%% file: main.tex
\documentclass[11pt,letterpaper]{mystyle}

\usepackage[authoryear,round, numbers]{natbib}

\makeatletter
\let\oldthebibliography\thebibliography
\let\endoldthebibliography\endthebibliography

\renewenvironment{thebibliography}[1]{%
  \NAT@numberstrue 
  \def\@biblabel##1{[##1]} 
  
  \setlength{\bibhang}{0pt}
  
  \oldthebibliography{#1}%
}{%
  \endoldthebibliography
}
\makeatother

\usepackage[utf8]{inputenc}
\usepackage[T1]{fontenc}
\usepackage{adjustbox}
\usepackage{float}
\usepackage{breakurl}    
\usepackage[edges]{forest}
\usepackage{subcaption}
\usepackage{soul}
\usepackage{multirow}
\usepackage[utf8]{inputenc} 
\usepackage{booktabs}       
\usepackage{amsfonts}       
\usepackage{nicefrac}      
\usepackage{microtype}     
\usepackage{xcolor}        
\usepackage{colortbl}
\usepackage{enumitem}
\usepackage{amssymb}
\usepackage{wrapfig}
\usepackage{courier}
\usepackage{bxcoloremoji}
\usepackage{CJKutf8}
\usepackage{xspace}
\usepackage{textcomp}
\usepackage{listings}
\usepackage{tabularray}
\usepackage{fontawesome5}
\usepackage{pifont}
\usepackage{makecell}
\usepackage{fancyhdr}
\usepackage{setspace}
\usepackage{threeparttable}
\usepackage{supertabular}
\usepackage{bm}
\usepackage{amsmath}
\usepackage{amsthm}
\usepackage{mathrsfs}
\usepackage{siunitx}
\usepackage{footmisc}
\usepackage{array}
\usepackage{CJK}
\usepackage{footnote}
\usepackage{tabularx}
\usepackage{tikz}

\newcolumntype{L}[1]{>{\raggedright\arraybackslash}p{#1}}
\usetikzlibrary{trees,positioning,shapes,shadows,arrows.meta}

\definecolor{hidden-draw}{RGB}{0,0,0}
\definecolor{hidden-blue}{RGB}{194,232,247}
\definecolor{hidden-orange}{RGB}{243,202,120}
\definecolor{hidden-yellow}{RGB}{242,244,193}
\definecolor{tree-level-1}{RGB}{245,20,85}
\definecolor{tree-level-2}{RGB}{246,86,118}
\definecolor{tree-level-3}{RGB}{248,177,193}
\definecolor{tree-leaf}{RGB}{176,230,198}

\tcbuselibrary{skins}

\definecolor{darkblue}{rgb}{0, 0, 0.5}
\definecolor{darkgreen}{RGB}{50,100,0}
\definecolor{darkred}{RGB}{200, 0, 0}
\definecolor{lightblue}{RGB}{220,235,250}
\hypersetup{colorlinks=true, citecolor=darkblue, linkcolor=darkblue, urlcolor=darkblue}

\definecolor{gray}{gray}{0.5}

\newcommand{\github}{\raisebox{-1.5pt}{\includegraphics[height=1.05em]{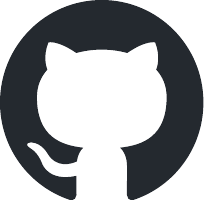}}}

\fancypagestyle{headstyle}{
    \fancyhead[C]{}
    \fancyhead[R]{}

}

\definecolor{hidden-red}{RGB}{205, 44, 36}
\definecolor{hidden-blue}{RGB}{194,232,247}
\definecolor{hidden-orange}{RGB}{243,202,120}
\definecolor{hidden-green}{RGB}{34,139,34}
\definecolor{hidden-pink}{RGB}{255,245,247}
\definecolor{hidden-black}{RGB}{20,68,106}
\definecolor{purple}{RGB}{144,153,196}
\definecolor{yellow}{RGB}{255,228,123}
\definecolor{hidden-yellow}{RGB}{255,248,203}
\definecolor{tkcolor}{RGB}{224,223,255}
\definecolor{darkblue}{rgb}{0, 0.40, 0.75}

\definecolor{lightblue}{RGB}{220,235,250}
\hypersetup{colorlinks=true, citecolor=darkblue, linkcolor=darkblue, urlcolor=darkblue}

\newcommand{\sps}{\,\,}

\usepackage{longtable}
\usepackage{arydshln}
\usepackage{array}
\usepackage[font=small,labelfont=bf]{caption}
\usepackage{colortbl} 
\usepackage{hyperref}

\usepackage{ragged2e}
\usepackage{tikz}
\usepackage{forest}
\useforestlibrary{edges}
\usetikzlibrary{arrows.meta,positioning,calc}

\tikzset{
  outline/.tip = {Latex[length=3mm]},
  box/.tip     = {Latex[length=3mm]},
}

\usepackage{bm}

\colorlet{outlineDraw}{black!35}
\colorlet{outlineBlue}{blue!12}
\colorlet{outlineYellow}{yellow!15}
\colorlet{outlineOrange}{orange!15}
\colorlet{outlineGreen}{green!12}
\colorlet{outlineGray}{gray!12}

\newtcolorbox{TakeawayBox}[2][]{takeawaybox,title=#2,#1}

\tcbset{
  takeawaysbox/.style={
    colback=lightblue!80,
    colframe=black,
    fonttitle=\bfseries\small,
    coltitle=white,
    colbacktitle=black,
    enhanced,
    attach boxed title to top left={xshift=2.5mm,yshift=-2.5mm},
    boxed title style={rounded corners, size=small, colframe=black, colback=black},
    width=\linewidth,
    arc=3.5mm
  }
}

\title{
Locate, Steer, and Improve: A Practical Survey of Actionable Mechanistic Interpretability in Large Language Models}

\author{
\textbf{Hengyuan Zhang$^{1, { } \coloremojicode{1F4AB}}$ \sps Zhihao Zhang$^{2, { } \coloremojicode{1F4AB}}$ \sps Mingyang Wang$^{3}$ \sps Zunhai Su$^{4}$ \sps Yiwei Wang$^{5}$} \\
\textbf{Qianli Wang$^{6}$ \sps Shuzhou Yuan$^{7}$ \sps Ercong Nie$^{3}$ \sps Xufeng Duan$^{8}$ \sps  Feijiang Han$^{9}$ \sps Qibo Xue$^{10}$ \sps  Zeping Yu$^{11}$}
\textbf{Chenming Shang$^{12}$ \sps Xiao Liang$^{13}$ \sps Jing Xiong$^{1}$ \sps Hui Shen$^{14}$ \sps Chaofan Tao$^{1}$ \sps Zhengwu Liu}$^{1}$
\textbf{Senjie Jin$^{2}$ \sps Zhiheng Xi$^{2}$ \sps Dongdong Zhang$^{15}$ \sps Sophia Ananiadou$^{11}$}  \\
\textbf{Tao Gui$^{2}$ \sps Ruobing Xie$^{16}$ \sps Hayden Kwok-Hay So$^1$ \sps Hinrich Schütze$^{3}$}   \\ \textbf{Xuanjing Huang$^2$ \sps Qi Zhang$^{2, { } \coloremojicode{2709}}$ \sps  Ngai Wong$^{1, { } \coloremojicode{2709}}$}  \\
$^1$The University of Hong Kong \sps  $^2$Fudan University \sps $^3$LMU Munich   
$^4$Tsinghua University \sps  $^5$Technische Universität Darmstadt \sps $^6$Technische Universität Berlin\\
 $^7$Technische Universität Dresden  \sps $^8$The Chinese University of Hong Kong \\
 $^{9}$University of Pennsylvania \sps $^{10}$Nanjing University \sps $^{11}$University of Manchester \\ $^{12}$Dartmouth College \sps
$^{13}$University of California, Los Angeles \\  $^{14}$University of Michigan \sps $^{15}$Microsoft \sps $^{16}$Tencent
}

\begin{document}

\begin{abstract}
  \vspace{5mm}
  \textbf{\large Abstract:}
Mechanistic Interpretability (MI) has emerged as a vital approach to demystify the opaque decision-making of Large Language Models (LLMs). 
However, existing reviews primarily treat MI as an observational science, summarizing analytical insights while lacking a systematic framework for \textit{actionable intervention}. 
To bridge this gap, we present a practical survey structured around the pipeline: \textit{``Locate, Steer, and Improve.''} We formally categorize \textit{Localizing} (diagnosis) and \textit{Steering} (intervention) methods based on specific \textit{Interpretable Objects} to establish a rigorous intervention protocol. 
Furthermore, we demonstrate how this framework enables tangible improvements in \textit{Alignment}, \textit{Capability}, and \textit{Efficiency}, effectively operationalizing MI as an actionable methodology for model optimization.
\textit{\textcolor{blue}{With actionable mechanistic interpretability evolving at a fast pace, we pledge to keep this survey up to date, ensuring it reflects the cutting-edge advances in this area.}}
  \vspace{5mm}

  $^{\coloremojicode{1F4AB}}$ \textit{Equal Contribution}    \quad $^{\coloremojicode{2709}}$ \textit{Corresponding Author} 

  \vspace{5mm}
  \textbf{Keywords}: Actionable Interpretability, Large Language Models, Localizing and Steering, Model Improvement
  \vspace{5mm}

  \coloremojicode{1F4C5} \textbf{Date}: April 13, 2026

  \github{} \textbf{Github Repo}: \url{https://github.com/rattlesnakey/Awesome-Actionable-MI-Survey}

  \coloremojicode{1F4E7} \textbf{Contact}: 
  \href{mailto:hengyuan.zhang88@gmail.com}{hengyuan.zhang88@gmail.com}
  \href{mailto:zhangzhihao19@fudan.edu.cn}{zhangzhihao19@fudan.edu.cn}
  \href{mailto:qz@fudan.edu.cn}{qz@fudan.edu.cn}
  \href{mailto:nwong@eee.hku.hk}{nwong@eee.hku.hk}

\end{abstract}
\maketitle

\vspace{3mm}
\pagestyle{headstyle}
\thispagestyle{empty}
\newpage
\tableofcontents

\clearpage

\section*{Paper Outline}
\label{app:paper_outline}
\tikzset{
  outline-box/.style={
    rectangle,
    draw=outlineDraw,
    rounded corners,
    text opacity=1,
    minimum height=1.5em,
    inner sep=2pt,
    align=center,
    fill opacity=.9,
    line width=0.6pt,
  },
  outline-leaf/.style={
    outline-box,
    align=left,
    inner xsep=3pt,
    inner ysep=4pt,
  }
}

\newcommand{\citeleaf}[1]{%
  \parbox[t]{\linewidth}{%
    \RaggedRight\footnotesize
    \setlength{\parskip}{0pt}\setlength{\parindent}{0pt}%
    \setlength{\emergencystretch}{2em}%
    #1%
  }%
}
\newcommand{\cites}[1]{\cite{#1}}

\begin{figure*}[!h]
\centering
\resizebox{!}{0.92\textheight}{%
\begin{forest}
forked edges,
for tree={
  grow=east,
  reversed=true,
  anchor=base west,
  parent anchor=east,
  child anchor=west,
  base=center,
  font=\large,
  outline-box,
  edge+={outlineDraw, line width=0.8pt},
  s sep=19pt,
  inner xsep=3pt,
  inner ysep=3pt,
  ver/.style={rotate=90, child anchor=north, parent anchor=south, anchor=center, transform shape},
},
where level=1{text width=9.5em, font=\large, text centered}{},
where level=2{text width=12em, font=\large, text centered}{},
where level=3{text width=17em, font=\large, text centered}{},
[
  {\textbf{Locate, Steer, and Improve: A Practical Survey of Actionable Mechanistic Interpretability in Large Language Models}},
  font=\large,
  ver,
  xshift=12mm, 
  s sep=29pt
  [
      {\textbf{Core Interpretable Objects of LLMs} (\S\ref{sec:core_objects})},
      fill=outlineBlue
      [
        {%
          \parbox[t]{\linewidth}{%
            \RaggedRight\large
            \setlength{\parskip}{0pt}\setlength{\parindent}{0pt}%
            \setlength{\emergencystretch}{2em}%
            \textbf{Token Embedding} (\S\ref{sec:token_embedding});\ %
            \textbf{Transformer Block and Residual Stream} (\S\ref{sect:block_stream});\ %
            \textbf{Multi-Head Attention (MHA)} (\S\ref{sec:mha});\ %
            \textbf{Feed-Forward Network (FFN)} (\S\ref{sec:ffn});\ %
            \textbf{Sparse AutoEncoder (SAE) Feature} (\S\ref{sec:sae_feature})%
          }%
        },
        outline-leaf,
        fill=outlineBlue,
        text width=71.8em 
      ]
  ]
  [, phantom, no edge, minimum height=125pt] 
  [
    {\textbf{Localizing Methods} (\S\ref{sec:localize_methods})},
    fill=outlineYellow
    [
      {\textbf{Magnitude Analysis} (\S\ref{sec:magtitude_analysis})},
      fill=outlineYellow
      [
        {\citeleaf{\cites{
        dettmers2022gpt3,tang-etal-2024-language,galichin2025have,
        zhang2025towards,an2025systematic,su2025kvsink,
        anthropic2023sae,su2025unveiling,
        jin2025massivevalues,bi2025unveiling,
        chuang2024dola,zhang-etal-2024-truthx,
        elhoushi-etal-2024-layerskip,men-etal-2025-shortgpt,
        xiao2023smoothquant,ashkboos2024quarot,yu2024super,cai2024pyramidkv,
        xiong2025dope,xiong2025atts,he2024zipcache,
        su2025rotatekv,yuan2025native,
        lai-etal-2024-style,liu2024unraveling,chen-etal-2024-learnable,chen2024qrnca,
        wang2025brainmap,andrylie2025sparseautoencoderscapturelanguagespecific,
        gurgurov2025languagearithmeticssystematiclanguage,
        xiao2023streamingllm,cancedda2024spectral,singh2024needs,wang2024transformers,
        zhou2025roleattentionheadslarge,sergeev2025optimizingmultimodallanguagemodels,
        sun-etal-2025-personality,bas2025steering,
        dumitru2024layer,tan2024dlo,lawson2025learningskipmiddlelayers}}},
        outline-leaf, 
        fill=outlineYellow,
        text width=58em 
      ]
    ]
    [
      {\textbf{Causal Attribution} (\S\ref{sec:causal_Attribution})},
      fill=outlineYellow
      [
        {\citeleaf{\cites{
        vig2020gender,meng2022ccs,zhang2023towards,stolfo-etal-2023-mechanistic,
        yucausal_emnlp2024,geiger2025causal,ferreira2025truthfulfabricatedusingcausal,
        yeo2025towards,ravindran2025adversarial,yuEntangledRepresentationsMechanistic2025,
        wang2023interpretability,geva-etal-2023-dissecting,tang-etal-2024-language,
        yu-ananiadou-2024-interpreting}}},
        outline-leaf,
        fill=outlineYellow,
        text width=58em 
      ]
    ]
    [
      {\textbf{Gradient Detection} (\S\ref{sec:gradient_detection})},
      fill=outlineYellow
      [
        {\citeleaf{\cites{
        li2016visualizing,sundararajan2017axiomatic,smilkov2017smoothgrad,shrikumar2017deeplift,
        enguehard2023sig,wu2023analyzing,hou2023layersaliency,tao2025sdtp,grains2507,
        dai2022knowledge,ircan_neurips2024,zhang2024cofitune,zhang2024improving,li-etal-2025-happened,li2025instructionreasoningdatashape,
        relp2025,gaf_acl2025,liu2025sensmerging,
        gmt2025,zhang2024linguistic,li-etal-2025-loracoe,
        atpstar2403.00745,wang2022interpretability,zhang2023towards,yin-neubig-2022-interpreting}}},
        outline-leaf, 
        fill=outlineYellow,
        text width=58em 
      ]
    ]
    [
      {\textbf{Probing} (\S\ref{sec:probing})},
      fill=outlineYellow
      [
        {\citeleaf{\cites{
        alain2016understanding,Probing_Classifiers,conneau2018,tenney2019bert,
        ravichander2020probing,juprobing_coling2024,
        arxiv2410_knowledgeconflict,zhang-etal-2024-truthx,orgad2025llms,you-etal-2025-probabilistic_emnlp2025,
        du2024tst,emnlp2025_headprobe,iclr2025_politicalprobe,
        kantamneni2025_sparseprobing,absorption2024}}},
        outline-leaf, 
        fill=outlineYellow,
        text width=58em 
      ]
    ]
    [
      {\textbf{Vocabulary Projection} (\S\ref{sec:vocab_project})},
      fill=outlineYellow
      [
        {\citeleaf{\cites{
        nostalgebraist2020,geva2021transformer,belrose2023eliciting,wang2023interpretability,jiang2024large,jiang2025interpretingeditingvisionlanguagerepresentations,wendler2024llamas,kargaran2025programming,phukan-etal-2024-peering,phukan-etal-2025-beyond,yugeswardeenoointerpreting,sakarvadia2023attention,yu2024understanding,jiang2025devils,kim2025interpreting,wang2025logitlens4llms,huo2024mmneuron,yuUnderstandingMitigatingGender2025a,shao2025benford,arad2025saes,dreyer2025attributing,muhamed2025decoding,gur2025enhancing,shu-etal-2025-survey}}},
        outline-leaf, 
        fill=outlineYellow,
        text width=58em 
      ]
    ]
    [
      {\textbf{Circuit Discovery} (\S\ref{sec:circuit_discovery})},
      fill=outlineYellow
      [
        {\citeleaf{\cites{
        elhage2021mathematical,olsson2022incontextlearninginductionheads,Hannacicuits_nips2023,yao2024circuits,
        goldowsky2023localizing,conmy2023automated,
        syedetal2024attribution,hanna2024have,huang-etal-2025-pierce,sundararajan2017axiomatic,
        stolfo-etal-2023-mechanistic,wang2023interpretability,
        pacd2025,mib2025,
        bricken2023monosemanticity,ameisen2025circuit,hanna-etal-2025-circuit}}},
        outline-leaf, 
        fill=outlineYellow,
        text width=58em 
      ]
    ]
  ]
  [
    {\textbf{Steering Methods} (\S\ref{sec:steer_methods})},
    fill=outlineYellow
    [
      {\textbf{Amplitude Manipulation} (\S\ref{sec:amplitude_manipulation})},
      fill=outlineYellow
      [
        {\citeleaf{\cites{
        tang-etal-2024-language,nie-etal-2025-mechanistic,
        goyalBreakingBadTokens2025,yeo-etal-2025-understanding,
        liuDevilNeuronsInterpreting2024,chandnaDissectingBiasLLMs2025,huang-etal-2025-pierce,
        liu2024unraveling,men-etal-2025-shortgpt,zhou2025on,niu-etal-2025-llama,
        ahsanElucidatingMechanismsDemographic2025,raimondiAnalysingMoralBias2025,
        gao2025hneuronsexistenceimpactorigin,pach2025sparse,galichin2025have,
        stoehr-etal-2024-activation,liu2025sensmerging,yao2025activation,wang2025two}}},
        outline-leaf, 
        fill=outlineYellow,
        text width=58em 
      ]
    ]
    [
      {\textbf{Targeted Optimization} (\S\ref{sec:target_optimization})},
      fill=outlineYellow
      [
        {\citeleaf{\cites{
        meng2022ccs,meng2023massediting,zhong2024seeking,zhang-etal-2023-fine,zhang2024cofitune,
        xu-etal-2025-lets,xirobusttickets,zhouORTicket,zhang2024linguistic,zhang2024lulafns,
        li2025safety,du2024tst,liPrecisionKnowledgeEditing2024,zhang2024interpreting,
        zhu-etal-2024-landermt,
        tang2024razorattention,NEURIPS2024_028fcbcf,NEURIPS2023_edbcb758,
        chen-etal-2025-inner,xia-etal-2025-tokenskip,
        tanvocab_2025arxiv,leeMechanisticUnderstandingAlignment2024}}},
        outline-leaf, 
        fill=outlineYellow,
        text width=58em 
      ]
    ]
    [
      {\textbf{Vector Arithmetic} (\S\ref{sec:vector_arithmetic})},
      fill=outlineYellow
      [
        {\citeleaf{\cites{
        rimsky-etal-2024-steering,van2024extending,lu2024investigating,postmus2024steering,
        turner2024steeringlanguagemodelsactivation,sharma2025steeringconceptualbiastransformer,
        wang2025beyond,bayat2025steering,weng2025safe,he2025saif,soo2025interpretable,goyalBreakingBadTokens2025,
        ilharcoediting2023,yadav2023ties,liu2025sensmerging,yao2025activation,
        shu-etal-2025-survey}}},
        outline-leaf, 
        fill=outlineYellow,
        text width=58em 
      ]
    ]
  ]
  [, phantom, no edge, minimum height=315pt]
  [
    {\textbf{Applications} (\S\ref{sec:applications})},
    fill=outlineOrange
    [
        \textbf{Improve Alignment} (\S\ref{sec:improve_alignment}), fill=outlineOrange
        [
            Safety and Reliability (\S\ref{sec:safety_reliabilty}), fill=outlineOrange
            [
              \citeleaf{
                \cites{huang-etal-2025-pierce,zhou2025on,chen2025towards,suauWhisperingExpertsNeural2024,gao2025hneuronsexistenceimpactorigin,templeton2024scaling,goyalBreakingBadTokens2025,yeo-etal-2025-understanding,zhao2025understanding,liPrecisionKnowledgeEditing2024,li2025safety,leeMechanisticUnderstandingAlignment2024,arditi2024refusal,zhao2025llms,yin2025refusalfallscliffsafety,ball2024understandingjailbreaksuccessstudy,wang2025surgical,wang2025refusal,ferreira2025truthfulfabricatedusingcausal,huang2025internalcausalmechanismsrobustly,chuang2024dola,chen2024incontext,zhang-etal-2024-truthx,orgad2025llms,he2025saif,stolfo2025improving,jiang2024refine,li2025training}
              },
              outline-leaf, fill=outlineOrange, text width=39em
            ]
        ]
        [
            Fairness and Bias (\S\ref{sec:fairness_bias}), fill=outlineOrange
            [
              \citeleaf{
                \cites{vig2020gender,caiLocatingMitigatingGender2024,ahsanElucidatingMechanismsDemographic2025,chandnaDissectingBiasLLMs2025,yuEntangledRepresentationsMechanistic2025,iclr2025_politicalprobe,liuDevilNeuronsInterpreting2024,guanMPFAligningDebiasing2025,chintamIdentifyingAdaptingTransformerComponents2023,yuUnderstandingMitigatingGender2025a}
              },
              outline-leaf, fill=outlineOrange, text width=39em
            ]
        ]
        [
            Persona and Role (\S\ref{sec:personalization}), fill=outlineOrange
            [
              \citeleaf{
                \cites{rimsky-etal-2024-steering,poterti-etal-2025-role,chen2025persona,handa2025personality,su-etal-2025-understanding,deng2025neuron,chen2024from,tak-etal-2025-mechanistic,yuan2025monolingual,ju2025probing,karny2025neural,banayeeanzade2025psychological,bas2025steering,lai-etal-2024-style}
              },
              outline-leaf, fill=outlineOrange, text width=39em
            ]
        ]
    ]
    [
        \textbf{Improve Capability} (\S\ref{sec:improve_capability}), fill=outlineOrange
        [
            Multilingualism (\S\ref{sec:multilingual_crosslingual}), fill=outlineOrange
            [
              \citeleaf{
                \cites{zhao-etal-2024-multilingual,gurgurov2025languagearithmeticssystematiclanguage,tang-etal-2024-language,liu-etal-2025-relation-specific,jing-etal-2025-lingualens,andrylie2025sparseautoencoderscapturelanguagespecific,brinkmann-etal-2025-large,wendler2024llamas,philippy2023identifying,mousi2024exploring,chi-etal-2023-cross,hinck-etal-2024-llava,zhang-etal-2025-shifcon,wang-etal-2025-lost-multilinguality,wang-etal-2025-language-mixing,nie-etal-2025-mechanistic,liu-etal-2025-tracing}
              },
              outline-leaf, fill=outlineOrange, text width=39em
            ]
        ]
        [
            Knowledge Management (\S\ref{sec:knowledge_management}), fill=outlineOrange
            [
              \citeleaf{
                \cites{meng2022ccs,meng2023massediting,chen2024querylocalization,chen2024qrnca,zhang2024lulafns,katz2024backwardlens,lai2025jola,muhamed2025dsg,goyalBreakingBadTokens2025,jin2024ph3,li2025taming,niu-etal-2025-llama,jin2025massivevalues,zhang2024linguistic,zhang2024cofitune,wu2024reft,du2024tst,emnlp2025_headprobe,chen2025stitching,yadav2023tiesmerging,liu2025sensmerging}
              },
              outline-leaf, fill=outlineOrange, text width=39em
            ]
        ]
        [
            Logic and Reasoning (\S\ref{sec:logic_reasoning}), fill=outlineOrange
            [
              \citeleaf{
                \cites{zhang2024interpreting,quirke2024understanding,yang2024chainofthoughtlargelanguagemodels,venhoff2025understandingreasoningthinkinglanguage,ward2025reasoningfinetuning_arxiv2025,troitskii-etal-2025-internal_emnlp2025,galichin2025have,wu2023analyzing,wangelicitingcot_aaai2026,sun-etal-2025-probing_emnlp2025,you-etal-2025-probabilistic_emnlp2025,cywinski2025interpretlatentreasoning,tanvocab_2025arxiv,hjer2025improvingreasoningperformancelarge,tang-etal-2025-unlocking,hong-etal-2025-reasoning,zhang2025uncoveringlatentchainthought,liu2025fractionalreasoninglatentsteering,sinii2025steeringllmreasoningbiasonly,li-etal-2025-feature}
              },
              outline-leaf, fill=outlineOrange, text width=39em
            ]
        ]
    ]
    [
        \textbf{Improve Efficiency} (\S\ref{sec:improve_efficiency}), fill=outlineOrange
        [
            Efficient Training (\S\ref{sec:efficient_training}), fill=outlineOrange
            [
              \citeleaf{
                \cites{zhu-etal-2024-landermt,song-etal-2024-sift,xu-etal-2025-lets,mondal-etal-2025-language,gurgurov2025sparsesubnetworkenhancementunderrepresented,zhao-etal-2024-multilingual,sergeev2025optimizingmultimodallanguagemodels,lai2025jola,li2025finetuningsubgraphsearchnew,hoogland2402developmental,singh2024needs,minegishi2025context,nandaprogress,liu2023omnigrok,furutatowards,qiye2024exploring,li2025find}
              },
              outline-leaf, fill=outlineOrange, text width=39em
            ]
        ]
        [
            Efficient Inference (\S\ref{sec:efficient_inference}), fill=outlineOrange
            [
              \citeleaf{
                \cites{xia-etal-2025-tokenskip,lei2025generictokencompressionmultimodal,guo-etal-2024-attention,ye2025fit,he2024zipcache,cai2024pyramidkv,tang2024razorattention,xiao2024duoattention,laitenberger2025layerswhenlearningskip,valade2024acceleratinglargelanguagemodel,elhoushi-etal-2024-layerskip,wang-etal-2023-hadskip,shelke2024towards,lawson2025learningskipmiddlelayers,men-etal-2025-shortgpt,lu2024not,su2025unveiling,liu2024unraveling,tan-etal-2024-neuron,dumitru2024layer,zhang2025towards,xiao2025exploring,ranjan2025mix,zeng2024lsaq}
              },
              outline-leaf, fill=outlineOrange, text width=39em
            ]
        ]
    ]
  ]
  [, phantom, no edge, minimum height=240pt]
  [
    {\textbf{Discussion and Challenges} (\S\ref{app:challenges})},
    fill=outlineGreen,
    text width=21em
  ]
  [
    {\textbf{Future Directions} (\S\ref{sec:future_directions})},
    fill=outlineGreen,
    text width=21em
  ]
]
\end{forest}%
}
\label{fig:outline-all}
\end{figure*}

\clearpage

\begin{figure*}[t]
    \centering
    \includegraphics[width=\linewidth]{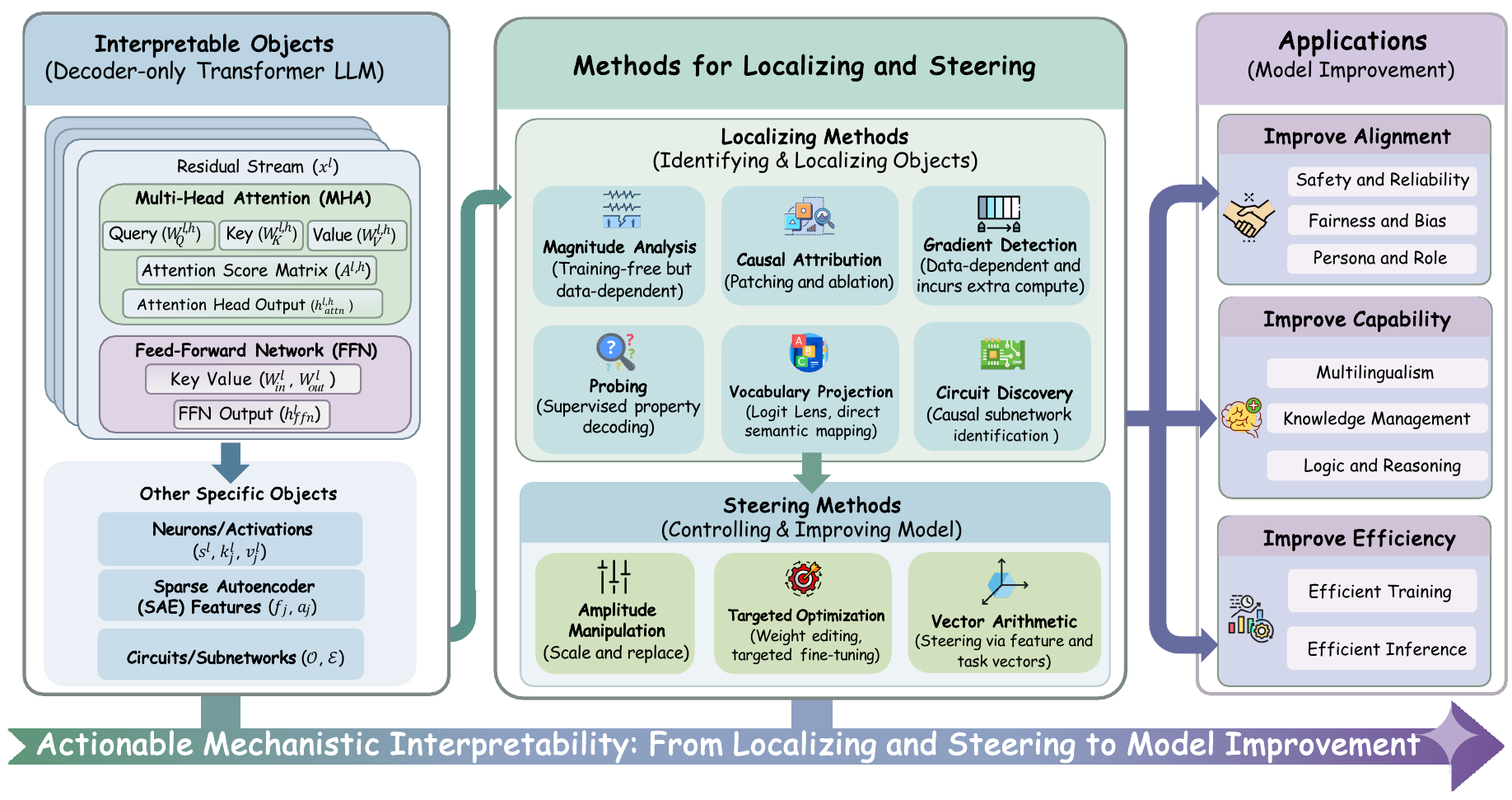}
    \vspace{-0.5cm}
    \caption{
    Overview of the paper structure. We begin by defining the core interpretable objects (\S\ref{sec:core_objects}) that form the foundation of our analysis. We then introduce a range of methods, ranging from localization (\S\ref{sec:localize_methods}) to steering (\S\ref{sec:steer_methods}). Finally, we illustrate how these methods can be applied to improve models (\S\ref{sec:applications}).}
    \vspace{-0.3cm}
    \label{fig:framework}
\end{figure*}

\section{Introduction}
\label{sec:intro}
Large Language Models (LLMs) have recently achieved remarkable success, demonstrating outstanding performance across a wide spectrum of applications, ranging from complex reasoning and multilingualism, to highly specialized domains~\citep{li2025system,ren2025deepseek,dubey2024llama,openai2024gpt4technicalreport,liang2025sws,qin2025survey,yang2024llm,chang2025treereview,li2024survey,zhao2024revolutionizing,yu-etal-2025-chain,qwen2025qwen25technicalreport}.
Despite these advancements, a critical challenge remains: the internal decision-making processes of these models are largely opaque, often operating as ``black boxes.''
This lack of transparency poses significant risks, particularly in safety-critical applications, and severely limits our ability to efficiently debug, control, and optimize model behaviors~\citep{huang2024exploring,hong2024cyclealign,zhang2025black}.
Consequently, \textit{Mechanistic Interpretability} (MI) has emerged as a pivotal research direction. 
Unlike traditional behavioral analysis, MI aims to ``reverse-engineer'' these complex neural networks, decomposing their intricate computations into understandable components and causal mechanisms~\citep{ferrando2024primer,zhao2024explainability,geiger2025causal}.

Current research in this field generally falls into two categories.
A significant body of work focuses on the \textit{theoretical and foundational aspects} of MI~\citep{rauker2023toward,allen2023physics,ferrando2024primer,zhao2024explainability,zheng2024attention,saphra2024mechanistic,lopez2025linguistic,geiger2025causal,gantla2025exploring}. These studies provide technical roadmaps for dissecting Transformer architectures and identifying fundamental units. 
However, they primarily prioritize \textit{scientific discovery}---aiming to elucidate the model's inner working mechanisms for the sake of understanding itself. They typically treat MI as an observational science, leaving the question of how to translate these microscopic insights into practical model improvements underexplored.

Recognizing the applied potential of interpretability, a second line of work has begun to \textit{bridge the gap between theoretical understanding and practical utilization}. These surveys discuss how MI techniques can be leveraged to aid downstream tasks or assist in specific domains~\citep{luo2024understanding,wu2024usable,rai2024practical,bereska2024mechanistic,lee2025interpretation,resck2025explainability,lin2025survey}.
However, despite their contributions, these existing reviews face two primary limitations that hinder broader adoption.
First, they often lack a \textit{sufficient categorization} and \textit{clear definition} of MI methods within practical application contexts. 
The distinction between diagnostic tools and intervention techniques is frequently blurred.
Second, their coverage of applications is often incomprehensive, and the illustration of methods is typically too general. 
This high-level abstraction makes it difficult for researchers to translate theoretical mechanistic insights into actionable interventions for specific problems.
This gap is particularly critical at the current stage of AI development. While the rapid progress of LLMs has been predominantly driven by \textit{external scaling factors}, e.g., larger models, more data, and increased computational resources, this paradigm for model improvement is increasingly encountering bottlenecks. 
We argue that further advancements must be refined through a deeper understanding of internal mechanisms, shifting the focus from simply scaling the ``outside'' to surgically improving the model from the ``inside.'' However, the field currently lacks a unified guide that systematically \textit{categorizes these internal methodologies} and presents a concrete \textit{pipeline for active model improvement.}

To address these challenges, we propose the \textit{``Locate, Steer, and Improve''} pipeline. This conceptual framework is designed to systematically transform MI from a passive observational science into an actionable intervention discipline. Our work makes the following key contributions:

\begin{itemize}[leftmargin=*]
    \item \textbf{1) A Rigorous Pipeline-Driven Framework:} We establish a structured framework for applying MI to real-world model optimization. We begin by defining the core \textbf{Interpretable Objects} within LLMs (e.g., neurons, attention heads, residual streams). Based on the application workflow, we clearly categorize methodologies into two distinct stages: \textbf{Localizing} (Diagnosis), which identifies the causal components responsible for specific behaviors, and \textbf{Steering} (Intervention), which actively manipulates these components to alter model outputs. Crucially, for each technique, we provide a detailed \textit{Methodological Formulation} along with its \textit{Applicable Objects} and \textit{Scope}, helping readers quickly understand the technical implementation and appropriate use cases.

    \item \textbf{2) Comprehensive Paradigms for Application:} We provide an extensive survey of MI applications organized around three major themes: \textit{Improve Alignment}, \textit{Improve Capability}, and \textit{Improve Efficiency}. These themes cover eight specific scenarios, ranging from safety and multilingualism to efficient training. Instead of merely listing relevant papers, we summarize representative \textit{MI application paradigms} for each scenario. This approach allows readers to quickly capture the distinct usage patterns of MI techniques across different application contexts, facilitating the transfer of methods to new problems.

    \item \textbf{3) Insights, Resources, and Future Directions:} We critically discuss the current challenges in actionable MI research and outline promising future directions. To facilitate further progress and lower the barrier to entry, we curate a comprehensive collection of \textbf{over 200 papers}, which are listed in Table~\hyperlink{paperlistall}{2}. These papers are systematically tagged according to their corresponding localizing and steering methods, providing a practical and navigable reference for the community.
\end{itemize}

\newpage

\section{Core Interpretable Objects of LLMs}
\label{sec:core_objects}
In this section, we establish a unified mathematical formulation for the core interpretable objects within LLMs. We focus specifically on the decoder-only Transformer architecture~\citep{radford2019language}, which serves as the predominant framework for contemporary state-of-the-art models~\citep{dubey2024llama,openai2024gpt4technicalreport,qwen2025qwen25technicalreport}. 
We present the core interpretable objects and their corresponding mathematical notations in Table~\ref{tab:interpretable_objects}.

\input{Tables/summary_object_notation}

\subsection{Token Embedding}
\label{sec:token_embedding}
The entry point of the model maps discrete tokens from a vocabulary $\mathcal{V}$ to continuous vector representations. We define the \textit{Embedding Matrix} as $\mathbf{W}_E \in \mathbb{R}^{|\mathcal{V}| \times d_{\text{model}}}$, where $|\mathcal{V}|$ denotes the vocabulary size and $d_{\text{model}}$ represents the hidden dimension of the model.
For a given input token $t_i$ at position $i$, its \textit{Token Embedding}---which also serves as the initial state of the residual stream, denoted as $\mathbf{x}_i^0$---is obtained by retrieving the corresponding vector from $\mathbf{W}_E$ and adding positional information:
\begin{equation}
\mathbf{x}_i^0 = \mathbf{W}_E[t_i] + \mathbf{p}_i
\end{equation}
\noindent where $\mathbf{p}_i$ is the positional embedding vector. It is worth noting that while earlier architectures used absolute positional embeddings added at the input, modern LLMs~\citep{dubey2024llama,openai2024gpt4technicalreport,qwen2025qwen25technicalreport} typically employ Rotary Positional Embeddings (RoPE)~\citep{su2024roformer}. In these architectures, positional information is applied directly to the query and key vectors within the attention mechanism rather than to the residual stream at the embedding layer.

\subsection{Transformer Block and Residual Stream}
\label{sect:block_stream}
Typically, an LLM is composed of $L$ stacked layers. Each layer $l$ consists of two primary blocks: a Multi-Head Attention (MHA) block and a Feed-Forward Network (FFN) block. The fundamental communication channel connecting these blocks is the \textit{residual stream}.

\begin{figure}[!ht]
\centering
\includegraphics[width=0.7\linewidth]{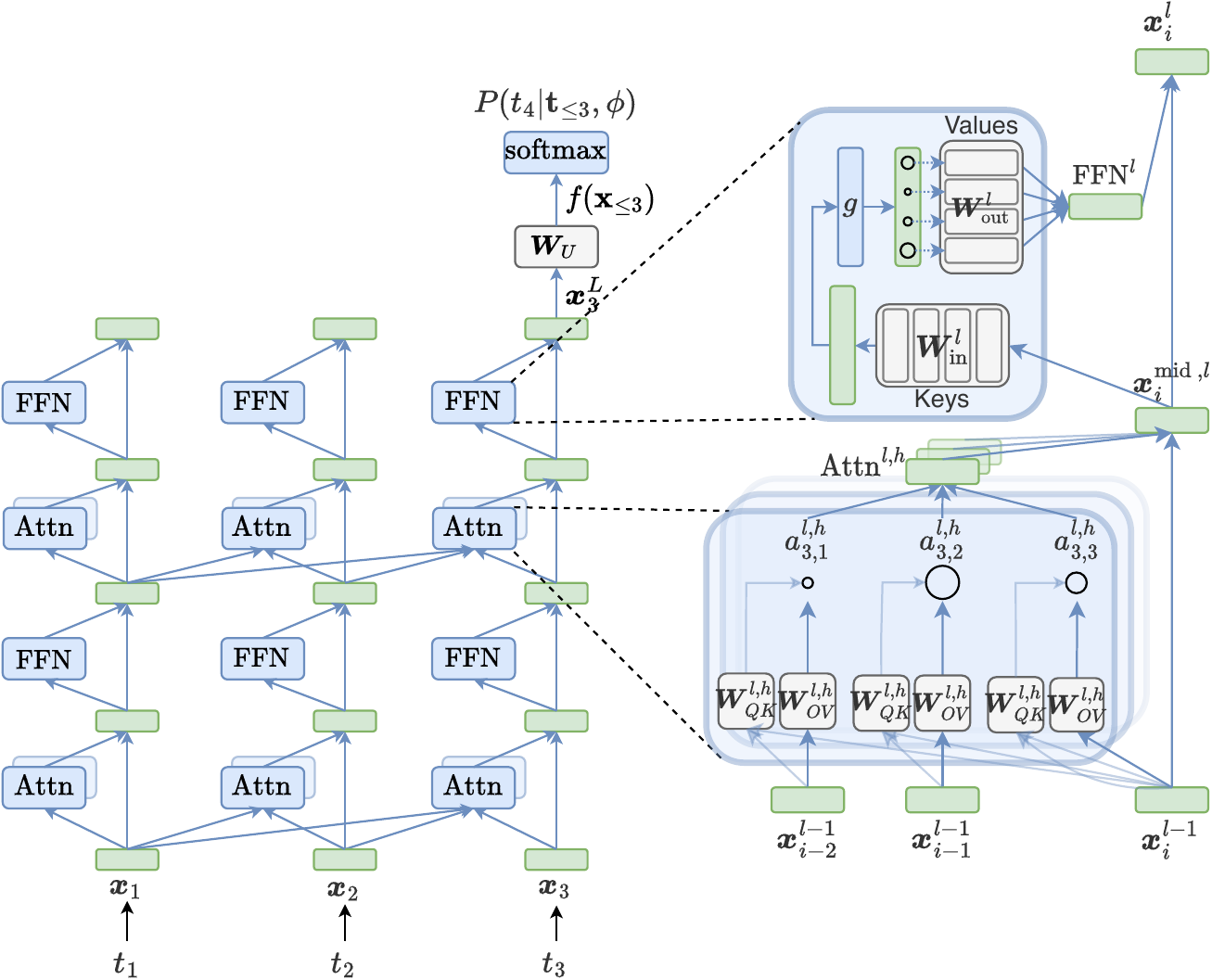}
\caption{The schematic of information flow within a standard Transformer block. The residual stream  ($\mathbf{x}_l$) serves as the backbone, while MHA and FFN act as additive branches that read from and write to this stream. Based on the figure from~\citet{ferrando2024primer}.}
\label{fig:transformer_flow}
\end{figure}

As illustrated in Figure~\ref{fig:transformer_flow}, the residual stream acts as the central ``highway'' for information propagation~\citep{elhage2021mathematical,bricken2021attention,meng2022ccs,meng2023massediting,zhang2024cofitune}. It preserves a shared memory state that is iteratively updated by the blocks. The update dynamics for the residual stream state\footnote{Here, $\mathbf{x}^l \in \mathbb{R}^{T \times d_\text{model}}$ represents the token-wise residual stream state of the input sequence at layer $l$, with $T$ tokens and hidden dimension $d_\text{model}$.} $\mathbf{x}^l$ at layer $l$ are defined as follows:
\begin{align}
\mathbf{x}^{l, \text{mid}} &= \mathbf{x}^l + \mathbf{h}_{\text{attn}}^{l}(\mathbf{x}^l) \\
\mathbf{x}^{l+1} &= \mathbf{x}^{l, \text{mid}} + \mathbf{h}_{\text{ffn}}^l(\mathbf{x}^{l, \text{mid}})
\end{align}
\noindent where $\mathbf{x}^{l, \text{mid}}$ represents the intermediate state after the MHA block but before the FFN block.\footnote{For simplicity and clarity in our mechanistic analysis, we omit Layer Normalization (LayerNorm or RMSNorm) terms from these equations. While crucial for training stability, normalization operations are often abstracted away in high-level interpretability studies to focus on the additive composition of features.}

This additive structure—where $\mathbf{x}^{l+1} = \mathbf{x}^l + \text{MHA}(\mathbf{x}^l) + \text{FFN}(\mathbf{x}^l + \text{MHA}(\mathbf{x}^l))$—is critical for MI analysis. 
It implies that features in the residual stream can be viewed as linear combinations of outputs from all previous components. This property enables the decomposition of the model's final prediction into individual component contributions, facilitating methods like ``Logit Lens''~\citep{nostalgebraist2020,wang2025logitlens4llms,liAnchoredAnswersUnravelling2025} and causal mediation analysis~\citep{meng2022ccs,meng2023massediting,goldowsky2023localizing,syedetal2024attribution,yeo2025towards}.

\subsection{Multi-Head Attention (MHA)}
\label{sec:mha}
The Multi-Head Attention mechanism allows tokens to contextualize information by attending to other positions in the sequence. It consists of $H$ independent heads, which primarily manage information routing and the resolution of contextual dependencies~\citep{elhage2021mathematical,olsson2022incontextlearninginductionheads,voita2019analyzing,feng2023language,men2024unlocking}.

\paragraph{Standard Formulation}
For a specific head $h$ at layer $l$, we define the learnable weight matrices as 
$\mathbf{W}_Q^{l,h}, \mathbf{W}_K^{l,h}, \mathbf{W}_V^{l,h} \in \mathbb{R}^{d_{\text{model}} \times d_{\text{head}}}$ 
and the output projection matrix as 
$\mathbf{W}_O^{l,h} \in \mathbb{R}^{d_{\text{head}} \times d_{\text{model}}}$.
Here, $T$ denotes the sequence length, the attention mechanism first computes the \textit{attention score} matrix
$\mathbf{A}^{l,h} \in \mathbb{R}^{T \times T}$, 
which represents the relevance of each token to every other token:
\begin{equation}
\mathbf{A}^{l,h} = 
\text{softmax}\!\left(
    \frac{(\mathbf{x}^{l} \mathbf{W}_Q^{l,h}) (\mathbf{x}^{l} \mathbf{W}_K^{l,h})^\top}{\sqrt{d_{\text{head}}}}
    + \mathbf{M}
\right),
\end{equation}
where $\mathbf{M} \in \mathbb{R}^{T \times T}$ denotes the attention mask that prevents attention to invalid positions 
(e.g., future tokens in causal attention or padding tokens).

Functionally, attention heads ``read'' information from the residual stream of previous tokens via the 
query--key subspace projections, and then ``write'' the attended information back to the current position 
via the value and output projections. 
The output for a single head $h$, denoted as $\mathbf{h}_{\text{attn}}^{l,h}$, is computed as:
\begin{equation}
\mathbf{h}_{\text{attn}}^{l,h} = 
\left[\mathbf{A}^{l,h} (\mathbf{x}^{l} \mathbf{W}_V^{l,h})\right] \mathbf{W}_O^{l,h}.
\end{equation}
The total output of the MHA block is the sum of the outputs from all $H$ heads: $\mathbf{h}_{\text{attn}}^{l} = \sum^{H}_{h=1}\mathbf{h}_{\text{attn}}^{l,h}$.

\paragraph{Mechanistic View: QK and OV Units}
While the standard formulation describes \textit{how} attention is computed, the \textit{unit} perspective~\citep{elhage2021mathematical} offers deeper insight into \textit{what} task each head performs. As illustrated in the detailed view of Figure~\ref{fig:transformer_flow}, each head can be decomposed into two functionally distinct units:

\textbf{1) The QK Unit ($\mathbf{W}_{QK}$):} This unit determines \textit{where} to attend. By merging the query and key matrices into a single low-rank matrix $\mathbf{W}_{QK}^{l,h} = \mathbf{W}_Q^{l,h} (\mathbf{W}_K^{l,h})^\top$, the attention pattern depends directly on the interaction between residual stream states. The attention score $a_{i,j}^{l,h}$ (e.g., $a_{3,1}$ in Figure~\ref{fig:transformer_flow}) is derived from the bilinear form $(\mathbf{x}_i^l)^\top \mathbf{W}_{QK}^{l,h} \mathbf{x}_j^l$.

\textbf{2) The OV Unit ($\mathbf{W}_{OV}$):} This unit determines \textit{what} information is transmitted. By merging the value and output matrices into $\mathbf{W}_{OV}^{l,h} = \mathbf{W}_V^{l,h} \mathbf{W}_O^{l,h}$, we can view the head's operation as reading a vector from the source token $j$, transforming it linearly via $\mathbf{W}_{OV}^{l,h}$, and adding it to the destination token $i$, weighted by the attention score. This separation allows researchers to classify heads into distinct roles, such as ``Induction Heads'' (which copy previous tokens) or ``Previous Token Heads''~\citep{olsson2022incontextlearninginductionheads,singh2024needs,wang2024transformers}.

\subsection{Feed-Forward Network (FFN)}
\label{sec:ffn}
\paragraph{Standard Formulation}
The Feed-Forward Network block acts as a position-wise feature transformer. Unlike attention heads, FFNs operate independently on each token position, applying non-linear transformations to the input. They are often conceptualized as ``Key-Value'' memories, where the first layer projects the stream into a high-dimensional state (detecting patterns or ``Knowledge Keys'') and the second layer writes the retrieved knowledge back to the stream~\citep{geva2021transformer,geva2022transformer,dai2022knowledge}.

Mathematically, the output of the FFN block $\mathbf{h}_{\text{ffn}}^l$ is given by :\footnote{While we use the standard formulation above to keep notation compact, it is important to note that many modern LLMs employ gated variants such as SwiGLU~\citep{shazeer2020gluvariantsimprovetransformer}.
These variants introduce an additional gating matrix $\mathbf{W}_{\text{gate}}^l$ and combine an element-wise gate with the projection before the final output: $\mathbf{h}_{\text{ffn}}^l
    =
    \left(\text{SiLU}(\mathbf{x}^{l,\text{mid}}\mathbf{W}_{\text{gate}}^l)
    \odot
    (\mathbf{x}^{l,\text{mid}}\mathbf{W}_{\text{in}}^l)\right)\mathbf{W}_{\text{out}}^l$. 
For the sake of generality, we present the standard FFN formulation here.}
\begin{equation}
\mathbf{h}_{\text{ffn}}^l = \sigma(\mathbf{x}^{l, \text{mid}} \mathbf{W}_{\text{in}}^l) \mathbf{W}_{\text{out}}^l
\end{equation}
where $\mathbf{x}^{l, \text{mid}}$ is the input to the FFN, and $\sigma$ is a non-linear activation function.
The weight matrices are defined as $\mathbf{W}_{\text{in}}^l \in \mathbb{R}^{d_{\text{model}} \times d_{\text{ffn}}}$ and $\mathbf{W}_{\text{out}}^l \in \mathbb{R}^{d_{\text{ffn}} \times d_{\text{model}}}$. 

\paragraph{Mechanistic View: Neurons}
In this context, the \textit{neuron} $j$ is defined as an atomic unit comprised of a pair of weights: the \textit{key weight} $\mathbf{k}_{j}^l$ (the $j$-th row of $\mathbf{W}_{\text{in}}^l$) and the \textit{value weight} $\mathbf{v}_{j}^l$ (the $j$-th column of $\mathbf{W}_{\text{out}}^l$). The intermediate state $\mathbf{s}^l = \sigma (\mathbf{x}^{l, \text{mid}} \mathbf{W}_{\text{in}}^l)$ represents the vector of \textit{neuron activation}.

\subsection{Sparse Autoencoder (SAE) Feature}
\label{sec:sae_feature}

While the internal objects described above (e.g., neuron activation $\mathbf{s}^l$ or residual stream state $\mathbf{x}^l$) are fundamental to the model's operation, they are often \textit{polysemantic}. This is due to the phenomenon of \textit{superposition}, where neural networks represent more features than they have physical neurons by encoding them as nearly orthogonal directions in the high-dimensional activation space~\citep{elhage2022superposition}. Consequently, a single neuron may activate for multiple unrelated concepts, making direct interpretation difficult.

\begin{figure}[!ht]
\centering
\centerline{\includegraphics[width=0.7\columnwidth]{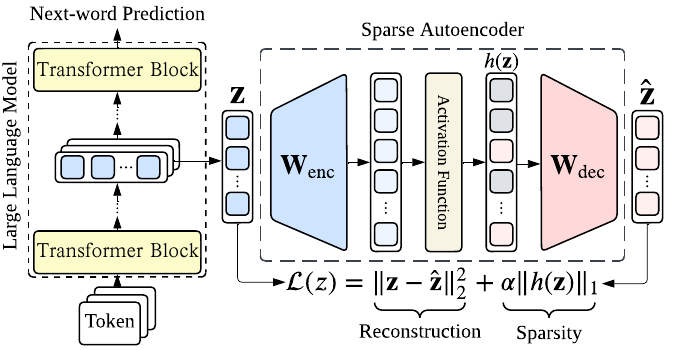}}
\caption{The framework of Sparse Autoencoders (SAEs). The SAE acts as an independent module attached to a frozen LLM, expanding dense representations into a sparse, overcomplete set of interpretable features via an encoder-decoder architecture. Based on the figure from~\citet{shu-etal-2025-survey}.}
\label{fig:sae_framework}
\end{figure}

Sparse Autoencoders (SAEs) provide a principled method to resolve this by disentangling dense, polysemantic representations into \textit{monosemantic features}~\citep{bricken2023monosemanticity}. As illustrated in Figure~\ref{fig:sae_framework}, an SAE acts as a ``microscope'' for the LLM. It projects low-dimensional dense activations into a higher-dimensional sparse latent space, effectively ``unpacking'' the superposition.

\paragraph{Mathematical Formulation}
SAEs are trained in a layer-wise manner as independent modules attached to a specific object of a frozen LLM. They can be applied to nearly all internal objects, including neuron activation $\mathbf{s}^l$, residual stream state $\mathbf{x}^l$, MHA output $\mathbf{h}_{\text{attn}}^{l}$, and FFN output $\mathbf{h}_{\text{ffn}}^{l}$~\citep{lieberum-etal-2024-gemma,he2024llama}. 
For instance, when applying an SAE to reconstruct a residual stream state $\mathbf{x}^l$, the forward pass is defined as:
\begin{align}
\mathbf{a} &= \sigma(\mathbf{x}^l\mathbf{W}_{\text{enc}} + \mathbf{b}_{\text{enc}})\\ \quad \hat{\mathbf{x}}^l &= \mathbf{a}\mathbf{W}_{\text{dec}} + \mathbf{b}_{\text{dec}}
\end{align}
\noindent where $\mathbf{W}_{\text{enc}} \in \mathbb{R}^{d_{\text{model}} \times d_{\text{SAE}}}$ and $\mathbf{W}_{\text{dec}} \in \mathbb{R}^{d_{\text{SAE}} \times d_{\text{model}}}$ are learnable weights. A critical hyperparameter here is the \textit{Expansion Factor}—the ratio of $d_{\text{SAE}}$ to $d_{\text{model}}$. To capture the vast number of features hidden in superposition, $d_{\text{SAE}}$ is typically set to be $16\times$ to $128\times$ larger than the model dimension~\citep{cunningham2023sparse,templeton2024scaling,bloom2024gpt2residualsaes,ghilardi2024efficient,mudide2024efficient,lieberum-etal-2024-gemma,he2024llama}.

The training objective is to minimize the reconstruction error while enforcing sparsity on the latent activations $\mathbf{a}$:
\begin{equation}
\mathcal{L} = \|\mathbf{x}^l - \hat{\mathbf{x}}^l\|_2^2 + \lambda \|\mathbf{a}\|_1
\end{equation}
In this framework, the \textit{SAE feature} $\mathbf{f}_j$ (the $j$-th row of $\mathbf{W}_{\text{dec}}$) represents a distinct semantic direction in the activation space. The \textit{SAE feature activation} $a_j$ (the $j$-th element of $\mathbf{a}$) quantifies the strength of this feature in the current input. Crucially, this decomposition transforms opaque vectors into an actionable vocabulary, allowing researchers to steer model behavior by targeting these granular, interpretable features~\citep{templeton2024scaling,lieberum-etal-2024-gemma,he2025saif,xu2025beyond,cho2025toward,li2025training}.

\paragraph{Challenges}
Training high-quality SAEs presents several practical and conceptual challenges. One common issue is \textit{dead latents}, where a substantial fraction of latent features remain inactive throughout training, effectively wasting model capacity. Methods such as \textit{ghost gradients} and periodic \textit{resampling} are often used to reactivate such units and improve feature utilization~\citep{bricken2023monosemanticity,shu-etal-2025-survey}. Another challenge is \textit{feature absorption}, in which broad and high-frequency features dominate training and suppress more specific, low-frequency features, making the learned representation less fine-grained and less interpretable. To address these issues, several improved variants, including \textit{Gated SAEs}, \textit{Top-K SAEs}, \textit{BatchTopK SAEs}, \textit{JumpReLU SAEs}, and \textit{Binary SAEs}, have been proposed to improve feature quality, sparsity control, and reconstruction fidelity~\citep{gao2024scaling,rajamanoharan2024improving,bussmann2024batchtopk,rajamanoharan2024jumping,cho2025binary}.
Beyond optimization difficulty, computational cost is a major barrier to the broader adoption of SAEs. Because SAEs are typically trained separately for individual layers, scaling them to modern LLMs requires repeating this process across many layers and often across multiple activation sites. As model size increases, the number of layers, hidden dimensions, and candidate latent features also grows, which substantially increases training time, memory usage, and storage cost. This challenge is further amplified by the highly overcomplete latent spaces commonly used in SAE training, which are often necessary to obtain sparse and interpretable features~\citep{ghilardi2024efficient,templeton2024scaling,mudide2024efficient}. As a result, training high-quality SAEs at scale remains expensive and can limit their use to only a subset of layers or models.
A further concern is faithfulness. Since SAEs act as replacement or surrogate models for the original activations, their usefulness depends critically on how well they reconstruct the original internal states while preserving downstream behavior. If reconstruction quality is imperfect, even small residual errors may accumulate and cause the SAE-reconstructed activations to deviate from the original model dynamics~\citep{lieberum-etal-2024-gemma,he2024llama}. 

\newpage

\section{Localizing Methods}
\label{sec:localize_methods}
Localizing Methods aim to identify interpretable objects that are responsible for a particular behavior or encode specific information. 
These techniques serve as a diagnostic step to narrow down the search space to manageable functional units. 
By pinpointing key components such as specific neurons, attention heads, or SAE features, they provide the necessary foundation for subsequent detailed mechanism analysis and targeted model steering.

\subsection{Magnitude Analysis}
\label{sec:magtitude_analysis}

\paragraph{Methodological Formulation}
\textit{Magnitude Analysis} methods serve as a fundamental heuristic in interpretability, operating on the premise that internal elements with larger numerical values often exert greater influence on the model's computation. 
It scores internal objects via a scalar function to identify salient components~\citep{dettmers2022gpt3,tang-etal-2024-language,galichin2025have}.

Formally, consider a set of internal objects $\mathcal{O} = \{o_1, o_2, \dots, o_N\}$, where each $o_j$ represents a candidate element (e.g., a specific weight parameter row, a neuron, an SAE feature, or an attention head). 
We define an \textit{Importance Score} $s_j$ for each object using a magnitude function $f(\cdot)$:
\begin{equation}
s_j = f(o_j), \quad \text{e.g., } s_j = \lVert o_j \rVert_p \text{ or } s_j = \max_{k} |(o_j)_k|
\end{equation}
Common choices for $f(\cdot)$ include the $L_2$-norm ($\lVert \cdot \rVert_2$) to measure the aggregate energy, the $L_\infty$-norm (max-value) to capture peak activation, or frequency-based metrics. Based on these scores, a subset of salient objects $\mathcal{O}_{\text{salient}}$ is selected for further inspection or intervention, typically via a thresholding mechanism or a top-$k$ ranking:
\begin{equation}
\mathcal{O}_{\text{salient}} = \{ o_j \mid s_j \geq \tau \} \quad \text{or} \quad \operatorname*{arg\,topk}_{j \in \{1,\dots,N\}} s_j
\end{equation}

\begin{figure}[!ht]
    \centering
    \includegraphics[width=0.9\linewidth]{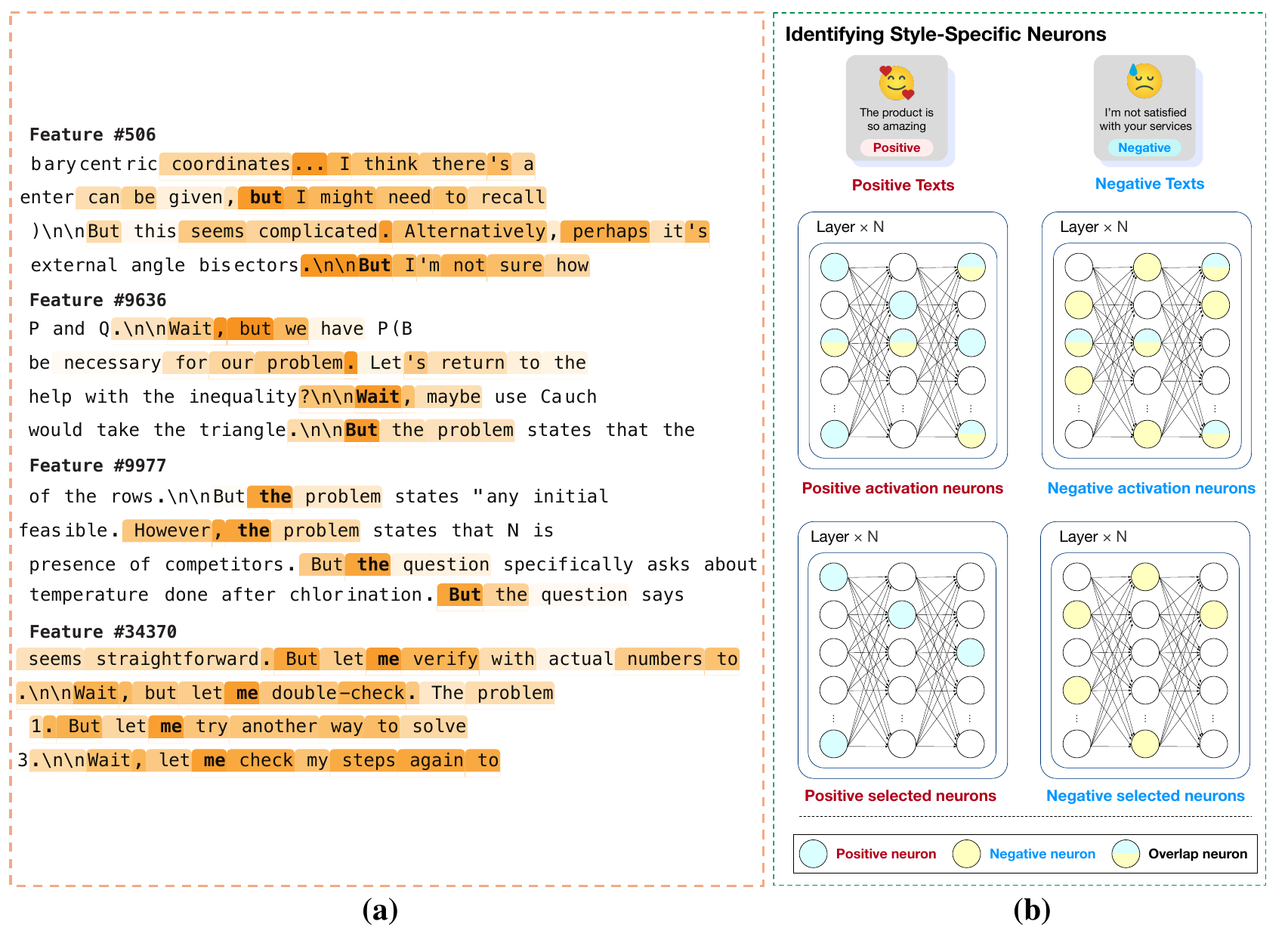}
    \caption{\textbf{Localization via Magnitude Analysis.}  \textbf{(a)} Discovery of SAE reasoning features~\citep{galichin2025have}: SAE features are scored using \textit{ReasonScore}, which aggregates activation magnitude and frequency during reasoning steps, isolating sparse features that encode cognitive behaviors like uncertainty or reflection. \textbf{(b)} Identification of Style-Specific Neurons~\citep{lai-etal-2024-style}: Neurons are ranked by their average activation magnitude on style-specific corpora, revealing clusters that selectively activate for distinct linguistic styles.}
    \label{fig:magnitude_analysis_examples}
\end{figure}

\paragraph{Applicable Objects}
This method applies broadly to both static structure and dynamic computation. 
We categorize the applicable objects as follows:

\textbf{1) Static Parameters:} In the context of model weights, \textit{Magnitude Analysis} is often used to identify outliers or ``heavy hitters'' without running inference. Researchers typically compute per-weight or per-row norms of weight matrices (e.g., $\lVert \mathbf{W}_{\text{in}}[j,:] \rVert$) to highlight parameters that dominate the inner product computations. 
These high-magnitude weights are often associated with critical knowledge storage or outlier features~\citep{dettmers2022gpt3,xiao2023smoothquant,ashkboos2024quarot,yu2024super,cai2024pyramidkv,xiong2025dope,xiong2025atts,he2024zipcache,zhang2025towards,an2025systematic,su2025kvsink,su2025rotatekv,yuan2025native,jin2025massivevalues,han2026zerotuning}.

\textbf{2) Dynamic Components (Neurons, SAE Features, or Attention Heads):} For functional units whose activity varies with input, ranking them by their activation statistics helps localize specialized capabilities~\citep{tang-etal-2024-language,lai-etal-2024-style,galichin2025have,liu2024unraveling,chen-etal-2024-learnable,chen2024qrnca,bi2025unveiling,wang2025brainmap,andrylie2025sparseautoencoderscapturelanguagespecific,gurgurov2025languagearithmeticssystematiclanguage}.

\begin{itemize}[leftmargin=*]
    \item \textbf{Specialized Neurons and SAE Features:} By feeding domain-specific datasets into the model and monitoring activations (e.g., neuron activation state $\mathbf{s}^l$ or SAE feature activation state $\mathbf{a}$), researchers can isolate components dedicated to specific concepts.
    For instance, in the context of higher-level reasoning, \citet{galichin2025have} utilized SAEs to disentangle the residual stream state $\mathbf{x}^l$. 
    As shown in Figure~\ref{fig:magnitude_analysis_examples} (a), they proposed a metric called \textit{ReasonScore}, which aggregates the activation frequency and magnitude of SAE features $a_j$ specifically during ``reasoning moments'' (e.g., when the model meets tokens like ``Wait'' or ``Therefore''). 
    By ranking features based on this score, they successfully localized \textit{Reasoning-Relevant} SAE features that encode abstract concepts like uncertainty or exploratory thinking.
    Similarly, for style transfer, \citet{lai-etal-2024-style} employed \textit{Magnitude Analysis} to identify \textit{Style-Specific Neurons}. As illustrated in Figure~\ref{fig:magnitude_analysis_examples} (b), they calculated the average activation magnitude of FFN neurons across corpora with distinct styles (e.g., positive vs. negative). Neurons that exhibited significantly higher average activation for the source style compared to the target style were identified as ``Source-Style Neurons,'' serving as candidates for subsequent deactivation.

    \item \textbf{Attention Heads:} The magnitude and distribution of \textit{attention scores} ($\mathbf{A}^{l,h}$) serve as a direct indicator of a head's functional role~\citep{xiao2023streamingllm,cancedda2024spectral,singh2024needs,wang2024transformers,bi2025unveiling,zhou2025roleattentionheadslarge,sergeev2025optimizingmultimodallanguagemodels}.
    For instance, \citet{zhou2025roleattentionheadslarge} introduced the \textit{Safety Head ImPortant Score (Ships)}, which aggregates attention weights on refusal-related tokens to localize ``Safety Heads'' critical for model alignment.
    In the multimodal domain, \citet{sergeev2025optimizingmultimodallanguagemodels} and \citet{bi2025unveiling} measured the concentration of attention mass on image tokens versus text tokens, successfully pinpointing heads responsible for visual perception and cross-modal processing.
    Similarly, \citet{singh2024needs} measured ``induction strength''---derived from the attention probability assigned to token repetition patterns---to track the formation and importance of Induction Heads.
\end{itemize}

\textbf{3) Layer-wise Representations:} Furthermore, measuring the magnitude of \textit{layer-wise distances} reveals structural roles. 
Comparing representations across contrastive inputs (e.g., $\lVert \mathbf{x}^l - \mathbf{x}'^{\,l} \rVert$) localizes layers where task-specific information diverges most strongly~\citep{chuang2024dola,zhang-etal-2024-truthx,sun-etal-2025-personality,bas2025steering}, whereas comparing consecutive layers (e.g., $\lVert \mathbf{x}^l - \mathbf{x}^{l+1} \rVert$) identifies layers with minimal state updates, pointing to redundant computation~\citep{dumitru2024layer,elhoushi-etal-2024-layerskip,tan2024dlo,lawson2025learningskipmiddlelayers,men-etal-2025-shortgpt}.

\paragraph{Characteristics and Scope}
The scope of \textit{Magnitude Analysis} for dynamic quantities is characterized as \textbf{training-free but data-dependent}.
\begin{itemize}[leftmargin=*]
    \item \textbf{Advantages:} It does not require training auxiliary classifiers or performing computationally expensive backward passes. This makes it highly scalable and suitable for analyzing large models in real-time.
    \item \textbf{Limitations:} It serves primarily as a \textit{lightweight heuristic}. High activation magnitude implies high presence but does not guarantee causal necessity (e.g., a high-magnitude feature might be cancelled out by a subsequent layer). Furthermore, its success relies heavily on the quality of the input data; if the dataset fails to elicit the specific behavior, the relevant components will remain dormant. Therefore, \textit{Magnitude Analysis} is typically used as a ``first-pass'' screening tool to filter candidate objects for more rigorous verification methods.
\end{itemize}

\subsection{Causal Attribution}
\label{sec:causal_Attribution}

\paragraph{Methodological Formulation}
\textit{Causal Attribution} methods constitute the gold standard for localization in MI. Unlike correlation-based analyses, these techniques identify which internal objects are \emph{causally responsible} for a specific model behavior by systematically measuring the effect of controlled interventions~\citep{vig2020gender,meng2022ccs,zhang2023towards,stolfo-etal-2023-mechanistic,yucausal_emnlp2024,geiger2025causal,ferreira2025truthfulfabricatedusingcausal,yeo2025towards}.

Formally, let $F(\cdot)$ denote a scalar model output of interest, such as the logit or probability of a target token. Let $o$ be an internal object (e.g., a neuron activation $s^l_j$ or a head output $\mathbf{h}^{l,h}_{\text{attn}}$) defined in \S\ref{sec:core_objects}. 
To evaluate the \textit{causal effect} of $o$, we compare the model's output under a counterfactual intervention against the baseline state:
\begin{equation}
\operatorname{do}(o \leftarrow \tilde{o}): \Delta F(o) = F\!\left(\operatorname{do}(o \leftarrow \tilde{o})\right) - F(o)
\end{equation}
where $F(o)$ represents the model's behavior in the standard ``clean'' run, and $\operatorname{do}(o \leftarrow \tilde{o})$ represents the intervention where the object $o$ is forced to take on a modified value $\tilde{o}$ while all other causal factors are held constant (ceteris paribus).
The intervention typically takes two forms: \textit{Patching} (where $\tilde{o}$ is an activation computed from a counterfactual input) or \textit{Ablation} (where $\tilde{o}$ is set to zero or a mean vector). A large magnitude $|\Delta F(o)|$ indicates that the object $o$ acts as a critical mediator or information node for the behavior encoded by $F$.

\begin{figure}[!t]
\centering
\centerline{\includegraphics[width=0.95\columnwidth]{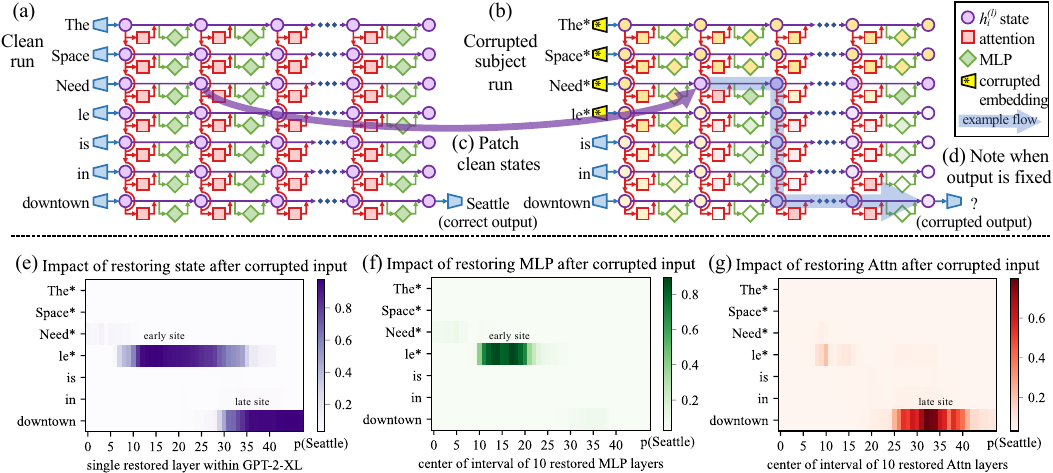}}
\caption{Overview of \textit{Causal Tracing}. The method identifies critical internal states by creating a corrupted run (noising the subject ``Space Needle'') and systematically restoring clean states to see which ones recover the prediction ``Seattle''. The heatmap results reveal that factual information is processed in early MLP layers at the subject position and later transferred to the final token via attention. Based on the figure from~\citet{meng2022ccs}.}
\label{fig:causal_tracing}
\end{figure}

\paragraph{Applicable Objects}
This analysis primarily targets \textbf{dynamic objects} involved in the inference process, including the residual stream state $\mathbf{x}^l$, the output of FFN $\mathbf{h}_{ffn}^l$, and the output of specific attention head $\mathbf{h}_{attn}^{l,h}$.

\textbf{1) Patching (Interchange Intervention):} This approach replaces an object computed from the original input with one computed from a \textit{counterfactual input} to isolate specific information pathways. By systematically patching across layers and positions, one can localize exactly where task-specific information (e.g., factual knowledge) is introduced or transformed~\citep{meng2022ccs,zhang2023towards,yeo2025towards,ravindran2025adversarial,yuEntangledRepresentationsMechanistic2025}.

We exemplify this mechanism using \textit{Causal Tracing}~\citep{meng2022ccs}, a representative technique designed to localize factual associations (e.g., ``The Space Needle'' $\rightarrow$ ``Seattle''). 
As illustrated in Figure~\ref{fig:causal_tracing}, this process involves three key steps:
\begin{itemize}
    \item \textbf{Corrupted Run (Intervention):} First, the specific knowledge is erased from the model's computation. A \textit{corrupted input} is created by adding Gaussian noise to the embeddings of the subject tokens (e.g., ``Space Needle''), causing the probability of the correct prediction (``Seattle'') to drop significantly.
    \item \textbf{Patched Run (Restoration):} The core operation systematically restores specific internal states. For a specific layer $l$ and token position $i$, the method copies the hidden activation from a separate original \textit{clean run} and pastes (restores) it into the corrupted computation graph.
    \item \textbf{Effect Measurement:} The causal effect is quantified by the \textit{Indirect Effect (IE)}, which measures how much of the original target probability is recovered by this restoration. A high IE score implies that the patched state at $(l, i)$ carries critical information.
\end{itemize}
Through this rigorous process, \citet{meng2022ccs} revealed that factual recall relies on two distinct localized mechanisms: an early retrieval phase in the \textit{FFN blocks} at subject tokens, and a late information transport phase in the \textit{MHA blocks} at the final token.

\textbf{2) Ablation (Knockout):} Alternatively, ablation-based attribution explicitly ``zeros out'' or removes objects, such as masking specific attention heads $\mathbf{h}_{attn}^{l,h}$ or neurons, and measures the resulting performance drop to determine their causal necessity~\citep{wang2023interpretability,geva-etal-2023-dissecting,yucausal_emnlp2024,tang-etal-2024-language,yu-ananiadou-2024-interpreting}.
This rigorous verification has been applied across various domains:
\citet{wang2023interpretability} and \citet{yucausal_emnlp2024} employed ablation to isolate minimal heads responsible for indirect object identification and in-context learning, respectively.
In the context of specialized capabilities, \citet{yu-ananiadou-2024-interpreting} utilized pruning (permanent ablation) to identify heads critical for arithmetic reasoning, while \citet{tang-etal-2024-language} masked specific neurons to demonstrate the existence of language-specific functional regions.
Furthermore, \citet{geva-etal-2023-dissecting} applied blocking interventions to dissect the precise roles of FFN value vectors in factual recall mechanisms.

\paragraph{Characteristics and Scope}
The scope of \textit{Causal Attribution} is characterized as \textbf{rigorously causal but computationally intensive}.
\begin{itemize}[leftmargin=*]
    \item \textbf{Advantages:} Unlike \textit{Magnitude Analysis} (\S\ref{sec:magtitude_analysis}), which only establishes correlation, \textit{Causal Attribution} provides definitive evidence that a component is a functional driver of the model's output. This allows researchers to distinguish essential mechanisms from features that are highly activated but causally irrelevant to the specific behavior.
    \item \textbf{Limitations:} This rigor incurs a significant computational overhead. Verifying causality typically requires intervening on objects individually and performing a separate forward pass for each intervention. Consequently, the cost scales linearly with the number of objects analyzed, making it prohibitively expensive for dense, sweeping searches over large models. This inefficiency often necessitates the use of \textit{Gradient Detection} (\S\ref{sec:gradient_detection}), which utilizes gradients to rapidly approximate these causal effects, enabling efficient screening before performing expensive, fine-grained interventions.
\end{itemize}

\subsection{Gradient Detection}
\label{sec:gradient_detection}

\paragraph{Methodological Formulation}
\textit{Gradient Detection} methods localize influential internal objects by scoring them with the sensitivity of a scalar target $F(x)$ (e.g., a logit, margin, or loss) with respect to an object $o_j$:
$s_j(x)=\phi(\nabla_{o_j}F(x),o_j)$, where common instantiations include the gradient norm $s_j=\|\nabla_{o_j}F(x)\|$ and the gradient--input score $s_j=\nabla_{o_j}F(x)^\top o_j$ \citep{li2016visualizing,sundararajan2017axiomatic}.
These scores serve as fast, first-order proxies for intervention effects.
Specifically, under an additive modification $o_j \mapsto o_j + \Delta o_j$, a first-order Taylor expansion yields
\begin{equation}
F(o_j + \Delta o_j) - F(o_j)
= \nabla_{o_j}F(x)^\top \Delta o_j + \mathcal{O}(\|\Delta o_j\|^2),
\end{equation}
indicating that the dot product $\nabla_{o_j}F(x)^\top \Delta o_j$ represents the directional derivative of $F$ along $\Delta o_j$.
A common local “removal” surrogate sets $\Delta o_j = -o_j$, giving $F(o_j - o_j) - F(o_j) \approx -\nabla_{o_j}F(x)^\top o_j$, which motivates using $\nabla_{o_j}F(x)^\top o_j$ (or its magnitude) as a signed influence score. 

To mitigate saturation and explicitly model the notion of “absence,” \textit{Integrated Gradients} (IG) attribute the change from a baseline $\tilde{o}_j$ to the input $o_j$ by integrating gradients along the straight-line path $\gamma(\alpha) = \tilde{o}_j + \alpha (o_j - \tilde{o}_j)$:
\begin{equation}
\mathrm{IG}_k(o_j; \tilde{o}_j)
= (o_j - \tilde{o}_j)_k 
\int_{0}^{1} 
\frac{\partial F(\gamma(\alpha))}{\partial \gamma_k}\, d\alpha,
\qquad
\sum_{k} \mathrm{IG}_k(o_j; \tilde{o}_j)
= F(o_j) - F(\tilde{o}_j),
\end{equation}
where $k$ indexes the components of $o_j$, and each $\mathrm{IG}_k$ quantifies the contribution of the $k$-th component to the output difference $F(o_j) - F(\tilde{o}_j)$.
In practice, the integral is approximated by an $m$-step Riemann sum over $\alpha = t/m$ \citep{sundararajan2017axiomatic}.
Scores are typically computed over a dataset $\mathcal{D}$ and aggregated to stabilize rankings (e.g., $\mathbb{E}_{x \sim \mathcal{D}}[s_j(x)]$ or $\mathbb{E}[|s_j(x)|]$), without explicitly applying perturbations during scoring.

\begin{figure}[!t]
  \centering
  \includegraphics[width=\linewidth]{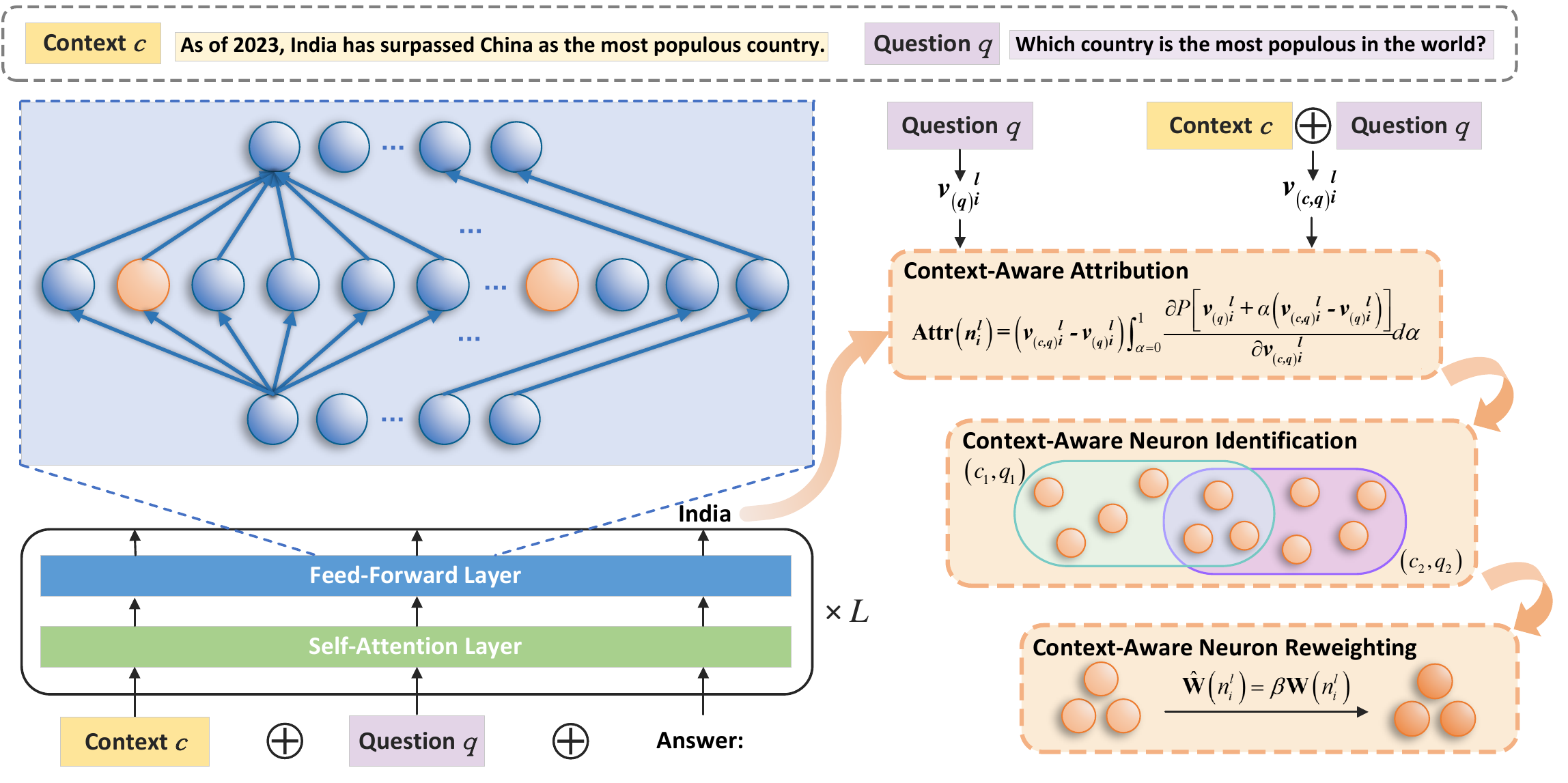}
  \caption{\textbf{Neuron-level gradient-based localization for mitigating knowledge conflicts.}
   It first calculates the \textit{Integrated Gradients} score for each neuron to measure its contribution to processing the context. It then identifies context-aware neurons by taking the intersection of neurons with the highest scores. Subsequently, the identified neurons are reweighted to guide the model to be more aligned with the contextual knowledge, ensuring greater fidelity to the context. Based on the figure from~\citet{ircan_neurips2024}.}
  \label{fig:ircan_fig1_gradient_neurons}
\end{figure}

\paragraph{Applicable Objects}
Because $F$ is differentiable with respect to any internal object $o_j$ (Table~\ref{tab:interpretable_objects}), \textit{Gradient Detection} applies uniformly across \textit{inputs}, \textit{activations}, and \textit{parameters}. Below we expand the object categories and make explicit the correspondence between symbols and the underlying model components.

\textbf{1) Inputs and Layer-wise States ($\mathbf{x}^0_i, \mathbf{x}^l$):}
For input embeddings $\mathbf{x}^0_i$ and the residual stream state $\mathbf{x}^l$, gradients directly quantify how sensitive $F(x)$ is to changes in specific prompt components and their propagated representations.
In practice, one computes $\nabla_{\mathbf{x}^0_i}F(x)$ or $\nabla_{\mathbf{x}^l}F(x)$ and derives token-level influence, such as the gradient norm $\|\nabla_{\mathbf{x}^0_i}F(x)\|$, the gradient--input score $\nabla_{\mathbf{x}^0_i}F(x)^\top \mathbf{x}^0_i$, or integrated gradients \citep{enguehard2023sig}.
Aggregating these scores across positions $i$ (optionally across layers $l$) yields a ranked view of which tokens or contextual spans are most responsible for a target output, as used to analyze CoT prompting~\citep{wu2023analyzing} and which depth regions contribute most strongly to the formation of that output \citep{hou2023layersaliency}, with closely related layer-/token-saliency signals also supporting dynamic token pruning \citep{tao2025sdtp} and inference-time steering \citep{grains2507}.

\textbf{2) Intermediate Outputs:}
Beyond inputs, \textit{Gradient Detection} can score internal computational units whose activations vary with the input.

\begin{itemize}[leftmargin=*]
    \item \textbf{Neurons ($\mathbf{s}^l$):} 
    A standard neuron-level object is the FFN activation vector $\mathbf{s}^l$ at layer $l$.
    Gradients $\nabla_{\mathbf{s}^l}F(x)$ can be converted into per-neuron scores to rank neurons by their local influence on $F$.
    This has been used to localize knowledge- or context-sensitive neurons and analyze their dependencies \citep{dai2022knowledge,ircan_neurips2024,zhang2024cofitune,zhang2024improving,li-etal-2025-happened,li2025instructionreasoningdatashape}.
    Figure~\ref{fig:ircan_fig1_gradient_neurons} illustrates a concrete LLM-specific instance: \citet{ircan_neurips2024} computes \textit{Integrated Gradients} scores to identify neurons most responsible for processing contextual cues under knowledge conflicts (via a context-aware attribution and a high-score intersection criterion), and then reweights the identified neurons to promote context-consistent generation.

    \item \textbf{Attention Head Outputs ($\mathbf{h}_{\text{attn}}^{l,h}$):}
    \textit{Gradient Detection} also applies to attention-related activations such as the attention head output $\mathbf{h}_{\text{attn}}^{l,h}$.
    Computing $\nabla_{\mathbf{h}_\text{attn}^{l,h}}F(x)$ and scalarizing it with $s_j(x)=\phi(\nabla_{\mathbf{h}_{\text{attn}}^{l,h}}F(x), \mathbf{h}_\text{attn}^{l,h})$ yields head-level rankings that can highlight salient heads or attention submodules for further analysis and subsequent intervention \citep{relp2025,gaf_acl2025,liu2025sensmerging}.
\end{itemize}

\textbf{3) Parameters ($\mathbf{W}_{Q/K/V/O}^{l,h},  \mathbf{W}_{\text{in/out}}^l$):}
Because $F$ is differentiable with respect to model weights, \textit{Gradient Detection} can score parameters at multiple granularities.
At the block level, common targets include attention projection matrices $\mathbf{W}_{Q/K/V/O}^{l,h}$ and FFN matrices $\mathbf{W}_{\text{in/out}}^l$. 
Gradients such as $\nabla_{\mathbf{W}_{Q}^{l,h}}F(x)$ can be turned into scalar salience measures (e.g., $\|\nabla_{\mathbf{W}}F(x)\|$) to rank influential attention/FFN modules \citep{relp2025,gaf_acl2025,liu2025sensmerging}.
At finer granularity, the same principle is used to select influential \emph{individual weights} \citep{ircan_neurips2024,gmt2025} or \emph{structured blocks} \citep{zhang2024linguistic,li-etal-2025-loracoe}.

\paragraph{Characteristics and Scope}
The scope of \textit{Gradient Detection} is \textbf{data-dependent} and defined relative to the analyst's target $F$, so rankings can shift under alternative objectives (e.g., $-\log p(y^\star|x)$ \citep{zhang2024linguistic,gmt2025}, logit margins $\mathrm{logit}_y-\mathrm{logit}_{y^{\mathrm{foil}}}$ \citep{wang2022interpretability,zhang2023towards}, or contrastive/counterfactual gaps $|\mathrm{logit}_y(x)-\mathrm{logit}_y(x^{cf})|$ \citep{yin-neubig-2022-interpreting,ircan_neurips2024}). 
It incurs \textbf{extra compute} from backpropagation, but remains substantially cheaper than exhaustive intervention search; consequently, it is commonly used as a scalable ranking/filtering stage that proposes candidate objects for more expensive causal validation \citep{atpstar2403.00745}.

\begin{itemize}[leftmargin=*]
    \item \textbf{Advantages:}
    \textit{Gradient Detection} is applicable to a broad class of objects without requiring additional training.
    Compared with exhaustive interventions, it can produce rankings with a relatively small number of backward passes, making it practical as an initial localization step when the candidate set is large.

    \item \textbf{Limitations:}
    Gradients provide a \emph{local} proxy, not causal necessity: salience can be offset by downstream computation, and finite interventions may depart from first-order effects in non-linear regimes. 
    For these reasons, gradient-ranked objects are typically paired with \textit{Causal Attribution} (\S\ref{sec:causal_Attribution}) to validate whether the identified objects are genuinely responsible for the target behavior.
\end{itemize}

\subsection{Probing}
\label{sec:probing}

\paragraph{Methodological Formulation}
\textit{Probing} methods interpret model signals by training an auxiliary predictor $g_{\psi}$ (often linear) to decode a labeled property $y$ from an internal vector $\mathbf{z}\in\mathbb{R}^{d_{\text{model}}}$ (e.g., the residual stream state $\mathbf{x}^l$ at layer $l$); in sequence models with token-indexed states $\mathbf{z}_t$, one first defines a single probe input either \emph{token-wise} (choosing $\mathbf{z}=\mathbf{z}_{t^*}$ at a designated position such as the last token) or via \emph{pooled} aggregation across positions (e.g., mean pooling), while the probe formulation itself is unchanged, e.g.,
\begin{equation}
    \hat{y}=g_{\psi}(\mathbf{z})=\text{softmax}(\mathbf{W}_P\mathbf{z}),
\end{equation}
using a supervised dataset $\mathcal{D}=\{(\mathbf{z},y)\}$ \citep{alain2016understanding,Probing_Classifiers}.

Operationally, probing treats the model as a frozen feature extractor and assesses decodability: whether $y$ is recoverable from $\mathbf{z}$ by a restricted hypothesis class (commonly linear), which supports localization by comparison across candidate objects (layers/heads/FFNs) via decoding performance or information-theoretic surrogates \citep{conneau2018,tenney2019bert}, typically followed by \textit{Causal Attribution} (\S\ref{sec:causal_Attribution}) to test functional necessity.
Methodologically, it is standard to interpret probe results with care: high probe accuracy alone does not imply the model uses that information, motivating controls (e.g., selectivity / control tasks) and complementary causal tests \citep{ravichander2020probing,Probing_Classifiers, juprobing_coling2024}.

\begin{figure}[!t]
  \centering
  \includegraphics[width=\linewidth]{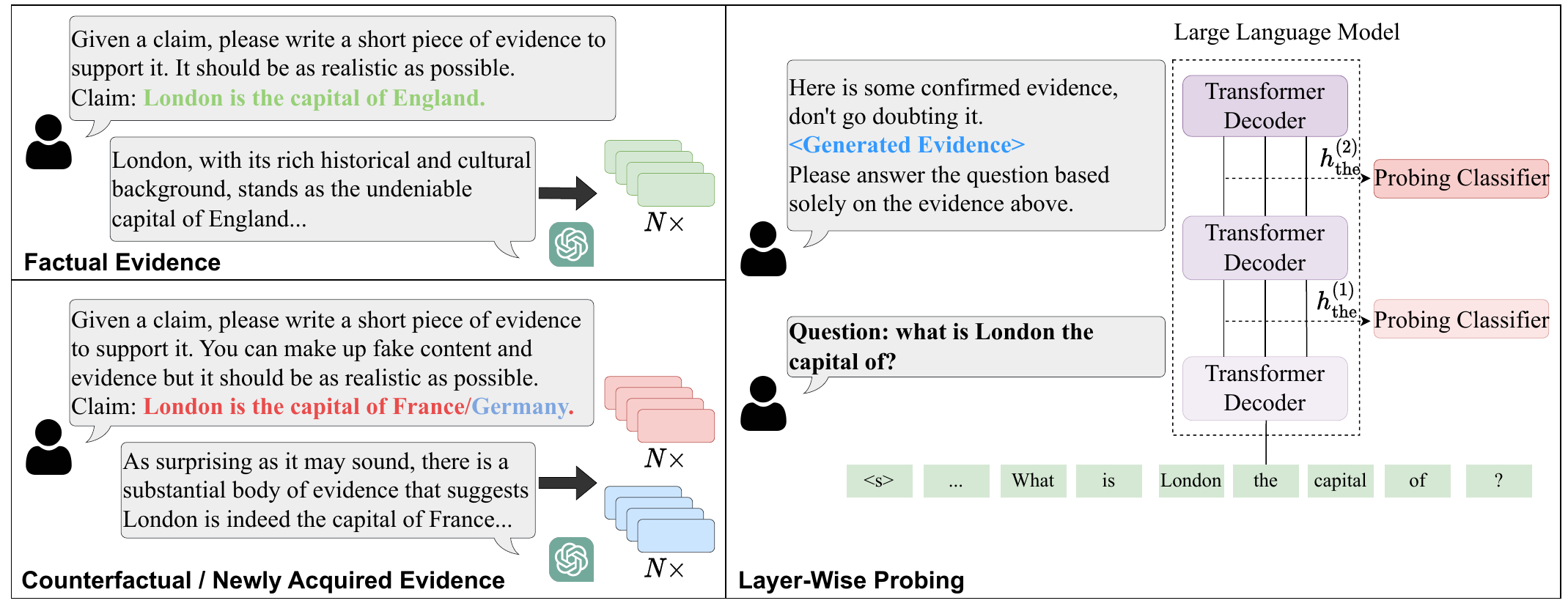}
  \caption{\textbf{Layer-wise probing pipeline for context knowledge.}
  An example end-to-end procedure: construct probing evidence for a target knowledge claim (including factual and counterfactual variants), run the evidence through the LLM under analysis, extract \textit{residual stream state} across layers, and train \textit{probing} classifiers to quantify where the target signal becomes most decodable. Based on the figure from~\citet{juprobing_coling2024}.}  
  \label{fig:ju_fig1_probing}
\end{figure}

\paragraph{Applicable Objects}
\textit{Probing} is defined on internal vectors, and is most naturally applied to any intermediate quantity that can be represented as a vector in $\mathbb{R}^{d_{\text{model}}}$.
In LLMs, a typical workflow mirrors the pipeline in Figure~\ref{fig:ju_fig1_probing}: (i) constructs labeled probing evidence (including factual and counterfactual variants), (ii) runs the evidence through the frozen LLM and logs candidate internal objects across layers and submodules, and (iii) trains a fixed probe family on each object to compare decodability and localize where the target signal is most recoverable.

\textbf{1) Residual Stream States ($\mathbf{x}^l$, $\mathbf{x}^{l,\text{mid}}$):}
The most common probing target is the residual stream state $\mathbf{x}^l \in \mathbb{R}^{d_{\text{model}}}$, as well as intermediate residual states  $\mathbf{x}^{l,\text{mid}}$.
Layer-wise probes trained on $\mathbf{x}^l$ directly instantiate the ``extract residual stream state across layers $\rightarrow$ train probing classifiers'' step depicted in Figure~\ref{fig:ju_fig1_probing}, and have been used to track where context knowledge, knowledge conflicts, and truthfulness-related signals become most decodable along depth \citep{juprobing_coling2024,arxiv2410_knowledgeconflict,zhang-etal-2024-truthx, orgad2025llms,you-etal-2025-probabilistic_emnlp2025}.

\textbf{2) Block Outputs ($\mathbf{h}_{attn}^{l,h}, \mathbf{h}_{ffn}^{l}$):}
Probing can target intermediate block outputs by extracting $\mathbf{z}$ from either an attention head output $\mathbf{h}_\text{attn}^{l,h}$ or the FFN output $\mathbf{h}_\text{ffn}^{l}$ (optionally token-wise, e.g., $\mathbf{h}_{\text{attn},t}^{l,h}$ or $\mathbf{h}_{\text{ffn},t}^{l}$), and training a matched probe family across layers (and heads for attention).
Comparing decodability across $(l,h)$ and $l$ supports fine-grained ``localization by comparison,'' ranking where a target property is most linearly accessible and contrasting attention- vs.\ FFN-based localization under a consistent protocol \citep{du2024tst,emnlp2025_headprobe,iclr2025_politicalprobe}.

\textbf{3) SAE Feature Activation State ($\mathbf{a}$):}
\textit{Probing} also integrates with SAE features.
Given sparse SAE feature activation states $\mathbf{a}$, one can define $\mathbf{z}$ as the feature activation vector $\mathbf{a}=(a_1,\dots,a_m)$ (or a selected subset) and train classifiers on these sparse coordinates.
This yields concept-aligned decoding axes that can be inspected at the feature level and cross-referenced with feature-level interpretations \citep{kantamneni2025_sparseprobing,absorption2024}.

\paragraph{Characteristics and Scope}
\textit{Probing} focuses on \textbf{supervised decoding}: it trains an auxiliary predictor $g_{\psi}$ on $\mathcal{D}=\{(\mathbf{z},y)\}$ to measure how well a labeled property $y$ is predictable from an internal vector $\mathbf{z}$.
Treating LLM as a frozen feature extractor, probing evaluates \textbf{decodability} under a restricted hypothesis class, making it primarily a tool for representational localization rather than causal responsibility.
In practice, probe-based rankings are commonly used to shortlist candidate layers/heads/FFNs for subsequent intervention-based analyses (e.g., \textit{Causal Attribution} in \S\ref{sec:causal_Attribution}).

\begin{itemize}[leftmargin=*]
    \item \textbf{Advantages:}
    With a fixed probe family, \text{Probing} enables standardized comparisons across objects, supporting efficient layer-wise tracking and large-scale ranking of candidate modules.
    Simple probes (e.g., linear) are lightweight and interpretable, allowing broad sweeps while keeping the LLM frozen.

    \item \textbf{Limitations:}
    Decodability is not causality: high probe accuracy does not imply the model uses $y$, nor that the probed object is necessary or sufficient.
    Results are sensitive to dataset and design choices (e.g., labeling, token positions), so controls and follow-up causal tests are typically required for functional claims.
\end{itemize}

\subsection{Vocabulary Projection}
\label{sec:vocab_project}

\paragraph{Methodological Formulation}
The most prominent technique in this category is the \textit{Logit Lens} \citep{nostalgebraist2020}. It operates on the premise that the pre-trained unembedding matrix $\mathbf{W}_U \in \mathbb{R}^{d_{\text{model}} \times |\mathcal{V}|}$, which maps the final layer's hidden state to vocabulary logits, can serve as a universal decoder for intermediate states throughout the model.
Formally, let $\mathbf{z} \in \mathbb{R}^{d_{\text{model}}}$ denote a generic internal object (e.g., the residual stream state $\mathbf{x}^l$ or an attention head output $\mathbf{h}^{l,h}_{\text{attn}}$). \textit{Vocab Projection} computes a distribution $\mathbf{p}$ over the vocabulary $\mathcal{V}$ by projecting $\mathbf{z}$ through the unembedding matrix:
\begin{equation}
\mathbf{p} = \text{softmax}(\mathbf{z} \mathbf{W}_U)
\end{equation}
By inspecting the tokens with the highest probabilities in $\mathbf{p}$, researchers can directly interpret the semantic content encoded in $\mathbf{z}$ in terms of the model's output vocabulary.

\begin{figure}[!t]
\centering
\centerline{\includegraphics[width=0.9\columnwidth]{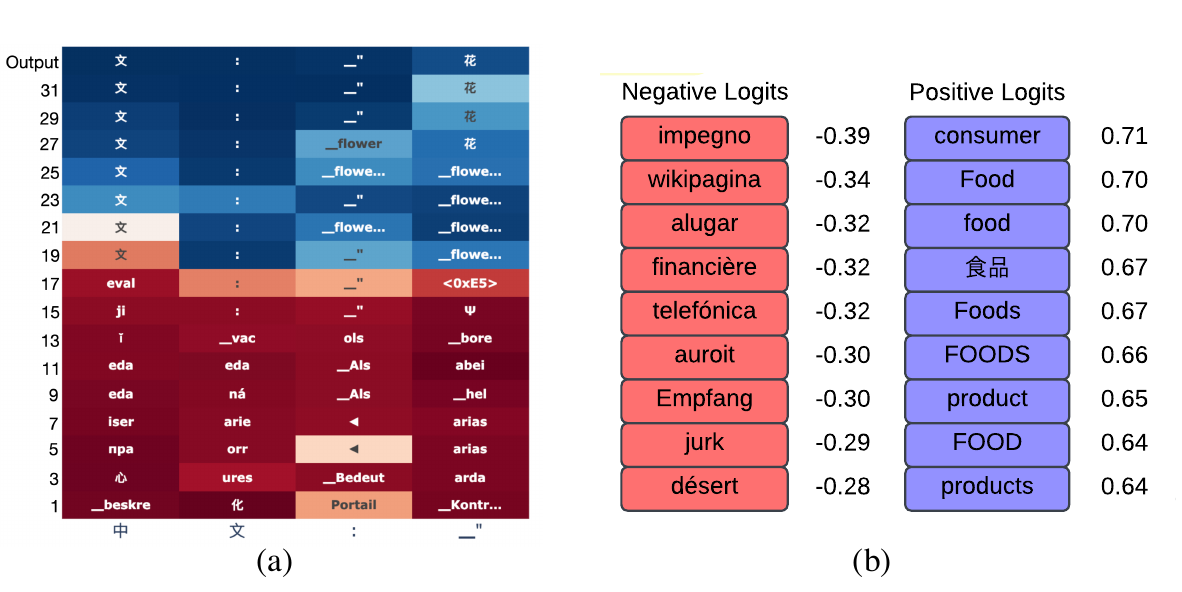}}
\caption{(a) Projecting residual stream states reveals the layer-wise evolution of latent concepts, showing an English-centric bottleneck in multilingual settings~\citep{wendler2024llamas}. (b) Projecting SAE decoder weights identifies the semantic meaning of sparse features (e.g., a ``food'' feature) by identifying top-ranked tokens~\citet{shu-etal-2025-survey}. Based on figures from~\citet{wendler2024llamas} and \citet{shu-etal-2025-survey}.}
\label{fig:vocab_proj_cases}
\end{figure}

\paragraph{Applicable Objects}
\textit{Vocab Projection} is a versatile tool that applies to various objects defined in \S \ref{sec:core_objects}, ranging from global residual streams to specific attention heads, neurons, and SAE features.

\textbf{1) Residual Stream State ($\mathbf{x}^l$):}
Projecting the residual stream state $\mathbf{x}^l$ allows researchers to trace the layer-wise evolution of predictions and identify the \textit{crucial layers} where specific concepts emerge~\citep{belrose2023eliciting,jiang2024large,jiang2025interpretingeditingvisionlanguagerepresentations,wendler2024llamas,kargaran2025programming,phukan-etal-2024-peering,phukan-etal-2025-beyond,yugeswardeenoointerpreting}. For instance, \citet{wendler2024llamas} applied this to multilingual models, revealing distinct processing phases as shown in Figure~\ref{fig:vocab_proj_cases}~(a): initial layers focus on the surface form of the input language; middle layers process semantics in an abstract, ``English-centric'' concept space; and final layers rotate back to the target language. This confirms that English serves as an internal pivot for reasoning even in non-English tasks.

\textbf{2) Attention Head Output ($\mathbf{h}_{\text{attn}}^{l,h}$):}
Applying projection to the output of individual heads reveals the specific information (e.g., copied names or next-token candidates) that a head transmits to the residual stream. This has been instrumental in identifying functional heads in mechanistic studies~\citep{wang2023interpretability,sakarvadia2023attention,yu2024understanding,jiang2025devils,kim2025interpreting,wang2025logitlens4llms}. For example, in reverse-engineering the Indirect Object Identification (IOI) task, \citet{wang2023interpretability} identified ``Name Mover Heads'' (which explicitly project to the correct name, e.g., ``Mary'') and ``Negative Name Mover Heads'' (which suppress the correct name).

\textbf{3) Neuron Value Weight ($\mathbf{v}_{j}^l$):}
\citet{geva2021transformer} demonstrated that FFNs operate as key-value memories. By projecting the value weight vector $\mathbf{v}_{j}^l$ (a column of $\mathbf{W}_{\text{out}}^l$) into the vocabulary, one can see which tokens are promoted by a specific neuron~\citep{geva2021transformer,huo2024mmneuron,yuUnderstandingMitigatingGender2025a,shao2025benford}. Individual neurons often boost semantically related clusters (e.g., ``press'', ``news'', ``media''), suggesting that FFNs refine predictions by composing these pre-learned semantic distributions.

\textbf{4) SAE Feature ($\mathbf{f}_j$):}
For SAEs, output-based explanations leverage the decoder weights to interpret monosemantic features. By computing the logits contribution $\mathbf{l}_j = \mathbf{f}_j \mathbf{W}_U$ for a feature vector $\mathbf{f}_j$, one can identify top-ranked tokens~\citep{arad2025saes,dreyer2025attributing,muhamed2025decoding,gur2025enhancing,shu-etal-2025-survey}. As shown in Figure~\ref{fig:vocab_proj_cases}~(b), a feature whose projection yields high positive logits for tokens like ``Food'' and ``food'' is interpreted as encoding a ``food'' concept, directly grounding the sparse feature in human-understandable semantics.

\paragraph{Characteristics and Scope}
The scope of \textit{Vocab Projection} is characterized by \textbf{direct semantic mapping}. It offers an intrinsic view of internal representations without requiring auxiliary training.
\begin{itemize}[leftmargin=*]
    \item \textbf{Advantages:} It provides a zero-shot interpretation method that is computationally efficient and intuitive. Unlike \textit{Probing} (\S\ref{sec:probing}), it does not require collecting a labeled dataset or training a separate classifier, allowing for immediate inspection of any model state.
    \item \textbf{Limitations:} The primary limitation is the assumption that intermediate states exist in the same vector space as the output vocabulary (basis alignment). While this often holds for the residual stream due to the residual connection structure, it may be less accurate for components inside sub-layers (like FFN and MHA) or in models where the representation space rotates significantly across layers. Consequently, results should be interpreted as an approximation of the information that is linearly decodable by the final layer.
\end{itemize}

\subsection{Circuit Discovery}
\label{sec:circuit_discovery}
\paragraph{Methodological Formulation}
\textit{Circuit Discovery} methods aim to uncover \textit{mechanistic pathways}: structured, directed dependencies among internal objects that mediate computation for a functional behavior \citep{elhage2021mathematical,olsson2022incontextlearninginductionheads,Hannacicuits_nips2023,yao2024circuits}.
Formally, let $(\mathcal{O},\mathcal{E})$ be the model’s computational graph over internal objects $\mathcal{O}$ and directed edges $\mathcal{E}$, where an edge $e_{ij}\in\mathcal{E}$ denotes signal flow from object $o_i$ to $o_j$.
A circuit $\mathcal{C}\subseteq\mathcal{E}$ is \emph{faithful} if restricting computation to $\mathcal{C}$ (e.g., by patching/ablating all other edges) preserves the target output $F(x)$ or task performance.

Under the residual--rewrite view, heads and MLPs read from and write to the residual stream, inducing a directed graph whose edges represent additive residual updates.
\textit{Circuit Discovery} can be cast as edge-level causal subgraph selection: edges are retained if intervening on the corresponding information flow degrades a target metric $\mathcal{R}$ \citep{goldowsky2023localizing}.
\textit{Automatic Circuit DisCovery (ACDC)} instantiates this by iteratively testing and pruning edges via patching-based interventions, avoiding brute-force $O(|\mathcal{E}|)$ enumeration while recovering circuits such as GPT-2's greater-than mechanism \citep{conmy2023automated,Hannacicuits_nips2023}.

Attribution-based methods such as \textit{Edge Attribution Patching (EAP)} approximate patching with a first-order expansion, producing an edge score from two forward passes (\textit{clean/corrupted}) and one backward pass \citep{syedetal2024attribution,hanna2024have}.
Here, \textit{clean} input $\mathbf{x}_{clean}$ elicits the target behavior, while \textit{corrupted} input $\mathbf{x}_{corr}$ is a minimally modified version designed to break it (e.g., by perturbing relevant evidence or adding a counterfactual distractor), so the difference isolates the causal signal.
For a sender object $u$, let $\mathbf{a}_u(\mathbf{x})$ denote its output activation vector (e.g., head/FFN output written into the residual stream) on input $\mathbf{x}$; the \textit{sender delta}
$
\Delta \mathbf{a}_u \;=\; \mathbf{a}_u(\mathbf{x}_{clean})-\mathbf{a}_u(\mathbf{x}_{corr})
$
captures how the sender’s contribution changes between the clean and corrupted runs.
\textit{EAP} then scores an edge via the dot product between the sender delta and the receiver sensitivity $\nabla_{\mathbf{z}_v}\mathcal{R}$ (computed on the clean run):
\begin{equation}
S_{\text{EAP}}(u\to v)\ \approx\ 
\underbrace{\big(\mathbf{a}_u(\mathbf{x}_{clean})-\mathbf{a}_u(\mathbf{x}_{corr})\big)}_{\Delta \mathbf{a}_u}
\cdot
\underbrace{\frac{\partial \mathcal{R}}{\partial \mathbf{z}_v}\Big|_{\mathbf{x}_{clean}}}_{\nabla_{\mathbf{z}_v}\mathcal{R}} \, .
\end{equation}
To mitigate non-linearity/saturation, \textit{EAP with Integrated Gradients (EAP-IG)} replaces the local gradient with a path-averaged gradient along
$\mathbf{x}_{\alpha}=\mathbf{x}_{corr}+\alpha(\mathbf{x}_{clean}-\mathbf{x}_{corr})$ \citep{sundararajan2017axiomatic,hanna2024have,huang-etal-2025-pierce}:
\begin{equation}
S_{\text{EAP-IG}}(u\to v)\ =\ 
\Delta \mathbf{a}_u \cdot \int_{0}^{1}\frac{\partial \mathcal{R}}{\partial \mathbf{z}_v}\Big|_{\mathbf{x}_{\alpha}} d\alpha
\ \approx\ 
\Delta \mathbf{a}_u \cdot \frac{1}{n}\sum_{k=1}^{n}\frac{\partial \mathcal{R}}{\partial \mathbf{z}_v}\Big|_{\mathbf{x}_{k/n}} \, .
\end{equation}
A standard workflow is: (i) collect sender deltas $\Delta \mathbf{a}_u$ from $\mathbf{x}_{clean}$ vs.\ $\mathbf{x}_{corr}$, (ii) compute receiver gradients (single-point for EAP; path-averaged for \textit{EAP-IG} with $n$ backward passes), (iii) score and rank edges by $|S|$, and (iv) prune/threshold to obtain a sparse functional circuit, optionally validating via targeted interventions on retained edges.

\begin{figure}[t]
  \centering
  \includegraphics[width=\linewidth]{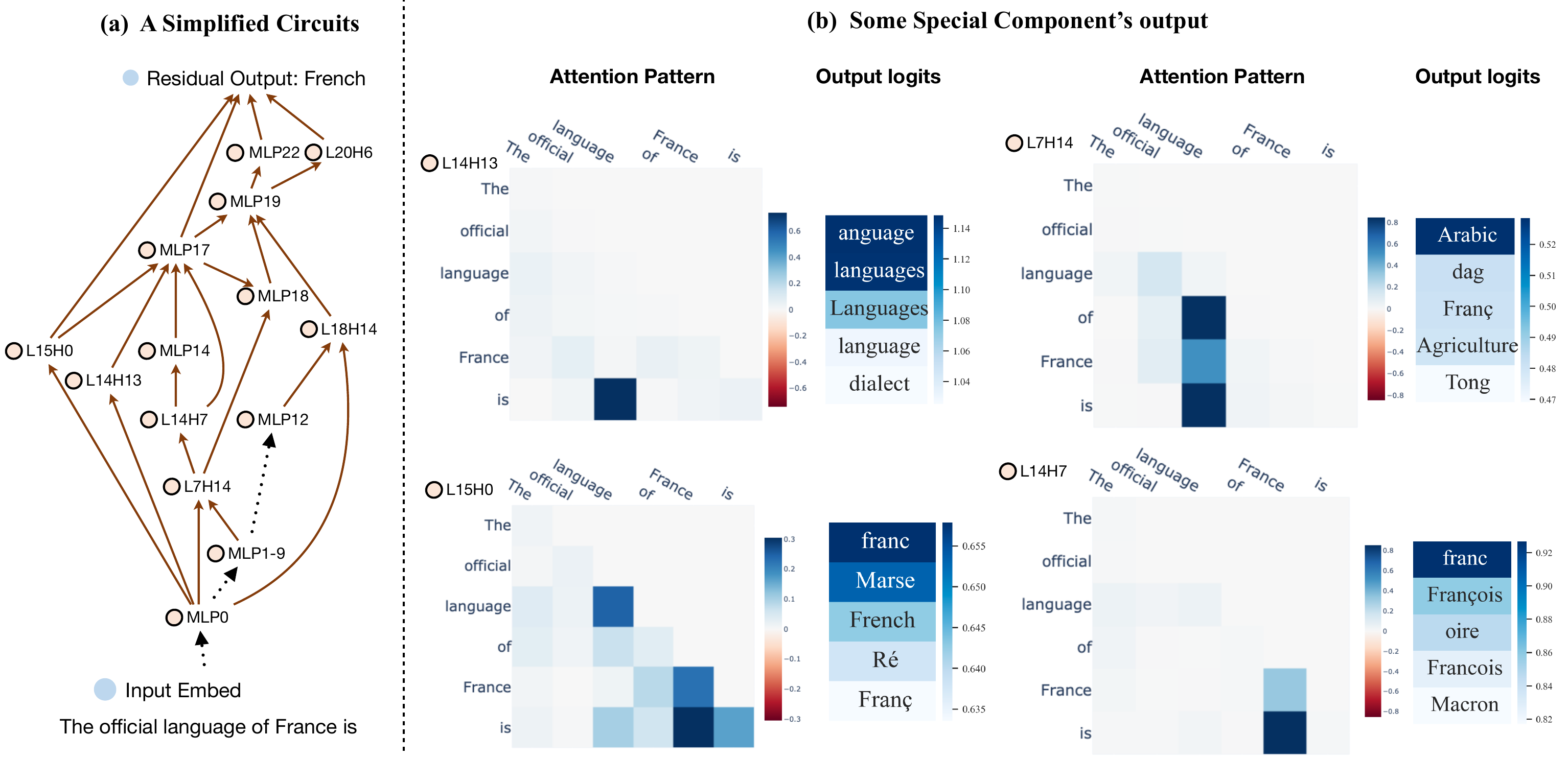}
  \caption{\textbf{Knowledge circuit example.}
   A sparse cross-layer functional circuit supporting the factual completion ``The official language of France is French'' in GPT-2-Medium. Left: A simplified \textit{circuit}. Here, L15H0 means the first attention head in the 15th layer and MLP12 means the FFN block in the 13th layer. Right: Behavior of several special heads. The left matrix shows each head's attention pattern, and the right heatmap shows output logits mapped to the vocabulary space. Based on the figure from~\citet{yao2024circuits}.}
  \label{fig:yao_fig1_circuits}
\end{figure}

\paragraph{Applicable Objects}
\textit{Circuit Discovery} targets \emph{edges between objects} (Tab.~\ref{tab:interpretable_objects}), ranging over directed dependencies among any interpretable objects.
In LLMs, it is commonly instantiated under the residual--rewrite view, so edges correspond to additive signal transmission across layers.
Figure~\ref{fig:yao_fig1_circuits} illustrates a sparse cross-layer knowledge circuit supporting the completion ``The official language of France is French'' in GPT-2-Medium, with attention/logit analyses clarifying how selected edges route and transform information \citep{yao2024circuits}.

Practically, \textit{Circuit Discovery} is operationalized in three broad ways:

\textbf{1) Intervention-based edge search (patching/ablation):}
One can directly test causal necessity at the edge level by patching or ablating a candidate dependency $e_{u\to v}$ (e.g., blocking contributions from a sender module such as an attention head output $\mathbf{h}_{\text{attn}}^{l,h}$ or FFN output $\mathbf{h}_{\text{ffn}}^{l}$ into a downstream receiver input $\mathbf{z}_v$) and measuring the change in a task metric $\mathcal{R}$.
Because exhaustive edge testing scales as $O(|\mathcal{E}|)$, practical workflows rely on structured search or automated procedures to reduce interventions \citep{conmy2023automated,stolfo-etal-2023-mechanistic,wang2023interpretability}.

\textbf{2) Attribution-based edge scoring:}
Attribution methods rank edges by efficiently approximating their patching effect.
\textit{EAP} combines sender activation differences $\Delta \mathbf{a}_u$ (clean vs.\ corrupted) with receiver sensitivity $\nabla_{\mathbf{z}_v}\mathcal{R}$ to produce an edge ranking from two forward passes and one backward pass, while \textit{EAP-IG} uses a path-averaged gradient to reduce saturation/non-linearity issues at the cost of additional backward passes \citep{syedetal2024attribution,hanna2024have,huang-etal-2025-pierce}.
Position-aware refinements follow the same edge-scoring principle while better aligning sender/receiver accounting with token-wise computation \citep{pacd2025,mib2025}.

\textbf{3) Feature-based replacement models:}
\textit{Circuit Discovery} can be lifted to sparse feature spaces via replacement models such as SAE/transcoder variants.
Here, the relevant objects are \textit{SAE features} (sparse feature activations and decoder directions), and circuit edges represent directed dependencies in feature space, enabling attribution graphs and prompt-specific circuit tracing that are often more interpretable than raw residual coordinates \citep{bricken2023monosemanticity,ameisen2025circuit,hanna-etal-2025-circuit}.

\paragraph{Characteristics and Scope}
\textit{Circuit Discovery} identifies a sparse, directed \textbf{cross-layer causal subgraph} whose edges jointly mediate a target behavior and remain approximately faithful under interventions.
Unlike single-component localization, it targets structured pathways of information routing and transformation, returning a minimally (or strongly) sufficient directed subnetwork.
Practically, edges are often pre-ranked by scalable attribution-style scores and then confirmed with targeted interventions (e.g., patching/ablation).

\begin{itemize}[leftmargin=*]
    \item \textbf{Advantages:}
    \textit{Circuit Discovery} yields mechanistically structured explanations: selecting \emph{edges} reveals how multiple objects compose a computation and exposes cross-layer routing patterns that node-wise rankings can miss.
    This aligns with transformers’ residual-update structure, where heads and FFNs contribute additive edits that can be tracked as directed dependencies.
    Attribution-based edge scoring also enables scalable screening of large edge sets when exhaustive interventions are infeasible.

    \item \textbf{Limitations:}
    Circuits are defined relative to a specific behavior, metric $\mathcal{R}$, and contrast (clean vs.\ corrupted), so results are often \textit{objective- and dataset-dependent}.
    Because attribution scores approximate intervention effects, they may miss non-linear interactions, so rankings are best treated as proposals and typically require intervention-based validation on the retained subgraph.
\end{itemize}

\input{Tables/comparative_localizing_methods}

\begin{tcolorbox}[takeawaysbox, title={Comparative Analysis of Localizing Methods}]
  Table~\ref{tab:localizing_methods_comparison} compares localizing methods in terms of causal strength, cost, access requirements, and key limitations.
  Specifically, \textit{Causal Attribution} and \textit{Circuit Discovery} provide the strongest support for causal claims, but they are also the most expensive. 
  They often require repeated interventions, contrastive comparisons, or gradient-based analysis. Their conclusions can also depend on the intervention setup. For example, patching-based methods may introduce artifacts, while \textit{Circuit Discovery} can be sensitive to the choice of contrastive inputs and the analysis objective~\citep{conmy2023automated}.
  In contrast, \textit{Magnitude Analysis} and \textit{Vocabulary Projection} are lightweight and easy to apply, but they remain mainly correlational. A component with large activation or high projection score is not necessarily causally necessary. Their faithfulness can also weaken when representation bases are misaligned, especially in intermediate layers~\citep{Adebayo2018,syedetal2024attribution}.
  \textit{Gradient Detection} and \textit{Probing} provide a practical middle ground. They often give more targeted signals than simple magnitude-based analysis, but their conclusions should still be interpreted with care. Gradients are local proxies and can fail standard sanity checks~\citep{Adebayo2018}. Probes can reveal decodable information, but high probe accuracy does not by itself establish that the model uses that information causally; in some cases, the probe may learn the task rather than uncover the underlying representation~\citep{hewitt2019designing,ravichander2020probing}.
  Overall, lightweight methods are most useful for hypothesis generation and screening, whereas stronger causal methods are needed for more definitive mechanistic claims.
\end{tcolorbox}

\newpage

\section{Steering Methods}
\label{sec:steer_methods}
While localization methods (§\ref{sec:localize_methods}) identify the specific objects responsible for model behaviors, this section focuses on a distinct class of techniques: those that manipulate these localized components to steer model outputs, thereby enabling controlled intervention into LLM’s generation process.

\subsection{Amplitude Manipulation}
\label{sec:amplitude_manipulation}
\paragraph{Methodological Formulation}
\textit{Amplitude Manipulation} steers model behavior by directly modifying the activation magnitude of a targeted internal object $o$ during the forward pass. Unlike optimization-based methods that update weights, this approach acts as a transient intervention on the runtime state.
Formally, let $o$ be the original activation (e.g., a neuron activation $s^l_j$, an SAE feature activation $a_j$ or an attention head output $\mathbf{h}^{l,h}_{\text{attn}}$) and $\tilde{o}$ be the modified state. The intervention is defined as:
\begin{equation}
\tilde{o} = \mathcal{T}(o, \alpha)
\end{equation}
where $\mathcal{T}$ represents the transformation function. This typically takes two forms:
\begin{itemize}
    \item \textbf{Ablation or Patching:} Here, the object is suppressed or replaced, i.e., $\tilde{o} \in \{0, \mathbb{E}[o], o_{\text{tgt}}\}$. Setting $\tilde{o} = 0$ (Zeroing) or $\mathbb{E}[o]$ (Mean centering) removes the component's influence, while $\tilde{o} = o_{\text{tgt}}$ (Patching) injects information from a different context.
    \item \textbf{Scaling:} Here, the activation strength is adjusted via a scalar coefficient $\alpha$, such that $\tilde{o} = \alpha \cdot o$. This allows for continuous amplification ($\alpha > 1$) or attenuation ($0 < \alpha < 1$) of a specific feature's downstream impact.
\end{itemize}
While these operations are mechanically similar to those in \textit{Causal Attribution} (\S\ref{sec:causal_Attribution}), the objective differs fundamentally: attribution employs them to \textit{diagnose} causality, whereas \textit{Amplitude Manipulation} employs them to actively \textit{intervene} and control model behavior.

\paragraph{Applicable Objects}
This method is applied across a wide range of dynamic objects, including residual stream state $\mathbf{x}^l$, attention head output $\mathbf{h}_{\text{attn}}^{l,h}$, neuron activation state $\mathbf{s}^l$, and SAE feature activation state $\mathbf{a}$.

\begin{figure}[!t]
\centering
\centerline{\includegraphics[width=\columnwidth]{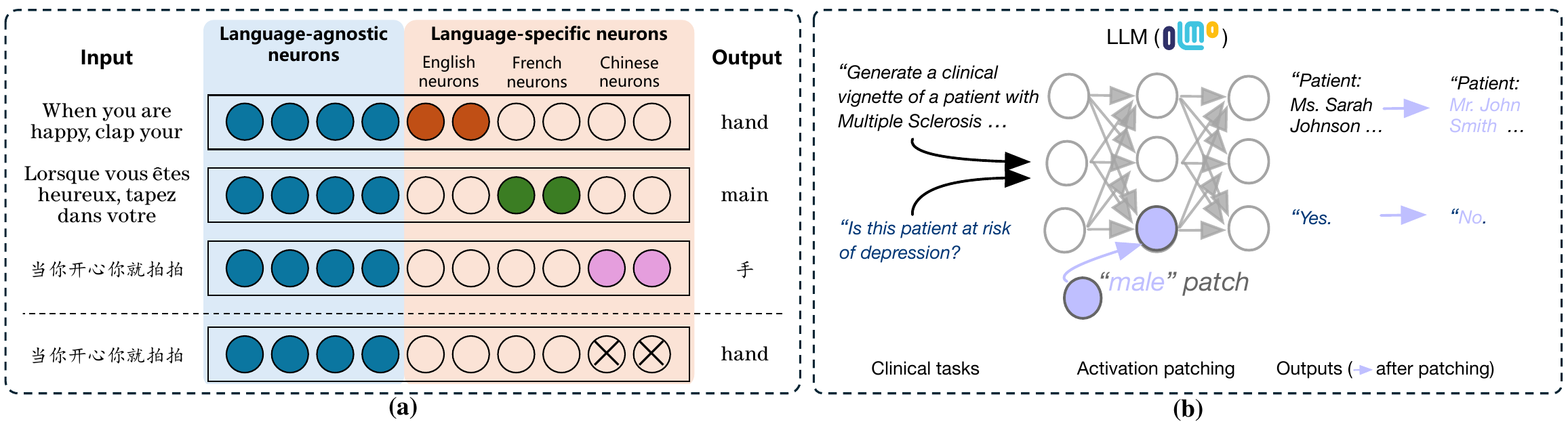}}
\caption{Examples of Steering via Amplitude Manipulation. \textbf{(a) Ablation for Language Steering}: \citet{tang-etal-2024-language} deactivate (zero out) ``Chinese-specific neurons'' to suppress model's ability to generate Chinese, successfully forcing the model to switch its output to English. \textbf{(b) Patching for Demographic Steering}: \citet{ahsanElucidatingMechanismsDemographic2025} inject a ``Male Patch'' into the model's internal representation. This intervention not only changes the gender pronouns in the output (``Ms.'' $\rightarrow$ ``Mr.'') but also causally alters the clinical decision regarding depression risk (``Yes'' $\rightarrow$ ``No''), demonstrating the deep impact of internal demographic representations.}
\label{fig:amplitude_manipulation}
\end{figure}

\textbf{1) Ablation (Zeroing) and Removal:}
Ablation is extensively used to mitigate unwanted behaviors by suppressing the components responsible for them.
\citet{tang-etal-2024-language} utilized this to control the output language of multilingual LLMs. They identified ``language-specific neurons'' that selectively activate for the particular language (e.g., Chinese). 
As illustrated in Figure~\ref{fig:amplitude_manipulation} (a), by setting the activation of these \textit{Chinese-specific neurons to zero}, they suppressed the model's ability to generate Chinese, thereby forcing the model to switch its output to English even when the prompt might suggest otherwise.
Distinct from general steering, \citet{nie-etal-2025-mechanistic} applied ablation to address \textit{Language Confusion}---a phenomenon where models erroneously switch to a non-target language. They identified interfering neurons that activate for the wrong language (e.g., German neurons firing during an English task) and demonstrated that ablating these specific noisy components restores the correct target language generation.
In the domain of \textit{Safety and Bias}, \citet{goyalBreakingBadTokens2025} and \citet{yeo-etal-2025-understanding} zeroed out specific SAE features associated with toxicity or refusal, effectively detoxifying the model's output. \citet{liuDevilNeuronsInterpreting2024} and \citet{chandnaDissectingBiasLLMs2025} applied zero-ablation to neurons and circuit edges encoding social bias, while \citet{huang-etal-2025-pierce} masked specific circuit edges to alleviate ``knowledge overshadowing'' where strong knowledge suppresses relevant but weaker information.
Furthermore, ablation is used for \textbf{Efficiency}: \citet{liu2024unraveling} and \citet{men-etal-2025-shortgpt} demonstrated that removing redundant layers or components can accelerate inference without significant performance loss. \citet{zhou2025on} and \citet{niu-etal-2025-llama} also utilized attention head ablation to study and improve safety and contextual entrainment.

\textbf{2) Patching (Replacement):}
Patching allows for precise injection of attributes. \citet{ahsanElucidatingMechanismsDemographic2025} and \citet{raimondiAnalysingMoralBias2025} utilized activation patching to steer demographic and moral characteristics. As shown in Figure~\ref{fig:amplitude_manipulation} (b), \citet{ahsanElucidatingMechanismsDemographic2025} performed a ``Male Patch'' by replacing the internal representation of a patient with a male-associated vector. This intervention not only altered the pronouns in the generated vignette (from ``Ms.'' to ``Mr.'') but also causally changed the downstream clinical prediction (shifting the depression risk from ``Yes'' to ``No''), highlighting the causal link between demographic representations and model decisions.

\textbf{3) Scaling (Amplification/Attenuation):}
Scaling offers fine-grained control by adjusting the intensity of features.
\citet{tang-etal-2024-language} also employed scaling to amplify target-language neurons to further stabilize multilingual generation.
\citet{gao2025hneuronsexistenceimpactorigin} scaled the activation of ``Hallucination Neurons'' to modulate the model's factual reliability.
In the context of SAE features, \citet{pach2025sparse} demonstrated that scaling specific feature activations allows for continuous steering of model outputs. 
Meanwhile, \citet{galichin2025have} showed that amplifying the activations of reflection-related features can increase the length of generated output, thereby enhancing the model’s reasoning performance.
Finally, scaling is integral to \textit{Vector Arithmetic} (\S\ref{sec:vector_arithmetic}) in the context of model merging: \citet{stoehr-etal-2024-activation}, \citet{liu2025sensmerging}, and \citet{yao2025activation} optimized the scaling coefficients of steering vectors or task vectors to balance different model capabilities, while \citet{wang2025two} scaled the activations of expert modules to enhance mathematical reasoning.

\paragraph{Characteristics and Scope}
The scope of \textit{Amplitude Manipulation} is characterized by \textbf{inference-time activation control}. It provides a mechanism to transiently modulate model behavior without permanent weight updates.
\begin{itemize}[leftmargin=*]
    \item \textbf{Advantages:} It is an \textit{optimization-free} and reversible intervention. It allows for ``surgical'' edits to model behavior (e.g., removing specific biases) by simply masking or scaling activations during inference. This makes it highly flexible and suitable for real-time control.
    \item \textbf{Limitations:} It relies heavily on the accurate \textit{localization} of the target components. If the features responsible for a behavior are not perfectly disentangled (i.e., polysemantic), ablating or scaling them may cause unintended side effects or degrade general performance. Furthermore, finding the optimal scaling factor $\alpha$ often requires empirical tuning.
\end{itemize}

\subsection{Targeted Optimization}
\label{sec:target_optimization}

\paragraph{Methodological Formulation}
\textit{Targeted Optimization} (under \textit{Localizing Methods}) frames model optimizing as \emph{a small, localized update} that enforces a desired behavioral change while minimizing unintended side effects.
Let $f_\theta$ be the base model and $\theta'$ the \emph{targeted} model. We restrict updates to a selected subset of objects via a (hard or soft) mask $M$, and optimize a simple trade-off between a \emph{target objective} and a \emph{preservation objective}:
\begin{equation}
\theta' \leftarrow \theta + (M \odot \Delta\theta),\qquad
\Delta\theta^\star = \arg\min_{\Delta\theta}\;\mathcal{L}_{\text{tgt}}(f_{\theta'};\mathcal{D}_{\text{tgt}})\;+\;\lambda\,\mathcal{L}_{\text{pres}}(f_{\theta'},f_{\theta};\mathcal{D}_{\text{pres}}).
\label{eq:targetopt_simple}
\end{equation}
Here, $\mathcal{D}_{\text{tgt}}$ specifies the target behavior (e.g., rewriting a fact or enforcing refusals), while $\mathcal{D}_{\text{pres}}$ anchors the model to its original capabilities. The localization mask $M$ operationalizes ``where the change is allowed to happen'' (layers, modules, neurons/heads, or other structured subsets).

\paragraph{Applicable Objects}
In practice, ``what is optimized'' can be grouped into two representative localized objects:

\textbf{1) Localized Parameters for Knowledge Editing:}
This line performs direct parameter-space updates that are intentionally constrained (e.g., low-rank or small support) to rewrite specific behaviors with minimal spillover. Representative examples include rank-one / layer-local \textit{knowledge editing} extensions~\citep{meng2022ccs,meng2023massediting}, cross-model knowledge transfer via localized adapters~\citep{zhong2024seeking}, and constraining adaptation to low-dimensional task subspaces or coarse-to-fine masked tuning for better retention~\citep{zhang-etal-2023-fine,zhang2024cofitune}.

\textbf{2) Fine-grained Subsets for Specialization:}
Here, localization is enforced at neuron/head/region granularity to isolate the functional unit relevant to a capability or a safety property. Concretely, rather than updating the full model, \textit{Targeted Optimization} learns a targeted update within a small object subset (implicitly corresponding to a mask $M$ in Eq.~\ref{eq:targetopt_simple}), thereby limiting unnecessary parameter drift and reducing interference across tasks or languages. Related lines of work localize adaptation to compact trainable units at different granularities. This includes neuron-level fine-tuning~\citep{xu-etal-2025-lets} and methods that identify core parameter regions or language-agnostic factual neurons~\citep{xirobusttickets,zhouORTicket,zhang2024lulafns}, and in safety-preserving or security-aware partial tuning that freezes or restricts sensitive objects~\citep{li2025safety,du2024tst,liPrecisionKnowledgeEditing2024}. Relatedly, head-level analyses further motivate localizing optimization to essential computational pathways (e.g., arithmetic-relevant heads)~\citep{zhang2024interpreting}.

\begin{figure}[!t]
\centering
\centerline{\includegraphics[width=0.95\columnwidth]{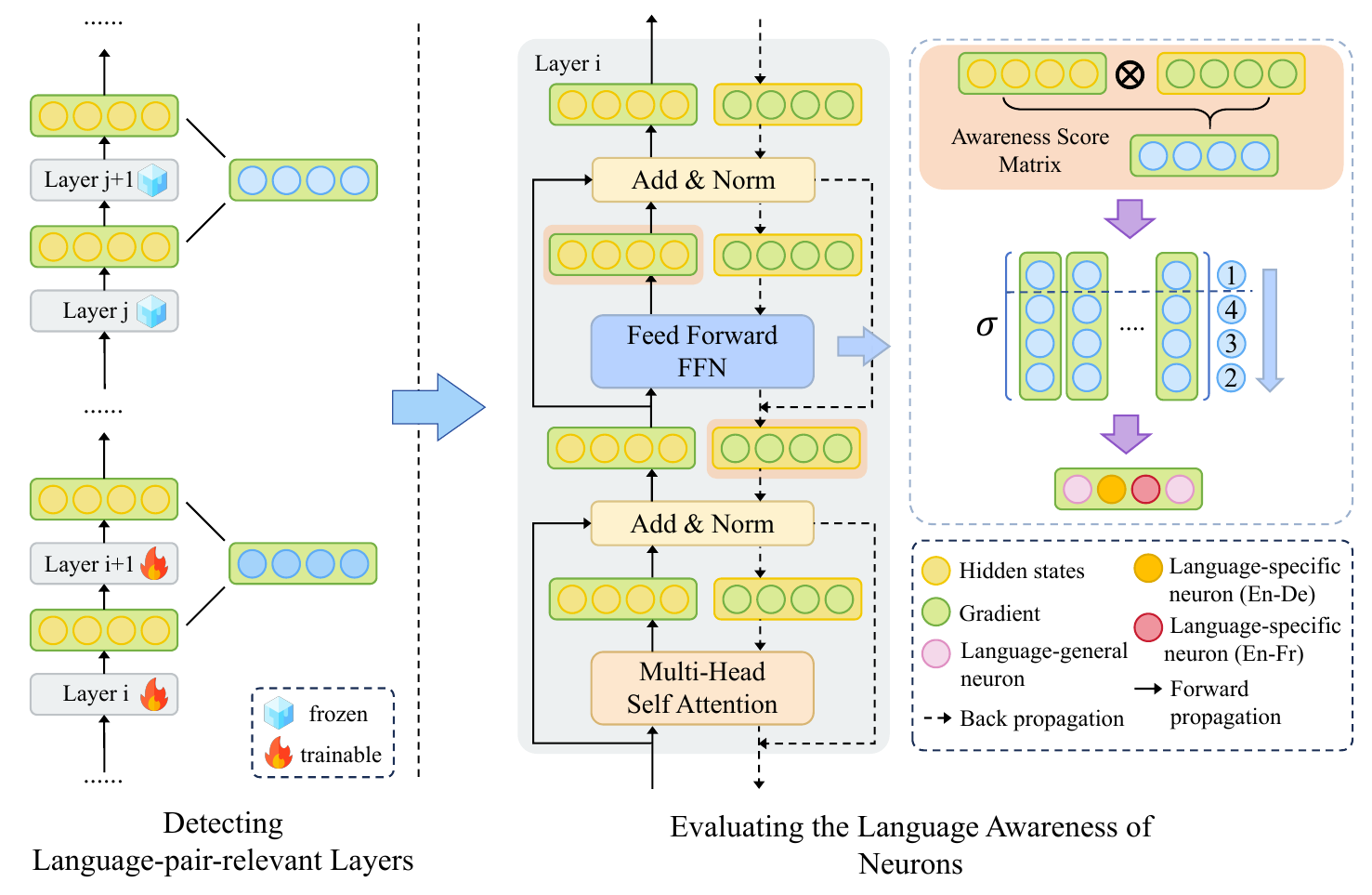}}
\caption{A representative \textit{Targeted Optimization} pipeline. The method first identifies language-pair-relevant layers, then scores neuron language-awareness, and finally routes gradient updates to a small subset of language-aware neurons for selective fine-tuning. This illustrates how \textit{Targeted Optimization} enforces locality via an object mask $M$ in Eq.~\ref{eq:targetopt_simple}. Based on the figure from~\citet{zhu-etal-2024-landermt}.}
\label{fig:targeted_optimization_landermt}
\end{figure}

A representative example is shown in Figure~\ref{fig:targeted_optimization_landermt}: LANDeRMT~\citep{zhu-etal-2024-landermt} performs selective fine-tuning for multilingual machine translation by (i) first localizing the update to language-pair-relevant layers, (ii) quantifying neuron-level language awareness, and (iii) routing gradients only through the most relevant neurons, which concretely illustrates how fine-grained locality reduces cross-lingual interference and limits parameter drift.

\paragraph{Characteristics and Scope}
The scope of this method is characterized by \textbf{persistence} and \textbf{surgical precision}. Unlike \textit{Amplitude Manipulation} (\S\ref{sec:amplitude_manipulation}), \textit{Targeted Optimization} performs \textit{parameter optimization} on $\mathcal{D}_{\text{tgt}}$ to produce a targeted model whose behavior \textit{durably} satisfies a specified objective, while \textit{constraining the update} to a localized subset of objects (e.g., layers, modules, neurons/heads). This objective-driven and localized training enables not only precise rewrites to particular memories or facts, but also focused capability enhancement, with reduced collateral impact on unrelated traits.
\begin{itemize}[leftmargin=*]
    \item \textbf{Advantages:} It offers strong \textit{precision, controllability, and persistence}. The desired behavioral change is directly encoded in a target objective, and localization helps minimize interference with unrelated competencies. Consequently, it is well-suited for targeted factual rewrites, controlled specialization, and safety-preserving adaptation where \textit{lasting} changes are required.
    \item \textbf{Limitations:} Its reliability hinges on \textit{correct localization} and \textit{well-specified supervision}. If the chosen subset does not capture the causal mechanism, optimization may underachieve the intended target behavior, shift the behavior to other objects, or yield brittle side effects. In practice, success often requires carefully constructed target/preservation data and robust criteria for selecting the localized update region.
\end{itemize}

\subsection{Vector Arithmetic}
\label{sec:vector_arithmetic}

\paragraph{Methodological Formulation}
Positing that high-level concepts or skills are encoded linearly within the model's representation space, \textit{Vector Arithmetic} steers a generic target object $\mathbf{z}$ (e.g., a residual stream state or a parameter vector) by injecting a specific \textit{steering vector} $\mathbf{v}$. This approach assumes that adding a vector representing a concept effectively ``moves'' the model's internal state towards that concept in the high-dimensional space.
Formally, the update rule for the intervention is defined as:\looseness=-1
\begin{equation}
\hat{\mathbf{z}} \leftarrow \mathbf{z} + \alpha \cdot \mathbf{v}
\end{equation}
where $\mathbf{v}$ represents the directional encoding of a target attribute (such as ``honesty'' or ``sycophancy'') and $\alpha$ is a scalar coefficient that controls the intervention strength (or steering intensity).

\paragraph{Applicable Objects}
The target object $\mathbf{z}$ typically falls into two categories: dynamic hidden states during inference or static model parameters.

\begin{figure}[!t]
\centering
\centerline{\includegraphics[width=0.9\columnwidth]{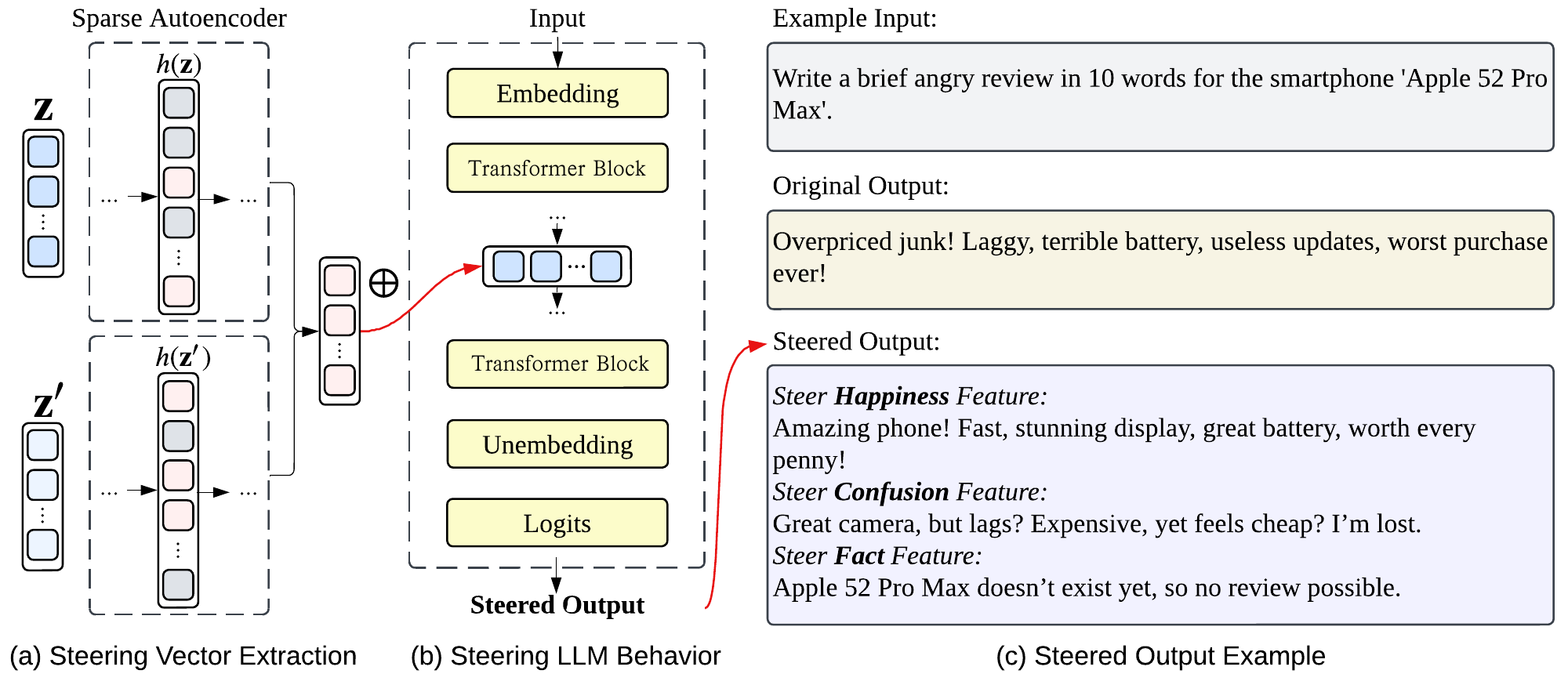}}
\caption{The pipeline for steering LLMs using SAE features. (a) \textbf{Steering Vector Extraction}: The target steering vector is derived by analyzing a set of prompts to identify features that distinguish a concept-rich state $\mathbf{z}'$ from a neutral state $\mathbf{z}$. The steering vector is computed as the weighted sum of these identified SAE features (i.e., decoder columns). (b) \textbf{Steering LLM Behavior}: This aggregated vector is injected into the Transformer's residual stream state $\mathbf{x}^l$ via vector addition. (c) \textbf{Steered Output Example}: Empirical results showing how steering specific features (e.g., Happiness, Confusion) drastically alters the model's generation style even when the original prompt implies a negative sentiment. Based on the figure from~\citet{shu-etal-2025-survey}.}
\label{fig:sae_steering}
\end{figure}

\textbf{1) Dynamic Hidden States:} The primary targets for runtime steering are the residual stream states $\mathbf{x}^l$ and the outputs of attention heads $\mathbf{h}_{\text{attn}}^{l,h}$. For these dynamic objects, the steering vector $\mathbf{v}$ is typically derived using one of two methods:
\begin{itemize}
    \item \textbf{Contrastive Activation Means:} This method, often referred to as ``Activation Addition'' or ``Mass-Mean Shift,'' assumes that a concept can be isolated by comparing the model's internal states across opposing contexts~\citep{rimsky-etal-2024-steering,van2024extending,lu2024investigating,postmus2024steering,turner2024steeringlanguagemodelsactivation,sharma2025steeringconceptualbiastransformer}. Formally, let $\mathcal{D}^+$ be a set of prompts eliciting the target behavior and $\mathcal{D}^-$ be a set eliciting the opposing behavior. The steering vector $\mathbf{v}$ is calculated as the difference between the centroids of the residual stream states $\mathbf{x}^l$ for these two sets:
    \begin{equation}
    \mathbf{v} = \boldsymbol{\mu}^+ - \boldsymbol{\mu}^- = \frac{1}{|\mathcal{D}^+|}\sum_{\mathbf{x}_i \in \mathcal{D}^+} \mathbf{x}^l_i - \frac{1}{|\mathcal{D}^-|}\sum_{\mathbf{x}_j \in \mathcal{D}^-} \mathbf{x}^l_j
    \end{equation}
    By adding $\alpha \cdot \mathbf{v}$ to the residual stream, we shift the model's current state towards the centroid of the positive behavior.

    \item \textbf{SAE Features:} SAEs offer a more precise way to derive $\mathbf{v}$ by utilizing monosemantic features~\citep{wang2025beyond,bayat2025steering,weng2025safe,he2025saif,soo2025interpretable,goyalBreakingBadTokens2025}. As illustrated in Figure~\ref{fig:sae_steering}, the process involves two steps:
    \begin{enumerate}
        \item \textbf{Feature Identification:} First, we collect residual stream states from a positive dataset $\mathcal{D}^+$ (eliciting the target concept, e.g., ``Happiness'') and a negative/neutral dataset $\mathcal{D}^-$. By passing these states through the SAE encoder, we calculate the \textit{differential activation score} $\delta_j$ for each feature $j$:
        \begin{equation}
        \delta_j = \mathbb{E}_{\mathbf{x} \in \mathcal{D}^+}[a_j(\mathbf{x})] - \mathbb{E}_{\mathbf{x} \in \mathcal{D}^-}[a_j(\mathbf{x})]
        \end{equation}
        where $a_j(\mathbf{x})$ denotes the $j$-th feature activation for input $\mathbf{x}$. Features with high positive $\delta_j$ constitute the set of ``Target Features'' $\mathcal{J}$ that specifically encode the desired trait.
        \item \textbf{Vector Construction:} The steering vector $\mathbf{v}$ is then synthesized as the weighted sum of these identified feature. Let $\mathbf{f}_j$ denote the $j$-th feature (the $j$-th column of the SAE decoder weights $\mathbf{W}_{\text{dec}}$). The steering vector is computed as:
        \begin{equation}
        \mathbf{v} = \sum_{j \in \mathcal{J}} \delta_j \cdot \mathbf{f}_j
        \end{equation}
        
    \end{enumerate}
    Finally, this obtained steering vector is injected into the model's residual stream during inference ($\hat{\mathbf{x}} \leftarrow \mathbf{x} + \alpha \cdot \mathbf{v}$). As shown in Figure~\ref{fig:sae_steering} (c), this enables precise manipulation of specific semantic traits like ``Happiness'' or ``Confusion'' to drastically alter generation styles while minimizing interference with unrelated concepts.
\end{itemize}

\textbf{2) Static Parameters:} For static weights, the steering vector $\mathbf{v}$ is explicitly defined as a \textit{Task Vector} in Model Merging~\citep{ilharcoediting2023,yadav2023ties,liu2025sensmerging,yao2025activation}. This vector is computed as the element-wise difference between the weights of a fine-tuned model and its pre-trained base ($\mathbf{v} = \mathbf{W}_{\text{ft}} - \mathbf{W}_{\text{base}}$), effectively encapsulating a transferable skill or behavior.
Recent advancements have evolved beyond simple element-wise addition by employing localization techniques to determine adaptive merging coefficients.
For instance, \citet{liu2025sensmerging} proposed \textit{Sens-Merging}, which utilizes \textit{Gradient Detection}-based sensitivity analysis to evaluate parameter importance, allowing for the precise balancing of weights based on their impact on task performance.
Complementarily, \citet{yao2025activation} introduced \textit{Activation-Guided Consensus Merging}, which leverages \textit{Magnitude Analysis} of internal representations. By calculating the mutual information between activations of the base and fine-tuned models, they derive layer-specific scaling coefficients to optimally integrate the task vector.

\paragraph{Characteristics and Scope}
The scope of this method is characterized by \textbf{additive directionality}. Unlike the precise rewriting in \textit{Targeted Optimization} (\S\ref{sec:target_optimization}), \textit{Vector Arithmetic} acts as a \textbf{steering force}, dynamically pushing the model towards a target attribute without permanently altering weights.
\begin{itemize}[leftmargin=*]
    \item \textbf{Advantages:} It is a lightweight and reversible intervention. Since it typically operates at inference time (for hidden states) or via simple weight addition, it does not require complex optimization or gradient descent during deployment. It allows for flexible control over model behavior by simply adjusting the steering coefficient $\alpha$.
    \item \textbf{Limitations:} The effectiveness relies on the ``Linear Representation Hypothesis.'' If the target concept is not encoded linearly or if the steering vector $\mathbf{v}$ is entangled with other concepts (lacking orthogonality), the intervention might introduce unintended side effects due to geometric interference. Conversely, when concept directions are highly orthogonal, they allow for robust compositionality and reliable steering without disrupting unrelated capabilities.
\end{itemize}

\input{Tables/comparative_steering_methods}

\begin{tcolorbox}[takeawaysbox, title={Comparative Analysis of Steering Methods}]
  Table~\ref{tab:steering_methods_comparison} compares steering methods in terms of steering strength, side effects, access requirements, cost, and key limitations.
  Among them, \textit{Targeted Optimization} offers the strongest and most persistent control because it directly modifies model parameters. It is usually more robust than inference-time interventions, and many methods explicitly include preservation objectives to reduce degradation of unrelated behaviors~\citep{meng2022ccs,meng2023massediting,zhang2024cofitune}. However, it requires full access to weights and gradients and comes with a much higher computational cost. 
  By contrast, \textit{Amplitude Manipulation} and \textit{Vector Arithmetic} are lightweight inference-time interventions. They are attractive in practice because they require only activation-level access and little additional compute. However, they are usually transient, more sensitive to prompt variation, and more vulnerable to side effects when the target feature is polysemantic or entangled with other behaviors~\citep{rimsky-etal-2024-steering,Tan2024}. Recent evidence further suggests that some activation steering methods can improve surface-level behavior while still underperforming on robustness or specificity, and may even increase jailbreak susceptibility when intervention fidelity is weak~\citep{Goyal2026,raedler2025the}.
  Overall, inference-time methods are useful when low cost and easy deployment are the main goals, whereas weight-update methods are preferable when stronger and more persistent control is required.
  \end{tcolorbox}

\newpage

\section{Applications}
\label{sec:applications}
Building on localizing methods (§\ref{sec:localize_methods}) that identify internal objects associated with specific behaviors and steering methods (§\ref{sec:steer_methods}) that intervene on these objects to modulate model outputs, this section summarizes how these lines of work translate into practical use cases. 
We organize the literature around three overarching objectives: \textit{alignment}, \textit{capability}, and \textit{efficiency}. 

\subsection{Improve Alignment}
\label{sec:improve_alignment}

\subsubsection{Safety and Reliability}
\label{sec:safety_reliabilty}
\begin{tcolorbox}[takeawaysbox, title={Summary of Application Paradigms}]
  MI improves safety and reliability in alignment applications primarily through two complementary, mechanism-aware intervention paradigms:
  
  \begin{enumerate}[leftmargin=*, label=\arabic*)]
  \item \textbf{Safety-Critical Component Manipulation.}  
  This paradigm focuses on identifying internal components that explicitly encode unsafe, harmful, or unreliable behaviors, such as toxicity, hallucination, or failed refusal, and intervening on these components directly. By localizing safety-critical attention heads, neurons, circuits, or SAE features, researchers apply inference-time \textit{Amplitude Manipulation} (\S\ref{sec:amplitude_manipulation}) to suppress unsafe activations, or use training-based \textit{Targeted Optimization} (\S\ref{sec:target_optimization}) to permanently rewrite safety-relevant parameters and internal signals.
  \item \textbf{Latent Safety and Reliability Representation Steering.}  
  This paradigm operates at the level of the residual stream, where abstract safety- and reliability-related concepts such as truthfulness, refusal, and instruction following are encoded as approximately linear directions. By identifying these directions using causal or contrastive analyses, models are steered via \textit{Vector Arithmetic} (\S\ref{sec:vector_arithmetic}) to correct hallucinations, enforce proper refusal behavior, or improve instruction adherence, while largely preserving general capabilities.
  \end{enumerate}
  \end{tcolorbox}
  \paragraph{1) Safety-Critical Component Manipulation} Unsafe or unreliable behaviors in LLMs have been shown to be mediated by relatively localized internal components. Accordingly, a body of work first localized safety-relevant objects and then intervened via targeted mechanistic techniques. At the attention level, \citet{zhou2025on} showed that a small subset of attention heads played a disproportionate role in safety-related behaviors, particularly refusal and rejection of harmful queries. Using \textit{Causal Attribution} to localize safety-critical heads and \textit{Amplitude Manipulation} to intervene, they demonstrated that suppressing these heads substantially weakened safety capability while modifying only a negligible fraction of parameters. 
  At the neuron level, several studies applied \textit{Magnitude Analysis} to identify neurons whose activations were strongly associated with unsafe or misaligned behaviors. \citet{zhao2025understanding} introduced \emph{safety neurons} and showed that a very small subset --- predominantly located in early self-attention layers --- collectively governed safety behavior; they then performed \textit{Targeted Optimization} by selectively tuning these neurons during training, significantly improving safety without degrading general performance. Complementarily, \citet{suauWhisperingExpertsNeural2024} used magnitude-based criteria to pinpoint toxicity-related neurons and applied \textit{Amplitude Manipulation} by scaling down their activations at inference time to mitigate toxic generations. Similarly, \citet{gao2025hneuronsexistenceimpactorigin} identified hallucination-associated neurons (\textit{H-neurons}) via \textit{Magnitude Analysis} and validated their causal impact through \textit{Amplitude Manipulation}, showing that suppressing these neurons reduced hallucinations without broadly affecting other capabilities. Beyond individual neurons, recent work leveraged SAEs to disentangle safety-related representations into interpretable features. Using \textit{Magnitude Analysis} over SAE feature activation states, \citet{templeton2024scaling} showed that SAE features extracted from LLMs exhibited strong monosemanticity, including features associated with harmful or toxic content.
  Building on this insight, \citet{goyalBreakingBadTokens2025} applied \textit{Amplitude Manipulation} to suppress selected SAE features and thereby detoxify model outputs. Likewise, \citet{yeo-etal-2025-understanding} performed SAE-based \textit{Magnitude Analysis} to identify harm- and refusal-related feature sets and validated their roles through targeted \textit{Amplitude Manipulation}, enabling fine-grained control and mechanistic insight into refusal behavior.
  
  While \textit{Amplitude Manipulation}-based interventions typically operate at inference time, several works pursued more persistent safety improvements through \textit{Targeted Optimization}. \citet{huang-etal-2025-pierce} identified safety-relevant circuits and updated only parameters within these circuits to mitigate harmful behaviors. At finer granularity, \citet{zhao2025understanding}, \citet{chen2025towards}, and \citet{liPrecisionKnowledgeEditing2024} showed that selectively updating neuron-associated weights enabled precise safety edits with minimal side effects. At a coarser level, \citet{li2025safety} demonstrated that safety behavior could be localized at the layer level, while \citet{leeMechanisticUnderstandingAlignment2024} analyzed how alignment objectives reshaped internal representations during optimization.

  \paragraph{2) Latent Safety and Reliability Representation Steering}
  
  A complementary line of research shows that many safety-relevant behaviors are encoded as approximately linear directions in LLM's latent space, motivating safety interventions based on \textit{Vector Arithmetic} in the residual stream.
  
  \citet{arditi2024refusal} and \citet{zhao2025llms} showed that refusal was encoded as a compact low-dimensional subspace identified via \textit{Causal Attribution}, and that some jailbreaks succeeded by suppressing this refusal signal via \textit{Vector Arithmetic} without changing the model’s harmfulness belief. Extending these findings to reasoning models, \citet{yin2025refusalfallscliffsafety} identified a refusal-cliff phenomenon using \textit{Probing}, where refusal intent was maintained during intermediate reasoning but was abruptly suppressed at the final generation stage, a failure mode attributed to a small set of refusal-suppressing attention heads.
  Building on these analyses, multiple studies identified actionable safety directions and applied steering interventions. \citet{wang2025surgical} proposed a training-free, single-vector ablation method (a form of \textit{Vector Arithmetic}) that selectively removed false refusal while preserving true refusal and general capabilities, enabling fine-grained safety calibration. \citet{wang2025refusal} further demonstrated that refusal directions were approximately universal across safety-aligned languages, helping to explain the effectiveness of cross-lingual jailbreaks as well as vector-based interventions.
  
  \textit{Vector Arithmetic}-based steering was also applied to hallucination reduction and factuality improvement. \citet{chuang2024dola} introduced contrastive layer decoding (a form of \textit{Vector Arithmetic}) during generation to amplify factual signals identified via \textit{Vocabulary Projection}. Similarly, \citet{zhang-etal-2024-truthx} identified a truthfulness direction in the residual space using \textit{Probing} and then edited it via \textit{Vector Arithmetic}, enabling controllable enhancement of truthful behavior. Complementarily, \citet{orgad2025llms} showed that hallucination-related representations could be detected internally via \textit{Probing} even when they were not expressed at the output level, highlighting the diagnostic value of latent safety signals. Finally, recent work applied \textit{Vector Arithmetic} to improve instruction-following reliability. \citet{he2025saif} leveraged SAE-derived directions to steer instruction adherence, while \citet{stolfo2025improving}, \citet{jiang2024refine}, and \citet{li2025training} demonstrated that instruction-following behavior could be improved through \textit{Vector Arithmetic} steering without full retraining.

\subsubsection{Fairness and Bias}
\label{sec:fairness_bias}
\begin{tcolorbox}[takeawaysbox, title={Summary of Application Paradigms}]

  MI facilitates the diagnosis and control of fairness-related biases (gender bias, distributed attribute and cultural bias signals, and evaluation bias) in LLMs through three primary paradigms:
  
  \begin{enumerate}[leftmargin=*, label=\arabic*)]
      \item \textbf{Gender Bias Localization and Selective Debiasing} This paradigm localized gender-bias mediation primarily via \textit{Causal Attribution} (\S\ref{sec:causal_Attribution}) and then reduced bias through either inference-time \textit{Amplitude Manipulation} (\S\ref{sec:amplitude_manipulation}) or persistent \textit{Targeted Optimization} (\S\ref{sec:target_optimization}) on the identified components.
      \item \textbf{Distributed Attribute and Cultural Bias Signals} This paradigm extended beyond gender to demographic, social, and cultural biases, often requiring broader searches over internal structures (including \textit{Magnitude Analysis} (\S\ref{sec:magtitude_analysis}) or \textit{Gradient Detection} (\S\ref{sec:gradient_detection})).
      \item \textbf{Evaluation Bias Engines in Judgment and Framing} This paradigm studied cognitive and judgment biases induced by prompt format or evaluation settings (e.g., positional anchoring and moral attribution), and mitigated them via inference-time controls such as attention/position re-assignment or targeted scaling, typically guided by \textit{Magnitude Analysis} (\S\ref{sec:magtitude_analysis}) and validated by \textit{Causal Attribution} (\S\ref{sec:causal_Attribution}).
  \end{enumerate}
  \end{tcolorbox}
  
  \paragraph{1) Gender Bias Localization and Selective Debiasing}
  Mechanistic studies of gender bias established a canonical fairness pipeline: first localizing bias mediation with \textit{Causal Attribution}, then steering the identified carriers via either transient inference-time control or persistent parameter updates.
  \citet{vig2020gender} provided an early template using causal mediation analysis in GPT-2 to quantify which internal components mediated gendered associations, and demonstrated mitigation by replacing bias-inducing activations with counterfactual ones, a direct instance of \textit{Amplitude Manipulation}.
  To achieve persistent mitigation, subsequent work increasingly shifted to selective updates of localized components via \textit{Targeted Optimization}. \citet{chintamIdentifyingAdaptingTransformerComponents2023} showed that responsibility for gender bias could concentrate in specific late-layer attention heads and reduced bias by fine-tuning only these components. \citet{caiLocatingMitigatingGender2024} characterized a division of labor where lower FFN blocks encoded bias-relevant information while upper attention modules exploited it, proposing an editing-style method to update the responsible subset. Finally, \citet{yuUnderstandingMitigatingGender2025a} refined the intervention granularity to the neuron level, identifying distinct “gender neurons” versus “general neurons” and introducing an interpretable neuron-editing procedure to reduce bias while preserving general performance.
  
  \paragraph{2) Distributed Attribute and Cultural Bias Signals}
  Beyond gender, mechanistic evidence suggested that demographic, social, and cultural biases are often encoded more diffusely. This motivated localization strategies that avoid assuming a single “bias module,” alongside mitigation strategies combining targeted suppression with global representational steering.
  In domain-conditioned settings like healthcare, \citet{ahsanElucidatingMechanismsDemographic2025} used activation patching, a form of \textit{Causal Attribution}, to localize racial information across multiple LLMs, reporting that racial signals are more scattered across early and middle FFN layers compared to gender. Similarly, \citet{yuEntangledRepresentationsMechanistic2025} adopted Patchscope-style interventions to “read out” cultural knowledge from internal representations. Rather than proposing a mitigation, their results focused on diagnosis, revealing how cultural salience and resource imbalance manifest as systematic representational asymmetries.
  Addressing broader societal biases, \citet{liuDevilNeuronsInterpreting2024} employed \textit{Gradient Detection} to identify neurons associated with multiple social attributes and demonstrated mitigation by suppressing their activations. To scale localization beyond hand-picked modules, \citet{chandnaDissectingBiasLLMs2025} combined \textit{Magnitude Analysis} over internal structures with causal validation to create a reusable recipe for bias analysis across attributes. Finally, acknowledging that values can be represented linearly, \citet{kimLinearRepresentationsPolitical2024} used \textit{Probing} to identify attention heads predicting political ideology, and then steered generations via \textit{Vector Arithmetic}.
  
  \paragraph{3) Evaluation Bias Engines in Judgment and Framing}
  A complementary thread targeted \emph{cognitive biases} arising from judgment heuristics, prompt formats, or decision framing, rather than demographic correlations. These works mitigated such biases through inference-time controls guided by importance signals via \textit{Magnitude Analysis} or validated by \textit{Causal Attribution}.
  For positional anchoring in multiple-choice questions (MCQs), \citet{liAnchoredAnswersUnravelling2025} identified higher-layer mechanisms in GPT-2 that preferentially routed evidence toward anchored option tokens, providing concrete intervention loci. Generalizing beyond MCQs, \citet{wangEliminatingPositionBias2025a} formulated position bias across judge-style evaluation and retrieval-augmented QA, introducing a mechanism to re-assign positions based on attention-derived importance signals. Complementarily, \citet{yuMitigatePositionBias} traced “lost-in-the-middle” failures to a positional hidden-state channel and proposed a search-and-scale procedure to rescale this channel, improving robustness on long-context benchmarks.
  Extending to domain-specific decision-making, \citet{diminoTracingPositionalBias2025} localized mid-to-late transformer layers as core “bias engines” driving positional skew in financial advisory tasks. Finally, regarding moral judgment, \citet{raimondiAnalysingMoralBias2025} analyzed the Knobe effect, localized its mediation to residual activations, and reduced the intentionality attribution gap by patching fine-tuned states with their pretrained counterparts, selectively reverting value shifts introduced during alignment.

\subsubsection{Persona and Role}
\label{sec:personalization}

\begin{tcolorbox}[takeawaysbox, title={Summary of Application Paradigms}]
  MI facilitates the analysis and control of LLM personas and roles through three primary paradigms, ranging from global representation engineering to fine-grained component editing:
  \begin{enumerate}[leftmargin=*, label=\arabic*)]
  \item \textbf{Global Persona Modulation via Vectors:} This paradigm posits that high-level personality traits (e.g., sycophancy, honesty) are encoded as linear directions within the global activation space. Researchers extract these ``Persona Vectors'' and apply \textit{Vector Arithmetic} (\S\ref{sec:vector_arithmetic}) to the residual stream, steering the model's behavior without altering weights.
  \item \textbf{Persona-Specific Component Editing:} Moving beyond global vectors, this approach identifies specific model components, such as individual neurons or attention heads, that serve as the physical carriers of personality traits. These components are then targeted via \textit{Amplitude Manipulation} (\S\ref{sec:amplitude_manipulation}) or refined through \textit{Targeted Optimization} (\S\ref{sec:target_optimization}) to achieve persistent behavioral changes.
  \item \textbf{Psychological Profiling and Diagnosis:} Instead of active intervention, this paradigm utilizes MI as a diagnostic tool. By employing \textit{Probing} (\S\ref{sec:probing}) techniques and analyzing activation geometry, researchers can locate where psychological traits emerge, validate the stability of roles, and predict model behaviors before generation occurs.
  \end{enumerate}
  \end{tcolorbox}
  
  \paragraph{1) Global Persona Modulation via Vectors}
  A growing body of work suggests that complex persona-specific behavioral traits can be manipulated by intervening in the global activation state of the model.
  \citet{rimsky-etal-2024-steering} utilized Contrastive Activation Addition, a form of \textit{Vector Arithmetic}, to steer models away from sycophantic and hallucinatory behaviors. By extracting steering vectors from the residual stream differences between positive and negative examples, they demonstrated that high-level alignment properties can be precisely modulated during inference without fine-tuning.
  Expanding on this, \citet{chen2025persona} developed an automated pipeline to extract ``Persona Vectors'' for arbitrary traits (e.g., ``evil'' or ``sycophantic'') using natural language descriptions. They found that these vectors not only allow for post-hoc steering but can also be used to predict and mitigate unintended persona shifts (e.g., emergent misalignment) that occur during fine-tuning by monitoring the projection of training data onto these vectors.
  \citet{poterti-etal-2025-role} applied this concept to professional domains, constructing ``Role Vectors'' (e.g., Chemist, Doctor) from model activations. Their analysis revealed that reinforcing these role-specific directions significantly improves performance on domain-specific tasks and even yields cross-domain benefits, suggesting that role-playing is mechanistically grounded.
  Furthermore, \citet{pai2025billy} proposed \textit{BILLY}, a training-free framework that blends multiple persona vectors (e.g., Creative Professional + Environmentalist) to simulate collective intelligence within a single model. This approach steers the model with a composite vector, enhancing creativity and diversity in generation without the computational cost of multi-agent systems.
  Similarly, \citet{sun-etal-2025-personality} explored the task vector (i.e., the steering vector in the context of model merging), extracting personality vectors by subtracting pre-trained weights from fine-tuned ones. They showed that these vectors can be linearly composed to continuously modulate trait intensity (e.g., Extraversion) across different models.
  Finally, \citet{handa2025personality} conducted a rigorous comparative study of personality manipulation methods using the Big Five traits. 
  Their results showed that \textit{Vector Arithmetic} provides a lightweight yet effective approach for controlling model personas at inference time.
  
  \paragraph{2) Persona-Specific Component Editing}
  Rather than steering the global state, this paradigm seeks to identify and edit the specific neural components responsible for personality expression.
  \citet{deng2025neuron} proposed \textit{NPTI}, a method that identifies ``personality-specific neurons'' by applying \textit{Magnitude Analysis} on the activation differences between opposing trait descriptions (e.g., Extraversion vs. Introversion). By selectively activating or deactivating these neurons via \textit{Amplitude Manipulation}, they achieved fine-grained control over the model's personality without model training.
  \citet{su-etal-2025-understanding} extended this to ethical values, introducing \textit{ValueLocate}. They constructed a dataset based on the \textit{Schwartz Values Survey}~\citep{schwartz1992universals}, a well-established framework that classifies values into four dimensions: Openness to Change, Self-transcendence, Conservation, and Self-enhancement. Using this dataset, they located value-critical neurons and demonstrated that controlling them via \textit{Amplitude Manipulation} can effectively alter the model's value orientation.
  Addressing the specific issue of sycophancy, \citet{chen2024from} identified a sparse set of attention heads ($\sim$4\%) that significantly contribute to ``yes-man'' behavior. They proposed \textit{Supervised Pinpoint Tuning}, a form of \textit{Targeted Optimization}, which fine-tunes only these specific heads while freezing the rest of the model, successfully mitigating sycophancy while preserving general reasoning abilities better than standard instruction tuning.
  
  \paragraph{3) Psychological Profiling and Diagnosis}
  MI techniques are also extensively used to understand how psychological constructs are represented internally by applying \textit{Probing}.
  \citet{tak-etal-2025-mechanistic} investigated emotion inference, finding that emotion processing is functionally localized in MHA units within middle layers. 
  They validated this by showing that interventions on latent ``appraisal concepts'' (e.g., pleasantness) predictably shift the generated emotional tone.
  \citet{yuan2025monolingual} explored how language identity affects psycholinguistic traits like sound symbolism and word valence. Their \textit{Probing} analysis revealed that these signals become decodable in deeper layers and that language conditioning (e.g., bilingual persona) significantly modulates internal representations.
  \citet{ju2025probing} introduced a layer-wise \textit{Probing} framework to analyze the Big Five personality traits, discovering that personality information is predominantly encoded in the middle and upper layers. They further proposed a method to edit response personality by applying \textit{Amplitude Manipulation} to perturb hidden states orthogonal to the probing boundaries.
  In the realm of truthfulness, \citet{joshi2024personas} proposed the ``persona hypothesis,'' suggesting LLMs model truthfulness by inferring a ``truthful persona'' from the context. They provided evidence that \textit{Probing} for this persona can predict the truthfulness of generated answers.
  \citet{ghandeharioun2024whos} utilized techniques like \textit{Patchscopes} to reveal ``latent misalignment,'' showing that user personas (e.g., ``altruistic'' vs. ``selfish'') significantly affect a model's willingness to answer harmful queries, mediated by internal interpretations of the user's intent.
  For user-facing transparency, \citet{karny2025neural} developed an interface that visualizes ``Persona Scores'' derived from neural activations. Their user study highlighted that users often miscalibrate their expectations of model behavior, and neural transparency tools can bridge this gap.
  Finally, \citet{banayeeanzade2025psychological} introduced \textit{PsySET}, their evaluation results showcase that although \textit{Vector Arithmetic} steering is effective for modulating persona traits, it can introduce unintended side effects, such as ``joy'' steering reducing privacy awareness or ``anger'' steering increasing toxicity, necessitating rigorous safety evaluations.
  \citet{bas2025steering} further differentiated between steering ``internal dispositions'' versus ``external knowledge'', finding that steering method such as \textit{Vector Arithmetic} is highly effective for latent traits (e.g., personality) but struggles with knowledge-heavy personas (e.g., specific public figures), where it often degrades coherence.

\subsection{Improve Capability}
\label{sec:improve_capability}

\subsubsection{Multilingualism}
\label{sec:multilingual_crosslingual}

\begin{tcolorbox}[takeawaysbox, title={Summary of Application Paradigms}]
  In multilingual and cross-lingual settings, MI enables targeted control and enhancement of language behavior in LLMs through two primary application paradigms:
  \begin{enumerate}[leftmargin=*, label=\arabic*)]
  \item \textbf{Language-Specific Component Manipulation:}  
  This paradigm focuses on identifying and intervening on internal components that are specifically responsible for processing individual languages. By localizing \emph{language-specific neurons} or \emph{SAE features} via \textit{Magnitude analysis} (\S\ref{sec:magtitude_analysis}), researchers directly manipulate their magnitudes through \textit{Amplitude Manipulation} (\S\ref{sec:amplitude_manipulation}) to control output language, enhance multilingual performance, or perform language-specific adaptation.
  \item \textbf{Cross-Lingual Representation Steering:}  
  This paradigm operates at the level of the residual stream, where multilingual representations are dynamically transformed across layers. By identifying language-related directions using \textit{Vocabulary Projection} (\S\ref{sec:vocab_project}) or \textit{Magnitude Analysis} (\S\ref{sec:magtitude_analysis}), models are steered via \textit{Vector Arithmetic} (\S\ref{sec:vector_arithmetic}) to align representations across languages, improve cross-lingual transfer, and mitigate language inconsistency or language mixing phenomena.
  \end{enumerate}
  \end{tcolorbox}
  
  \paragraph{1) Language-Specific Component Manipulation}
  
  A central line of multilingual MI research shows that multilingual capabilities in LLMs are supported by a relatively small subset of internal components exhibiting strong language specificity. Accordingly, existing work has focused on localizing these components and manipulating their activations to control output language or enhance multilingual performance.
  \citet{zhao-etal-2024-multilingual} formalized this observation through a layered \textit{Multilingual Workflow (MWork)}, showing that representations became English-centric in intermediate layers and were mapped back to the query language in later layers. They employed \textit{PLND} to localize language-specific neurons and showed that intervening on only a tiny fraction could sharply disrupt multilingual performance, while selectively updating these neurons via \textit{Target Optimization} enabled data-efficient language-specific adaptation. 
  Along similar lines, \citet{tang-etal-2024-language} introduced \textit{Language Activation Probability Entropy (LAPE)} as a \textit{Magnitude Analysis} tool to quantify cross-language activation selectivity. Their results showed that language-specific neurons concentrated in the bottom and top layers, and that applying \textit{Amplitude Manipulation} to these neurons provided direct control over output language, effectively reducing off-target generation. 
  Complementing these localization-driven studies, \citet{kojima-etal-2024-multilingual} analyzed neuron activation patterns across languages using \textit{Magnitude Analysis} and confirmed the functional importance of language-specific neurons through \textit{Amplitude Manipulation}, including targeted ablation and scaling. Together, these results reinforced component-level activation control as a practical mechanism for multilingual intervention.
  
  Beyond identifying individual language-specific neurons, \citet{gurgurov2025languagearithmeticssystematiclanguage} systematized neuron-level \textit{Amplitude Manipulation} through Language Arithmetic, demonstrating that language-specific neurons exhibited additive properties that enabled controlled language switching and interpolation via linear operations on activations.
  In parallel, related work extended this paradigm beyond language identity to other forms of linguistic specialization. \citet{liu-etal-2025-relation-specific} identified relation-specific neurons whose activation patterns generalized across languages in multilingual factual probing tasks, demonstrating that neuron-level specialization could transfer cross-lingually beyond language identity. 
  At a finer representational granularity, \citet{jing-etal-2025-lingualens} employed SAEs to extract and analyze a wide range of interpretable linguistic features whose activation patterns varied systematically across languages, and showed that \textit{Amplitude Manipulation} of these features could causally affect corresponding linguistic behaviors. 
  Similarly, \citet{andrylie2025sparseautoencoderscapturelanguagespecific} identified language-specific SAE features via \textit{Magnitude Analysis} and demonstrated that \textit{Amplitude Manipulation} on these features enabled fine-grained control over multilingual behavior. 
  Finally, \citet{brinkmann-etal-2025-large} showed that SAE-based representations captured shared, cross-lingual grammatical abstractions, with targeted feature-level analyses providing supporting evidence.
  
  \paragraph{2) Cross-Lingual Representation Steering in Residual Space}
  
  A second major paradigm improves multilingual behavior by intervening on internal representations in the residual stream, where language representations are progressively transformed and aligned across layers. \citet{chi-etal-2023-cross} showed that cross-lingual transfer could be activated without task-specific supervision by restructuring and aligning multilingual representations across model components, suggesting that pretrained models encoded latent cross-lingual structure that could be activated without end-task data. 
  To localize where multilingual representations diverged or aligned, several studies relied on \textit{Vocabulary Projection}. In particular, \citet{wendler2024llamas} revealed that multilingual models often operated in an English-centric latent space during intermediate layers, even for non-English inputs, motivating interventions in later layers to restore language-faithful generation.
  Complementary representation-space analyses further supported this view: \citet{philippy2023identifying} analyzed the relationship between language distance and representation divergence, while \citet{mousi2024exploring} studied alignment dynamics in shared multilingual spaces using clustering-based metrics, together characterizing how cross-lingual alignment evolved across layers.
  
  Building on these localization insights, subsequent work intervened more directly on internal representations to influence multilingual behavior. \citet{hinck-etal-2024-llava} analyzed English-dominant responses in vision-language models and showed that targeted \textit{Vector Arithmetic} on internal attention and hidden states could mitigate this bias. 
  More recent studies moved from localization to \emph{failure-mode diagnosis} with MI tools. Using layer-wise \textit{Vocabulary Projection} and representation analysis, \citet{wang-etal-2025-lost-multilinguality} attributed cross-lingual factual inconsistency to late-layer transitions into language-related subspaces, while \citet{liu-etal-2025-tracing} traced when multilingual factual knowledge emerged across pretraining checkpoints, providing a developmental view of cross-lingual consistency rather than a post-hoc intervention account.
  Complementarily, \citet{wang-etal-2025-language-mixing} analyzed language mixing in reasoning by characterizing when and how internal states drift between languages during generation. \citet{nie-etal-2025-mechanistic} further combined late-layer lens-style analysis with targeted neuron-level interventions (\textit{Amplitude Manipulation}) to mitigate language confusion. 

\subsubsection{Knowledge Management}
\label{sec:knowledge_management}

\begin{tcolorbox}[takeawaysbox, title={Summary of Application Paradigms}]
  MI enables precise analysis, control, and consolidation of model knowledge through three complementary paradigms, ranging from local intervention to global composition:
  \begin{enumerate}[leftmargin=*, label=\arabic*)]
      \item \textbf{Precise Knowledge Updating:} This paradigm identifies minimal carriers responsible for a target association using \textit{Causal Attribution} (\S\ref{sec:causal_Attribution}), optionally assisted by \textit{Gradient Detection} (\S\ref{sec:gradient_detection}) or \textit{Vocabulary Projection} (\S\ref{sec:vocab_project}). Once located, associations can be modified either persistently through \textit{Targeted Optimization} (\S\ref{sec:target_optimization}) or transiently via \textit{Amplitude Manipulation} (\S\ref{sec:amplitude_manipulation}), achieving high specificity while controlling generalization.
  
      \item \textbf{Knowledge Retention and Stability:} MI supports diagnosing interference under continual updates or context injections. Critical carriers are located mainly through \textit{Magnitude Analysis} (\S\ref{sec:magtitude_analysis}), \textit{Causal Attribution} (\S\ref{sec:causal_Attribution}) and \textit{Gradient Detection} (\S\ref{sec:gradient_detection}), and stability is maintained by either constraining training-time changes or applying inference-time \textit{Amplitude Manipulation} (\S\ref{sec:amplitude_manipulation}) to reduce drift.  
  
      \item \textbf{Knowledge Consolidation:} To integrate multiple skills or fine-tuned variants, MI identifies compatible subspaces or transferable feature bases using \textit{Gradient Detection} (\S\ref{sec:gradient_detection}), \textit{Magnitude Analysis} (\S\ref{sec:magtitude_analysis}) or \textit{Probing} (\S\ref{sec:probing}). These objects are then combined via \textit{Vector Arithmetic} (\S\ref{sec:vector_arithmetic}) to merge capabilities while preserving essential associations. 
  \end{enumerate}
  \end{tcolorbox}
  
  \paragraph{1) Precise Knowledge Updating}
  MI-based knowledge updating shares a common workflow: First localizes carriers responsible for a target association, then intervenes either at the parameter level (persistent) or activation level (reversible), with careful measurement of locality and collateral effects.
  
  \begin{itemize}[leftmargin=*]
      \item \textbf{Localized Parameter Rewriting:}
      A core result is that many factual associations are mediated by localized pathways, often concentrated in mid-layer FFN output $\mathbf{h}_\text{ffn}^{l}$ and neuron activation state~$\mathbf{s}^l$. 
      \citet{meng2022ccs} used \textit{Causal Attribution} to identify carriers responsible for factual recall and applied structured weight edits (on FFN matrices such as $\mathbf{W}_{\text{out}}^l$) to rewrite specific associations, a process referred to as \textit{Targeted Optimization}, providing a mechanistic alternative to diffuse fine-tuning. 
      Scaling beyond single edits, \citet{meng2023massediting} extended this paradigm to large edit batches by coordinating updates across multiple layers, demonstrating that persistent rewriting could remain localized while handling substantial edit volume. 
      Subsequent work refined the localization premise: \citet{chen2024querylocalization} argued that editability was frequently \emph{query-conditioned}, motivating consistency-aware localization under a broader Query Localization assumption, rather than a fixed set of knowledge neurons. 
      For long-form QA, \citet{chen2024qrnca} introduced QRNCA (a form of \textit{Causal Attribution}), which yielded actionable neuron groups that better tracked query semantics. 
      In multilingual settings, \citet{zhang2024lulafns} identified language-agnostic factual neurons via  \textit{Magnitude Analysis} and applied \textit{Targeted Optimization} on these shared neurons to improve cross-lingual edit consistency. 
      In backward propagation, \citet{katz2024backwardlens} complemented forward analyses with \textit{Vocabulary Projection} of backward-pass gradients, offering an orthogonal diagnostic on where learning signals concentrated during updates.
  
      \item \textbf{Activation-Space Editing and Unlearning:}
      When persistent rewrites are undesirable (e.g., reversible control or safety-motivated removal), activation-level interventions on residual stream states $\mathbf{x}^l$ at layer $l$, or on head/feature activations, provide a practical alternative. 
      \citet{lai2025jola} jointly localized and edited attention-head computations (intervening on attention head output~$\mathbf{h}^{l,h}_\text{attn}$) through gated activation control, instantiating a targeted form of \textit{Targeted Optimization}.
      SAE-based approaches decomposed residual stream states $\mathbf{x}^l$ into sparse features with activations~$\mathbf{a}$, enabling feature-level interventions:  
      \citet{muhamed2025dsg} proposed dynamic SAE guardrails that selected and scaled relevant features via \textit{Magnitude Analysis} to achieve precision unlearning with improved forget--utility trade-offs, while \citet{goyalBreakingBadTokens2025} applied \textit{Amplitude Manipulation} to steer toxicity-related SAE features (scaling selected $\mathbf{f}_j$ via their activations $a_j$) to reduce harmful generations with controlled fluency impact.
  \end{itemize}
  
  \paragraph{2) Knowledge Retention and Stability}
  Retention work traced failures induced by repeated updates or context injection to identifiable carriers, and stabilized behavior via inference-time suppression or training-time adaptation, guided by MI diagnostics. Residual stream states $\mathbf{x}^l$ and attention head outputs $\mathbf{h}_{\text{attn}}^{l,h}$ are often key objects of intervention.
  
  \begin{itemize}[leftmargin=*]
      \item \textbf{Conflict Suppression and Mitigation:}
      Failures under retrieval or context injection often arose from attention heads that mediated the integration of parametric memory and external evidence in the residual stream. 
      \citet{jin2024ph3} performed \textit{Causal Attribution} to localize conflict-mediating heads and applied test-time head suppression/patching, i.e., \textit{Amplitude Manipulation} over attention head output $\mathbf{h}_{\text{attn}}^{l,h}$, to rebalance memory vs.\ context usage. 
      \citet{li2025taming} further used \textit{Magnitude Analysis} to identify heads exhibiting superposition effects and applied targeted gating via \textit{Targeted Optimization} to stabilize behavior under conflicts. 
      Long-context distraction was traced to entrainment-related heads: \citet{niu-etal-2025-llama} localized such heads using \textit{Causal Attribution} and ablated or modulated their outputs ($\mathbf{h}_{\text{attn}}^{l,h}$), reducing echoing of irrelevant context tokens. 
      \citet{jin2025massivevalues} further characterized concentrated massive values in computations mediated by the Q/K weight matrices $\mathbf{W}_Q^{l,h}$ and $\mathbf{W}_K^{l,h}$ (reflected in attention scores $\mathbf{A}^{l,h}$ via \textit{Magnitude Analysis}), then guided \textit{Amplitude Manipulation} over corresponding head outputs $\mathbf{h}_{\text{attn}}^{l,h}$ to maintain contextual reading without disrupting magnitude-structured signals.
  
      \item \textbf{Constraining Continual Adaptation:}
      To reduce catastrophic forgetting, MI localized stability-critical carriers and restricted learning via \textit{Targeted Optimization}. 
      \citet{zhang2024linguistic} applied \textit{Gradient Detection} to identify a ``core linguistic'' parameter region and froze it, mitigating forgetting. 
      \citet{zhang2024cofitune} further constrained adaptation through coarse-to-fine module selection and soft masking, balancing specialty and versatility. 
      Representation-level interventions were also employed: \citet{wu2024reft} localized residual stream states $\mathbf{x}^l$ and applied lightweight edits on $\mathbf{x}^l$ with a frozen backbone (a form of \textit{Targeted Optimization}), improving stability relative to weight-centric updates. 
      Monitoring side effects, \citet{du2024tst} used \textit{Probing} over residual stream states and attention heads to detect security-relevant drift and selected safer module update schedules, enabling controlled adaptation.
  \end{itemize}
  
  \paragraph{3) Knowledge Consolidation}
  Consolidation composes multiple specialized models by combining internal carriers while controlling interference. A common approach represents each fine-tuned model as a parameter ``task vector'' (a delta from a shared base) and merges these deltas via \textit{Vector Arithmetic}.
  
  \citet{yadav2023tiesmerging} improved multi-model composition over naive averaging by first trimming task vectors (a form of \textit{Magnitude Analysis}), resolving sign conflicts, and then merging consistent update directions.
  \textit{Sens-Merging} \citep{liu2025sensmerging} further computed layer-wise sensitivity scores via \textit{Gradient Detection} to weight deltas during merging, yielding stronger merged performance across diverse capability suites. Differently, \citet{yao2025activation} used \textit{Magnitude Analysis} over layer-specific task vectors to derive importance scores that modulate merge weights, improving alignment of merged models with dominant capability directions.
  Beyond parameter deltas, \citet{chen2025stitching} showed that affine mappings between residual stream states (a form of \textit{Probing}) could transfer linear features across models, enabling consolidation at the level of feature bases and amortizing training cost across model sizes.

\subsubsection{Logic and Reasoning}
\label{sec:logic_reasoning}

\begin{tcolorbox}[takeawaysbox, title={Summary of Application Paradigms}]
  MI enhances the logical deduction and reasoning capabilities of LLMs through three distinct paradigms, moving from structural optimization to dynamic inference control:
  \begin{enumerate}[leftmargin=*, label=\arabic*)]
  \item \textbf{Specific Refinement of Numerical and Logical Components:} Instead of blindly updating all parameters, this paradigm involves localizing the specific neurons or attention heads responsible for numerical computation and logical operators. These critical carriers are then strengthened via \textit{Targeted Optimization} (\S\ref{sec:target_optimization}) to improve arithmetic precision.
  \item \textbf{Inference Trajectory Steering:} By isolating directions in the activation space that correspond to high-level reasoning strategies (e.g., ``step-by-step'' planning), researchers can modulate the model's cognitive process. This is achieved by injecting steering vectors via \textit{Vector Arithmetic} (\S\ref{sec:vector_arithmetic}) or amplifying specific features via \textit{Amplitude Manipulation} (\S\ref{sec:amplitude_manipulation}).
  \item \textbf{Stepwise Diagnosis and Correction:} To ensure reliability, monitors based on \textit{Probing} or \textit{Magnitude Analysis} are deployed to track internal states during reasoning. These tools diagnose logical fallacies or uncertainty in real-time, enabling selective self-correction before errors propagate.
  \end{enumerate}
  \end{tcolorbox}
  
  \paragraph{1) Specific Refinement of Numerical and Logical Components}
  LLMs often struggle with precise arithmetic operations. Rather than treating the model as a monolith, MI research has demonstrated that mathematical abilities are often localized within specific sub-modules.
  \citet{quirke2024understanding} conducted a granular circuit analysis of modular addition, identifying that specific attention heads and MLP layers form a dedicated algorithm for numerical processing. By characterizing these circuits, they demonstrated that targeted interventions on these specific components could predictably alter the model's output distribution.
  \citet{yang2024chainofthoughtlargelanguagemodels} analyzed the activation dynamics of CoT processes, revealing that reasoning tasks predominantly activate a broader set of neurons in the final layers compared to standard prompting.
  Leveraging such insights,
  \citet{zhang2024interpreting} proposed an ``identify-analyze-finetune'' pipeline. This method first identified ``reasoning-critical'' attention heads and FFNs via \textit{Causal Attribution}, then froze most model parameters and performed \textit{Targeted Optimization} exclusively on these identified components to boost computational performance.
  Similarly, \citet{tanvocab_2025arxiv} decomposed the language model policy into ``Internal Layer Policies.'' Identifying that early layers maintain high entropy to facilitate exploration, they proposed \textit{Bottom-up Policy Optimization (BuPO)}, a method that selectively optimizes these foundational layers to refine the model's internal reasoning policy efficiently.
  
  \paragraph{2) Inference Trajectory Steering}
  Beyond basic arithmetic, complex reasoning requires the adoption of effective strategies. MI methods enable the extraction and injection of these high-level cognitive patterns by manipulating the model's internal representations via \textit{Vector Arithmetic} and \textit{Amplitude Manipulation}.
  Researchers have extensively utilized \textit{steering vectors} to modulate reasoning behaviors.
  \citet{venhoff2025understandingreasoningthinkinglanguage} utilized contrastive activation means to extract ``Backtracking Steering Vectors,'' demonstrating that injecting this vector increases the model's tendency to self-correct.
  Expanding on this to address inefficient reasoning modes, \citet{zhang2025understanding} identified specialized attention heads correlated with distinct cognitive behaviors—such as verification and backtracking—and proposed intervening on these heads to dynamically steer models away from inefficient ``underthinking'' or ``overthinking'' trajectories.
  Similarly, \citet{hjer2025improvingreasoningperformancelarge} and \citet{tang-etal-2025-unlocking} derived control vectors from residual streams to elicit reasoning capabilities; notably, \citet{tang-etal-2025-unlocking} showed that ``Long Chain-of-Thought'' capabilities can be unlocked via representation engineering without extensive fine-tuning.
  \citet{hong-etal-2025-reasoning} identified a single linear feature direction that mediates the trade-off between reasoning and memorization, allowing for causal control over the model's problem-solving mode.
  \citet{zhang2025uncoveringlatentchainthought} discovered ``Latent CoT'' vectors that, when injected, induce reasoning patterns without explicit natural language prompting.
  For more granular control, \citet{liu2025fractionalreasoninglatentsteering} introduced ``Fractional Reasoning,'' which enables continuous adjustment of reasoning intensity at inference time by scaling latent steering vectors.
  Taking dynamic cognitive strategy selection a step further, \citet{nguyen2026atlas} introduced \textit{ATLAS}, which employs an external latent verifier to dynamically evaluate ongoing reasoning states and adaptively adjust steering intensity at test time, effectively tailoring the cognitive strategy to each specific problem instance.
  Efficiency is also a key benefit; \citet{sinii2025steeringllmreasoningbiasonly} demonstrated that training a single steering vector (bias-only adaptation) matches the reasoning performance of fully RL-tuned models.
  Taking a different approach, \citet{wangelicitingcot_aaai2026} proposed an optimization-based framework: instead of training weights, they optimized hidden representations directly to maximize the likelihood of reasoning paths, utilizing these optimized states to guide the model's trajectory.
  \citet{li-etal-2025-feature} proposed a dual framework, utilizing SAEs to extract interpretable reasoning features while also introducing an SAE-free algorithm to compute steering directions directly from residual activations.
  \citet{zhang2025fantastic} proposed the \textit{RISE} framework, applying SAEs to step-level activations to perform unsupervised discovery of disentangled features corresponding to complex reasoning behaviors like reflection and backtracking. Furthermore, they demonstrated that intervening on these specific feature activations enables the direct adjustment of the model's reasoning mode.
  \citet{galichin2025have} employed SAEs and introduced ``ReasonScore'' (a form of \textit{Magnitude Analysis}) to identify sparse features associated with uncertainty and exploratory thinking. By amplifying these features via \textit{Amplitude Manipulation}, they successfully guided the model toward more robust reasoning.
  \citet{troitskii-etal-2025-internal_emnlp2025} focused on latent states preceding ``wait'' tokens. They located specific features that promote or suppress these tokens and showed that modulating them fundamentally alters the subsequent reasoning process.
  Regarding latent states, \citet{cywinski2025interpretlatentreasoning} demonstrated the feasibility of transplanting reasoning patterns. By employing \textit{Causal Attribution} to localize critical latent vectors and subsequently applying patching (a form of \textit{Amplitude Manipulation}), they effectively forced the model to adopt specific latent reasoning paths.
  
  \paragraph{3) Stepwise Diagnosis and Correction}
  A major challenge in multi-step reasoning is error propagation. MI provides tools for real-time internal diagnosis.
  \citet{sun-etal-2025-probing_emnlp2025} introduced a \textit{Probing} framework to detect reasoning failures. They trained lightweight classifiers on the model's internal activations to distinguish between correct and hallucinatory reasoning steps, acting as an internal monitor to flag errors before the final output is generated.
  Taking a probabilistic perspective, \citet{you-etal-2025-probabilistic_emnlp2025} introduced \textit{ARES}, a framework that employs \textit{Magnitude Analysis} on the entailment probability of internal states. They found that distinct uncertainty patterns emerge when the model deviates from logic. Based on this, they proposed a self-correction mechanism: when the internal monitor detects high ``reasoning uncertainty,'' the model is triggered to backtrack and regenerate the current step, significantly improving the reliability of long-chain deductions.
  Finally, \citet{wu2023analyzing} employed \textit{Gradient Detection} to compute feature attribution scores for tokens within CoT traces. While primarily analytic, this method serves as a potential diagnostic tool by quantifying the semantic influence of each reasoning step, allowing researchers to verify whether the model is attending to relevant logic or spurious correlations during generation.

\subsection{Improve Efficiency}
\label{sec:improve_efficiency}

\subsubsection{Efficient Training}
\label{sec:efficient_training}
\begin{tcolorbox}[takeawaysbox, title={Summary of Application Paradigms}]
  MI noticeably enhances training efficiency by shifting model optimization from a ``black-box'' paradigm to one guided by internal redundant structures and evolutionary dynamics. This application primarily follows two paradigms:
  \begin{itemize}
  \item[1)] \textbf{Sparse Fine-tuning}: By uncovering the model's intrinsic sparsity, researchers isolate and update only critical subnetworks via \textit{Targeted Optimization} (\S\ref{sec:target_optimization}). Unlike standard PEFT methods that introduce external modules, this paradigm modifies model-intrinsic weights, often matching full model fine-tuning performance while drastically reducing computational and memory overhead.
  \item[2)] \textbf{Training Dynamics Monitoring}: Leveraging \textit{Magnitude Analysis} (\S\ref{sec:magtitude_analysis}) and singular learning theory, this paradigm develops internal metrics to track the emergence of specific capabilities and generalization phases. By capturing phase transitions that traditional validation loss may miss, it enables informed decisions on early stopping and prevents unnecessary computations.
  \end{itemize}
  \end{tcolorbox}
  
  \paragraph {1) Sparse Fine-tuning} Unlike PEFT methods that introduce external modules~\citep{li2021prefixtuningoptimizingcontinuousprompts,hu2022lora,liu-etal-2022-p}, achieve efficiency by fine-tuning intrinsic subsets, often matching or exceeding the performance of full fine-tuning.
  At the neuron granularity, researchers utilize diagnostic tools to pinpoint task-specific units. 
  \citet{zhu-etal-2024-landermt} proposed the \textit{LANDeRMT} framework, which employs Taylor expansion to evaluate the ``awareness score'' of FFN neurons for machine translation, enabling \textit{Gradient Detection}-based selective update of language-general and language-specific neurons to mitigate parameter interference. 
  \citet{song-etal-2024-sift} introduced \textit{SIFT}, which exploits the ``quasi-sparsity'' of pre-trained gradients—where the top 1\% of components can account for 99\% of the total gradient norm—using hook functions to perform memory-efficient in-place sparse updates via \textit{Gradient Detection}. 
  Similarly, \citet{xu-etal-2025-lets} developed \textit{NeFT}, which identifies sensitive neurons through \textit{Magnitude Analysis} by calculating the cosine similarity between weights before and after a brief full-parameter fine-tuning run. 
  Furthermore, \citet{mondal-etal-2025-language} and \citet{gurgurov2025sparsesubnetworkenhancementunderrepresented} leveraged \textit{Language Activation Probability Entropy} \citep{tang-etal-2024-language} to identify language-sensitive neurons via \textit{Magnitude Analysis}, achieving significant gains by updating less than 1\% of the model. 
  
  More granular approaches achieve massive efficiency by isolating extremely sparse mechanistic components. 
  \citet{zhao-etal-2024-multilingual} proposed \textit{Parallel Language-specific Neuron Detection}, identifying consistently activated neurons for specific languages without labeled data via \textit{Causal Attribution}; they found that deactivating just 0.13\% of these neurons causes a total loss in multilingual generation. 
  \citet{sergeev2025optimizingmultimodallanguagemodels} introduced \textit{Head Impact scores} based on \textit{Magnitude Analysis} to identify attention heads, demonstrating that fine-tuning only 0.01\% of parameters in the highest-impact layers significantly improves model understanding capability. 
  \citet{lai2025jola} proposed \textit{JOLA}, a framework that employs HardConcrete gates with expected-$L_0$ regularization to jointly learn which attention heads to edit and whether to apply additive or multiplicative interventions. 
  Furthermore, \citet{li2025finetuningsubgraphsearchnew} reframed fine-tuning as a ``subgraph search'' process, introducing a circuit-tuning algorithm that iteratively builds and optimizes a task-relevant circuit via \textit{Circuit Discovery} within the computational graph to preserve general model capabilities.
  
  \paragraph{2) Training Dynamic Monitoring} 
  The second paradigm predominantly leverages \textit{Magnitude Analysis} and other quantitative diagnostics to monitor the state evolution of internal objects, addressing the limitations of traditional validation loss in capturing critical phase transitions.
  
  In the context of \textit{Grokking}---where generalization emerges long after overfitting---MI metrics provide crucial signals that enable practitioners to \textit{confidently continue training despite zero progress in validation loss}. 
  Understanding that grokking arises from the competition between fast-learning memorization circuits and slow-learning but efficient generalization circuits~\citep{varma2023explaining,huang2024unified}, researchers have developed specific indicators to track the latter's formation. 
  \citet{nandaprogress} proposed \textit{Restricted Loss}, a metric derived by projecting weights onto the Fourier basis, which reveals that structured mechanisms form gradually during the apparent loss plateau. 
  Similarly, \citet{furutatowards} introduced \textit{Fourier Frequency Density (FFD)} to characterize the sparsity of internal representations; tracking \textit{FFD} allows for real-time assessment of generalizability, serving as a reliable proxy for circuit maturation. 
  Moving to early detection, \citet{notsawo2023predicting} analyzed the \textit{spectral signature} of the training loss curve itself, demonstrating that specific low-frequency oscillations in early epochs can effectively predict whether grokking will eventually occur, thus saving computational resources on unpromising runs. 
  In the realm of Mixture-of-Experts (MoE), \citet{li2025find} applied \textit{Magnitude Analysis} on router activations and proposed two pathway metrics---similarity and consistency. These metrics monitor how routing patterns evolve from random fluctuations to stable structures, serving as a precise indicator to determine the onset of grokking and enabling optimal early stopping.
  
  Beyond grokking, similar monitoring strategies are applied to the emergence of \textit{In-Context Learning (ICL)}. 
  \citet{hoogland2402developmental} utilized the \textit{Local Learning Coefficient (LLC)} from singular learning theory to quantify the geometry of the loss landscape. They observed that plateaus in the \textit{LLC} curve distinctively mark developmental stages (e.g., from bigram statistics to induction heads), allowing researchers to determine when a model has completed a specific structural transformation.
  Furthermore, \citet{minegishi2025context} extended this to In-Context Meta-Learning, developing circuit-specific metrics such as \textit{label attention scores}. By monitoring the shift in these metrics, they identified that models progress through multiple distinct phases (Non-Context $\to$ Semi-Context $\to$ Full-Context), providing a granular ``progress bar'' for the model's acquisition of meta-learning capabilities that is invisible to standard loss evaluation.

\subsubsection{Efficient Inference}
\label{sec:efficient_inference}
\begin{tcolorbox}[takeawaysbox, title={Summary of Application Paradigms}]
  MI facilitates the deployment and acceleration of LLMs---resulting in superior inference efficiency---by identifying and exploiting structural and functional redundancies. This is primarily achieved through two key paradigms:
  \begin{enumerate}[leftmargin=*, label=\arabic*)]
      \item \textbf{Selective Computation via Saliency Detection:} 
      This paradigm reduces computational overhead by localizing ``dispensable'' components. 
      At the \textit{data level}, MI helps identify redundant tokens or KV cache entries through \textit{Magnitude Analysis} (\S\ref{sec:magtitude_analysis}), \textit{Gradient Detection} (\S\ref{sec:gradient_detection}), and \textit{Circuit Discovery} (\S\ref{sec:circuit_discovery}), enabling the pinpointing of tokens or heads with minimal importance.
      At the \textit{model level}, importance metrics based on \textit{Magnitude Analysis} (\S\ref{sec:magtitude_analysis}) enable the dynamic skipping of redundant layers or Mixture-of-Experts (MoE) experts, facilitating ``on-demand'' computation.
      \item \textbf{Layer-Specific Adaptive Quantization:} 
      Rather than applying uniform bit-widths, this paradigm leverages mechanistic insights, including \textit{Magnitude Analysis} (\S\ref{sec:magtitude_analysis}), \textit{Gradient Detection} (\S\ref{sec:gradient_detection}), and \textit{Vocabulary Projection} (\S\ref{sec:vocab_project}) to assess the quantization sensitivity of different layers. 
      By allocating higher precision to ``irreplaceable'' layers and applying aggressive compression to more robust ones, models achieve superior memory-accuracy trade-offs, tailored to diverse hardware constraints.
  \end{enumerate}
  \end{tcolorbox}
  \paragraph{1) Selective Computation via Saliency Detection}
  The core premise of selective computation is that not all architectural or data components contribute equally to the final output. 
  MI provides a principled tool to quantify such contributions and to prune redundant components accordingly.
  \begin{itemize}[leftmargin=*]
  \item \textbf{Data Level}: 
  Researchers have developed advanced token- and KV-cache-level pruning strategies that leverage \textit{Magnitude Analysis} and \textit{Gradient Detection} to effectively identify and remove unimportant tokens.
  By leveraging \textit{Magnitude Analysis} to identify tokens with minimal contribution to the reasoning process in CoT sequences, \textit{TokenSkip}~\citep{xia-etal-2025-tokenskip} selectively skips these tokens, achieving substantial compression with negligible performance degradation.
  \citet{lei2025generictokencompressionmultimodal} explored explanation-driven token compression for multimodal LLMs, where \textit{Gradient Detection} is used to map attention patterns to explanation outcomes, enabling the effective pruning of visual tokens during the input stage.
  For KV cache-level pruning, \textit{FitPrune}~\citep{ye2025fit} and \textit{ZipCache}~\citep{he2024zipcache} employed \textit{Magnitude Analysis} saliency metrics to identify and retain critical KV states.
  \citet{guo-etal-2024-attention} introduced \textit{Value-Aware Token Pruning (VATP)}, which applied \textit{Magnitude Analysis} to attention scores and the L1 norm of value attention vectors to identify crucial tokens.
  Moving beyond token-wise pruning, \textit{Circuit Discovery} techniques have been applied to identify “Retrieval Heads” that are essential for long-context tasks, enabling non-critical heads to operate with a fixed-length KV cache \citep{tang2024razorattention,xiong2024uncomp,xiong2025parallelcomp,xiao2024duoattention}.
  
  \item \textbf{Model Level}: 
  MI-guided metrics enable the skipping of entire architectural blocks, such as redundant layers, MoE experts, or neurons, thereby facilitating inference acceleration with minimal impact on model performance.
  \citet{men-etal-2025-shortgpt} introduced “Block Influence” (BI), a similarity metric based on \textit{Magnitude Analysis} that compares the input and output of each layer. This technique effectively removes layers with minimal contribution to the representation space.
  Dynamic bypassing methods, such as \textit{GateSkip} \citep{laitenberger2025layerswhenlearningskip} and \textit{LayerSkip} \citep{elhoushi-etal-2024-layerskip}, employ learnable residual gates to skip layers during inference, also based on Magnitude Analysis. 
  Similarly, \textit{HadSkip} \citep{wang-etal-2023-hadskip} and \textit{SBERT} \citep{shelke2024towards} models leverage Magnitude Analysis to facilitate effective layer skipping.
  In MoE architectures, \citet{lu2024not} skipped unimportant experts during inference based on the \textit{Magnitude Analysis} of router scores.
  \citet{su2025unveiling} further identified \textit{Super Experts} by analyzing the \textit{Magnitude Analysis} of experts' output activations, showing that these experts are essential for logical reasoning and that pruning them leads to catastrophic performance degradation.
  Finally, by localizing specialized multilingual neurons \citep{liu2024unraveling} and language-specific sub-networks \citep{tan-etal-2024-neuron} through \textit{Magnitude Analysis} on their activations, LLMs can activate only the sub-circuits necessary for the specific task at hand.
  \end{itemize}
  
  \paragraph{2) Layer-Specific Adaptive Quantization}
  While standard quantization applies a uniform bit-width across all parameters, MI-driven research promotes mixed-precision quantization based on layer-wise ``functional saliency.''
  Many of these metrics are based on \textit{Magnitude Analysis} to identify sensitive layers.
  \citet{dumitru2024layer} proposed a pragmatic approach to measure layer importance by examining shifts in the embedding space or the presence of weight outliers, assigning higher bit-precision to layers that caused larger representational shifts.
  Similarly, \citet{zhang2025towards} introduced \textit{SensiBoost} and \textit{KurtBoost}, which used activation sensitivity and weight distribution kurtosis to identify layers that were "hard-to-quantize," allocating them more memory budget.
  \textit{LieQ}~\citep{xiao2025exploring} further uncovered a strong correlation between training-induced energy concentration and representational compactness, providing a geometry-driven sensitivity proxy for automatic bit-width allocation.
  Beyond static analysis, \textit{Mix-QViT}~\citep{ranjan2025mix} employed \textit{Layer-wise Relevance Propagation (LRP)}—a form of \textit{Gradient Detection}—to assess the contribution of each layer to the final classification, thereby guiding mixed-precision quantization in vision transformers.
  \textit{LSAQ} \citep{zeng2024lsaq} adaptively adjusted quantization strategies in real-time by applying \textit{Vocab Projection} to obtain the vocab distribution for each layer.
  It then calculated the Jaccard similarity between these distributions to identify sensitive layers, ensuring that they maintained high precision while more robust layers were aggressively compressed to meet the resource constraints of edge devices.

\newpage

\section{Discussion and Challenges}
\label{app:challenges}
While the actionable applications discussed in previous sections demonstrate the encouraging potential of MI, it is crucial to recognize that it is not a universal solution. There remain important challenges and boundary conditions that may limit its future scalability, reliability, and broader practical impact. We discuss these fundamental points below.\looseness=-1

\paragraph{Scalability Constraints and Granularity} MI remains difficult to scale beyond low-level components \citep{kharlapenko2025scaling,nikankin2025same}. While individual neurons or learned features are increasingly well-characterized \citep{duan2025unveiling, bricken2023monosemanticity}, identifying higher-level computational structures still relies heavily on manual inspection~\citep{he2024dictionary,Markscircuts_iclr2025,yao2024circuits,lindsey2025biology,nguyen2025challenges}. Although recent work has made progress toward automation~\citep{conmy2023automated,hanna2024have}, current methods often require substantial human intervention and do not robustly generalize across prompts, tasks, or models~\citep{prakash2024fine,hanna-etal-2025-circuit,li2025find}. Crucially, as model sizes scale beyond 100B parameters, the computational cost of fine-grained causal localization grows prohibitively, forcing a fundamental trade-off between localization granularity and computational feasibility. Prominent approaches such as SAEs rely on training surrogate models, introducing costs that grow with model size and feature dimensionality. Precisely attributing behavior to individual components would in principle require exhaustive interventions, but modern LLM scale renders such causal tracing computationally infeasible \citep{zhang2023towards,hanna2024have}. As a result, most analyses \citep{nanda2023attribution,syedetal2024attribution,yumaagnitude_emnlp2024,ameisen2025circuit} operate at coarser granularities or rely on heuristic approximations.

\paragraph{Distributed Mechanisms vs. Localized Sparsity} 
Current mechanistic analyses often face a fundamental trade-off between sparsity and completeness of representation~\citep{gao2025weight,pach2025sparse}. 
Many interpretability methods aim to force the model's internal representations into a small set of monosemantic components to make interpretation tractable. However, aggressively enforcing sparsity may prune or obscure components that are genuinely part of the true mechanism but do not fit a sparse pattern. This leads to a tension: methods that induce sparsity can improve interpretability but risk overlooking distributed or ``inactive'' subcomponents of genuine mechanisms. Acknowledging scenarios where localization may be fundamentally limited, and developing metrics that balance sparsity and mechanistic completeness, remains an open challenge.

\paragraph{Intervention Robustness and Side Effects} 
Interventions informed by MI, such as model editing or steering, often lack robustness and predictability~\citep{yin2024lofit,wang2025beyond}. Changes intended to modify a specific behavior can introduce unintended side effects on other tasks or domains, raising concerns about generalization and reliability ~\citep{jiang2024interpretable,zhang2024cofitune,zhang2025find,hsueh2024editing,xu2025easyedit2,da2025steering,braun2025beyond,zhang2025reinforcementfinetuningenablesmllms}. For instance, \citet{yuUnderstandingMitigatingGender2025a} demonstrate that modifying a very small number of neurons can lead to substantial degradation in overall language performance. The need for accurate target localization and steering methods that avoid collateral behavioral disruption remains a central technical challenge.

\input{Tables/evaluation_protocal}

\paragraph{Evaluation Protocols for Actionable MI}
A critical gap exacerbating these challenges is the lack of robust evaluation frameworks to assess the faithfulness of localization and explanation methods \citep{miller2024transformer}. Although some benchmarks~\citep{mib2025,parrack2025benchmarking,nguyen1786probing,wu2025axbench,karvonen2025saebench} have been proposed, there remains no consensus on metrics that can determine whether an identified component truly corresponds to the underlying causal mechanism. In the absence of reliable ground truth at the mechanism level, rigorous validation and comparison of MI methods is inherently challenging. 

To address the need for standardized actionable-MI evaluation, we preliminarily propose a minimal evaluation framework (as shown in Table~\ref{tab:evaluation_framework}) that covers three distinct task settings: math reasoning, safety, and knowledge editing. For each setting, we explicitly define (1) the features to localize or steer, (2) the primary evaluation metric, (3) essential side-effect checks, and (4) representative benchmark datasets.

This framework facilitates rigorous, multi-faceted evaluation. For example, to evaluate a localizing method targeting reflection features in mathematical reasoning, one might first identify candidate features and then apply targeted ablation. The resulting change in reflection token frequency on benchmarks like AIME25 can then serve as a quantitative measure of causal faithfulness. Similarly, to rigorously compare steering interventions, one can apply different steering methods to the exact same localized features and measure differences in task accuracy or behavioral shifts. For instance, researchers can directly compare the efficacy of (1) amplifying a reflection feature via \textit{Amplitude Manipulation}, (2) injecting it as a steering vector via \textit{Vector Arithmetic}, and (3) performing localized weight updates via \textit{Targeted Optimization}.

Furthermore, measuring unintended consequences is a first-class requirement. For each task setting, we design dedicated side-effect checks to evaluate the collateral impact of different steering methods:

\begin{itemize}[leftmargin=*]
\item \textbf{Math Reasoning:} Interventions aimed at enhancing reflection or arithmetic features should not degrade performance in other domains. Therefore, after steering a reflection-related feature, the model must also be evaluated on general knowledge and safety benchmarks to ensure its broader capabilities remain intact.
\item \textbf{Safety:} The objective is not only to increase the refusal rate on harmful prompts (measured by benchmarks such as SORRY-Bench~\citep{Xie2025}), but also to avoid over-refusal---that is, rejecting benign or harmless prompts. This represents a critical trade-off, as overly aggressive safety interventions may significantly harm the model's general helpfulness. Benchmarks such as OR-Bench~\citep{Cui2024} are specifically designed to quantify this negative side effect.
\item \textbf{Knowledge Editing:} For knowledge editing, \textit{locality} is the primary side-effect metric. It measures whether editing a specific fact (e.g., modifying a president's name) unintentionally alters unrelated knowledge in the model~\citep{Zhang2024knowledgeedit}. \textit{Fluency} is additionally essential, as it evaluates whether the model's generative quality degrades post-edit, such as by becoming repetitive or producing ungrammatical outputs.
\end{itemize}

\section{Future Directions}
\label{sec:future_directions}
Looking forward, several directions appear particularly promising for advancing MI.

\paragraph{Broadening the Architectural Scope}
While our formalization in \S\ref{sec:core_objects} is primarily centered on decoder-only Transformer LLMs, an important future direction is to extend the ``Locate, Steer, and Improve'' framework to broader model architectures. Although our pipeline is defined in the decoder-only setting, many of the localizing and steering methodologies we categorize are general in spirit and can, in principle, be transferred to other architectures once their corresponding interpretable objects are identified. For example, compared with decoder-only dense models, MoE architectures introduce an additional routing mechanism, making router states, routing activations, routing decisions, and expert activations natural functional objects beyond standard neural activations \citep{shazeer2017outrageously,fedus2022switch,zhou2022expertchoice,wang2025semanticrouting}. This suggests that localization tools developed for dense models, such as \textit{Magnitude Analysis}, can be extended from ordinary neural activations to router activations, in order to identify influential routes or experts and subsequently steer behavior through interventions on the corresponding expert modules \citep{wang2025two,wang2025semanticrouting,routers2024visionmoe}. Similarly, MLLMs augment text-only LLMs with vision encoders and cross-modal interaction modules \citep{liu2023visual,li2023blip2,alayrac2022flamingo,lin2025survey}. Their interpretable objects therefore go beyond language-side activations to include visual token representations, image-patch attention maps, cross-modal attention patterns, and modality-level information-flow pathways, which can likewise be analyzed using existing tools in our taxonomy. For instance, \textit{Magnitude Analysis} can be applied to visual or multimodal activations, including those in the vision encoder and multimodal attention stack, while \textit{Vocabulary Projection} can be adapted to inspect how visual representations align with linguistic concepts \citep{neo2024towards,yu2024understanding,bi2025unveiling,zhang2025crossmodal}. At the same time, fully realizing actionable MI in these architectures will likely require more than direct adaptation: future work should develop architecture-specific localizing and steering techniques that explicitly account for distinctive structural properties such as sparse routing in MoE models and tightly entangled cross-modal representations in MLLMs \citep{lin2025survey}.

\paragraph{Integration with Cognitive Science} 
A key priority for mechanistic interpretability is to move from isolated, low-level analyses toward integrated, system-level explanations. Most existing MI work focuses on task-specific and localized mechanisms, such as knowledge neurons, safety-related neurons, arithmetic heads, or specific task circuits~\citep{yao2024knowledge,chen2024incontext,xiao2025taskcircuit,wang2024bpo,chen2026residual,liang2026training,zhang2024interpreting,xiong2026expressionsyntaxinformationbottleneck,xiong2026mmformalizer,gurgurov2025languagearithmeticssystematiclanguage,li2025safety}. While informative, these approaches offer limited insight into how models organize computation more broadly~\citep{zhao2024explainability}. 
In contrast, cognitive science conceptualizes cognition through higher-order organizations. This includes broad processing paradigms like System 1 and System 2 reasoning~\citep{li2025system}, as well as distinct functional subsystems governing attention, memory, language, and executive control~\citep{morgan1927introduction, gruber2004executive,gruszka2010handbook,zhang2019cognitive}.
Future research should draw explicit parallels between MI-discovered mechanisms and these established cognitive architectures. For instance, exploring whether specific localized circuits function analogously to working memory or attention control, could reveal if LLM internal structures exhibit organizational principles analogous to human cognitive systems~\citep{geiger2025causal}. Furthermore, this integration can foster a bidirectional synergy: MI findings might inform cognitive theories of human reasoning and language processing, while established cognitive frameworks provide a robust blueprint for understanding and evaluating LLMs.

\paragraph{Theoretical Foundations} 
In parallel, stronger theoretical foundations are needed. Connecting internal representations to principles from cognitive science~\citep{davies2024cognitive, wulff2025advancing,zhang2025question,ren2025large} or information theory~\citep{conklin2024representations} may help unify disparate MI findings and reduce reliance on ad-hoc interpretations. A principled framework could also clarify what kinds of internal structures should be expected in large-scale models and why~\citep{kendiukhov2025review}.

\paragraph{From Interpretation to Interpretable Design} 
Finally, an emerging direction is the progression from interpretation to intervention and, ultimately, model design. Insights from MI are increasingly used not only to explain behavior, but also to edit, steer, or modularize models. This direction connects naturally to earlier work on intrinsically interpretable models, such as Concept Bottleneck Models~\citep{ismail2024concept,sunconcept2024,shang2024understanding,shang2024incremental,tan2024interpreting,hu2025editableconceptbottleneckmodels,zhao2025partially} and Weight-sparse transformers~\citep{gao2025weight}, which enforce transparency through architectural constraints. However, despite their interpretability benefits, such models typically underperform black-box architectures on large-scale, complex tasks~\citep{srivastava2024vlg}. Looking forward, a key challenge is to bridge this gap by designing interpretable backbone architectures that can serve as viable alternatives to transformers, achieving interpretability by construction while maintaining performance comparable to state-of-the-art black-box models. In this sense, interpretability-informed design may move beyond post-hoc analysis toward fundamentally more controllable, customizable, and transparent model architectures.

\section{Conclusion}
\label{sec:conclusion}
In this survey, we systematically reframe MI from a predominantly observational endeavor into a practical, actionable paradigm. By organizing existing methods around the unified pipeline of ``Locate, Steer, and Improve'', we clarify how interpretable objects can be precisely localized, causally manipulated, and ultimately leveraged to enhance alignment, capability, and efficiency in LLMs.
Our analysis highlights that many recent advances---ranging from safety and persona alignment, to knowledge editing, and further to sparse fine-tuning---are most effective when grounded in explicit mechanistic intervention. We further discuss key challenges and future directions in \S\ref{app:challenges}, with the goal of providing a coherent foundation for future research that tightly integrates interpretability, intervention, and model design. Ultimately, we hope this perspective will accelerate the transition toward more powerful, transparent, and reliable LLMs.

\section*{Limitation}
\label{sec:limitation}
This survey focuses on MI for dense LLMs and does not systematically cover methods specific to other architectures and modalities. In particular, Mixture-of-Experts (MoE) models introduce routing mechanisms and sparsely activated experts, while vision–language models and vision-only models rely on modality-specific representations and architectural components that pose distinct interpretability challenges. Nevertheless, many of the methods discussed in this work are conceptually general and, with appropriate adaptation, can be applied to MoE models and multimodal architectures, for example by operating on expert-level activations or modality-specific residual streams. A comprehensive and systematic treatment of these architectures is therefore left to future work.

In addition, the field currently lacks unified benchmarks or standardized evaluation protocols for localization methods, making it difficult to rigorously compare approaches or to assess whether the identified model components are causally optimal. This limitation also affects downstream applications, where interventions often rely on a single localization method without formal guarantees. Some works partially mitigate this issue by combining multiple localization techniques and examining whether they converge on similar model components, but developing principled and reproducible evaluation frameworks remains an open challenge.

\newpage
\onecolumn
\appendix

\newpage
\section{Summary of Surveyed Papers}
\label{app:paper_list}

\definecolor{colorObj}{RGB}{0, 51, 102} 
\newcommand{\ObjStyle}[1]{\textcolor{colorObj}{\textbf{\texttt{\footnotesize #1}}}} 

\def\ObjToken{\ObjStyle{Token Embedding}}
\def\ObjRes{\ObjStyle{Residual Stream}}
\def\ObjAttn{\ObjStyle{MHA}}
\def\ObjFFN{\ObjStyle{FFN}}
\def\ObjNeuron{\ObjStyle{Neuron}}
\def\ObjSAE{\ObjStyle{SAE Feature}}

\definecolor{colorLoc}{RGB}{0, 100, 0}
\newcommand{\LocStyle}[1]{\textcolor{colorLoc}{\textsf{\footnotesize #1}}}
\def\LocMag{\LocStyle{Magnitude Analysis}}
\def\LocCausal{\LocStyle{Causal Attribution}}
\def\LocGrad{\LocStyle{Gradient Detection}}
\def\LocProb{\LocStyle{Probing}}
\def\LocVocab{\LocStyle{Vocab Projection}}
\def\LocCirc{\LocStyle{Circuit Discovery}}

\definecolor{colorStr}{RGB}{139, 0, 0}
\newcommand{\StrStyle}[1]{\textcolor{colorStr}{\textit{\footnotesize #1}}}
\def\StrAmp{\StrStyle{Amplitude Manipulation}}
\def\StrOpt{\StrStyle{Targeted Optimization}}
\def\StrVec{\StrStyle{Vector Arithmetic}}

\def\None{\textcolor{gray}{-}}

\newcolumntype{L}[1]{>{\raggedright\arraybackslash}p{#1}}
\newcolumntype{C}[1]{>{\centering\arraybackslash}p{#1}}

\hypertarget{paperlistall}{}
\begin{small} 
\begin{longtable}{L{3.6cm} C{2.9cm} C{2.9cm} C{3.2cm} C{1.0cm} C{0.5cm} C{0.5cm}}
\caption{Summary of Surveyed Papers. We annotate each paper with tags for its Core Interpretable Objects (\S\ref{sec:core_objects}), Localizing Methods (\S\ref{sec:localize_methods}), and Steering Methods (\S\ref{sec:steer_methods}). For studies employing multiple objects or localizing/steering methods, we annotate the primary tag. The symbol ``-'' in the Steering Method column denotes works that apply localized mechanistic insights directly for analysis or monitoring, without employing active intervention techniques. The surveyed papers were collected through searches on major academic platforms, including \textit{arXiv}, \textit{OpenReview}, \textit{ACM Library}, and \textit{Semantic Scholar}, using keywords such as ``Mechanistic Interpretability'', ``Sparse Autoencoder'', ``Neuron'', and ``Circuit'', and were categorized based on their \textit{primary} interpretable object, localizing method, and steering method.}\label{tab:paper_list_all} \\
\toprule
\textbf{Paper} & \multicolumn{1}{c}{\textbf{Object}} & \multicolumn{1}{c}{\textbf{Localizing Method}} & \multicolumn{1}{c}{\textbf{Steering Method}} & \textbf{Venue} & \textbf{Year} & \textbf{Link} \\
\midrule
\endfirsthead

\toprule
\textbf{Paper} & \multicolumn{1}{c}{\textbf{Object}} & \multicolumn{1}{c}{\textbf{Localizing Method}} & \multicolumn{1}{c}{\textbf{Steering Method}} & \textbf{Venue} & \textbf{Year} & \textbf{Link} \\
\midrule
\endhead

\bottomrule
\endfoot

\rowcolor{gray!15} 
\multicolumn{7}{c}{\textbf{\textit{Safety and Reliability (Improve Alignment)}}} \\
\midrule
\citeauthor{zhou2025on} & \ObjAttn & \LocCausal & \StrAmp & ICLR & 2025 & \href{https://openreview.net/forum?id=h0Ak8A5yqw}{Link} \\

\citeauthor{huang-etal-2025-pierce} & \ObjAttn & \LocCirc & \StrOpt & EMNLP & 2025 & \href{https://aclanthology.org/2025.emnlp-main.781/}{Link} \\

\citeauthor{jiang2024refine} & \ObjAttn & \LocCausal & \StrOpt & ArXiv & 2024 & \href{https://arxiv.org/abs/2406.12227}{Link} \\

\citeauthor{chen2025towards} & \ObjNeuron & \LocCausal & \StrAmp & ArXiv & 2025 & \href{https://openreview.net/forum?id=1NkrxqY4jK}{Link} \\

\citeauthor{suauWhisperingExpertsNeural2024} & \ObjNeuron & \LocMag & \StrAmp & ICML & 2024 & \href{https://openreview.net/forum?id=2P6GVfSrfZ}{Link} \\

\citeauthor{gao2025hneuronsexistenceimpactorigin} & \ObjNeuron & \LocMag & \StrAmp & ArXiv & 2025 & \href{https://arxiv.org/abs/2512.01797}{Link} \\

\citeauthor{zhao2025understanding} & \ObjNeuron & \LocMag & \StrOpt & ICLR & 2025 & \href{https://openreview.net/forum?id=yR47RmND1m}{Link} \\

\citeauthor{liPrecisionKnowledgeEditing2024} & \ObjNeuron & \LocMag & \StrOpt & ArXiv & 2025 & \href{https://arxiv.org/abs/2410.03772}{Link} \\

\citeauthor{templeton2024scaling} & \ObjSAE & \LocMag  & \StrAmp & Blog & 2024 & \href{https://transformer-circuits.pub/2024/scaling-monosemanticity/index.html}{Link} \\

\citeauthor{goyalBreakingBadTokens2025} & \ObjSAE & \LocMag  & \StrAmp & EMNLP & 2025 & \href{https://aclanthology.org/2025.emnlp-main.641/}{Link} \\

\citeauthor{yeo-etal-2025-understanding} & \ObjSAE & \LocMag  & \StrAmp & EMNLP & 2025 & \href{https://aclanthology.org/2025.findings-emnlp.338/}{Link} \\

\citeauthor{li2025training} & \ObjSAE & \LocMag & \StrVec & ArXiv & 2025 & \href{https://arxiv.org/abs/2506.07691}{Link}
\\ 

\citeauthor{weng2025safe} & \ObjSAE & \LocMag & \StrAmp & ArXiv & 2025 & \href{https://arxiv.org/abs/2509.18127}{Link} \\

\citeauthor{wu2025axbench} & \ObjSAE & \LocMag & \StrVec & ICML & 2025 & \href{https://openreview.net/forum?id=K2CckZjNy0}{Link} \\

\citeauthor{he2025saif} & \ObjSAE & \LocMag & \StrVec & ArXiv & 2025 & \href{https://arxiv.org/abs/2502.11356}{Link} \\

\citeauthor{li2025safety} & \ObjRes & \LocCausal & \StrOpt & ICLR & 2025 & \href{https://openreview.net/forum?id=kUH1yPMAn7}{Link} \\

\citeauthor{hedstromsteer} & \ObjRes & \LocProb & \StrVec & ICML & 2025 & \href{https://arxiv.org/abs/2510.13290}{Link} \\

\citeauthor{leeMechanisticUnderstandingAlignment2024} & \ObjRes & \LocProb & \StrOpt & ICML & 2024 & \href{https://proceedings.mlr.press/v235/lee24a.html}{Link} \\

\citeauthor{arditi2024refusal} & \ObjRes & \LocCausal & \StrVec & NeurIPS & 2024 & \href{https://openreview.net/forum?id=pH3XAQME6c}{Link} \\

\citeauthor{zhao2025llms} & \ObjRes & \LocCausal & \StrVec & NeurIPS & 2025 & \href{https://openreview.net/forum?id=zLkpt30ngy}{Link} \\

\citeauthor{yin2025refusalfallscliffsafety} & \ObjRes & \LocProb & \StrVec & ArXiv & 2025 & \href{https://arxiv.org/abs/2510.06036}{Link} \\

\citeauthor{ball2024understandingjailbreaksuccessstudy} & \ObjRes & \LocCausal & \StrVec & ArXiv & 2024 & \href{https://arxiv.org/abs/2406.09289}{Link} \\

\citeauthor{wang2025surgical} & \ObjRes & \LocCausal & \StrVec & ICLR & 2025 & \href{https://openreview.net/forum?id=SCBn8MCLwc}{Link} \\

\citeauthor{wang2025refusal} & \ObjRes & \LocCausal & \StrVec & NeurIPS & 2025 & \href{https://openreview.net/forum?id=eWxKpdAdXH}{Link} \\

\citeauthor{ferreira2025truthfulfabricatedusingcausal} & \ObjRes & \LocCausal & \StrVec & ICML & 2025 & \href{https://arxiv.org/abs/2504.05294}{Link} \\

\citeauthor{huang2025internalcausalmechanismsrobustly} & \ObjRes & \LocCausal & \StrVec & ICML & 2025 & \href{https://openreview.net/forum?id=wGFEzfhFae}{Link} \\

\citeauthor{pan2025the} & \ObjRes & \LocCausal & \StrVec & ICML & 2025 & \href{https://openreview.net/forum?id=pH3XAQME6c}{Link} \\

\citeauthor{chuang2024dola} & \ObjRes & \LocVocab & \StrVec & ICLR & 2024 & \href{https://openreview.net/forum?id=Th6NyL07na}{Link} \\

\citeauthor{chen2024incontext} & \ObjRes & \LocVocab & \StrVec & ICML & 2024 & \href{https://openreview.net/forum?id=s3e8poX3kb}{Link} \\

\citeauthor{yan2026spurious} & \ObjRes & \LocVocab & \StrAmp & ArXiv & 2026 & \href{https://arxiv.org/abs/2601.11061}{Link} \\

\citeauthor{zhang-etal-2024-truthx} & \ObjRes & \LocProb & \StrVec & ACL & 2024 & \href{https://aclanthology.org/2024.acl-long.483/}{Link} \\

\citeauthor{orgad2025llms} & \ObjRes & \LocProb & \StrVec & ICLR & 2025 & \href{https://openreview.net/forum?id=KRnsX5Em3W}{Link} \\

\citeauthor{stolfo2025improving} & \ObjRes & \LocGrad & \StrVec & ICLR & 2025 & \href{https://openreview.net/forum?id=wozhdnRCtw}{Link} \\

\citeauthor{nguyen2025multi} & \ObjRes & \LocGrad & \StrVec & ACL & 2025 & \href{https://arxiv.org/abs/2502.12446}{Link} \\

\citeauthor{du2024tst} & \ObjToken & \LocGrad & \StrVec & ArXiv & 2025 & \href{https://arxiv.org/abs/2507.18043}{Link} \\

\rowcolor{gray!15} 
\multicolumn{7}{c}{\textbf{\textit{Fairness and Bias (Improve Alignment)}}} \\
\midrule

\citeauthor{vig2020gender} & \ObjAttn & \LocCausal & \StrAmp & NeurIPS & 2020 & \href{https://proceedings.neurips.cc/paper/2020/file/92650b2e92217715fe312e6fa7b90d82-Paper.pdf}{Link} \\

\citeauthor{chintamIdentifyingAdaptingTransformerComponents2023} & \ObjAttn & \LocCausal & \StrOpt & ACLWS & 2023 &
\href{https://aclanthology.org/2023.blackboxnlp-1.29/}{Link} \\

\citeauthor{wangEliminatingPositionBias2025a} & \ObjAttn & \LocMag & \StrAmp & ICLR & 2025 &
\href{https://openreview.net/forum?id=fvkElsJOsN}{Link} \\

\citeauthor{iclr2025_politicalprobe} & \ObjAttn & \LocProb & \StrVec & ICLR & 2025 &
\href{https://openreview.net/forum?id=rwqShzb9li}{Link}
\\

\citeauthor{diminoTracingPositionalBias2025} & \ObjAttn & \LocMag & \None & ICAIF & 2025 &
\href{https://arxiv.org/abs/2508.18427}{Link} \\

\citeauthor{chandnaDissectingBiasLLMs2025} & \ObjAttn & \LocMag & \StrAmp & TMLR & 2025 &
\href{https://openreview.net/forum?id=EpQ2CBJTjD}{Link} \\

\citeauthor{caiLocatingMitigatingGender2024} & \ObjFFN & \LocCausal & \StrOpt & ICIC & 2024 & \href{https://arxiv.org/abs/2403.14409}{Link} \\

\citeauthor{ahsanElucidatingMechanismsDemographic2025} & \ObjFFN & \LocCausal & \StrAmp & EMNLP & 2025 & \href{https://aclanthology.org/2025.findings-emnlp.789}{Link} \\

\citeauthor{liAnchoredAnswersUnravelling2025} & \ObjFFN & \LocVocab & \StrOpt & ACL & 2025 &
\href{https://aclanthology.org/2025.findings-acl.124/}{Link} \\

\citeauthor{yuUnderstandingMitigatingGender2025a} & \ObjNeuron & \LocCirc & \StrOpt & ArXiv & 2025 & \href{https://arxiv.org/abs/2501.14457}{Link} \\

\citeauthor{liuDevilNeuronsInterpreting2024} & \ObjNeuron & \LocGrad & \StrAmp & ICLR & 2024 &
\href{https://openreview.net/forum?id=SQGUDc9tC8}{Link} \\

\citeauthor{yuEntangledRepresentationsMechanistic2025} & \ObjRes & \LocCausal & \None & ArXiv & 2025 &
\href{https://arxiv.org/abs/2508.08879}{Link} \\

\citeauthor{guanMPFAligningDebiasing2025} & \ObjRes & \None & \StrAmp & ICML & 2025 & \href{https://arxiv.org/abs/2507.02595}{Link} \\

\citeauthor{yuMitigatePositionBias} & \ObjRes & \LocMag & \StrAmp & ACL & 2025 &
\href{https://aclanthology.org/2025.findings-acl.316/}{Link} \\

\citeauthor{raimondiAnalysingMoralBias2025} & \ObjRes & \LocCausal & \StrAmp & ArXiv & 2025 &
\href{https://arxiv.org/abs/2510.12229}{Link} \\

\rowcolor{gray!15} 
\multicolumn{7}{c}{\textbf{\textit{Persona and Role (Improve Alignment)}}} \\
\midrule

\citeauthor{su-etal-2025-understanding} & \ObjNeuron & \LocCausal & \StrAmp & EMNLP & 2025 & \href{https://aclanthology.org/2025.findings-emnlp.501/}{Link} \\

\citeauthor{deng2025neuron} & \ObjNeuron & \LocCausal & \StrAmp & ICLR & 2025 & \href{https://openreview.net/forum?id=LYHEY783Np}{Link} \\

\citeauthor{lai-etal-2024-style} & \ObjNeuron & \LocMag & \StrAmp & EMNLP & 2024 & \href{https://aclanthology.org/2024.emnlp-main.745}{Link}  \\

\citeauthor{chen2024from} & \ObjNeuron & \LocCausal & \StrOpt & ICML & 2024 & \href{https://openreview.net/forum?id=d2vONO90Rw}{Link} \\

\citeauthor{lu2026assistant} & \ObjRes & \LocMag & \StrVec & ArXiv & 2026 & \href{https://arxiv.org/abs/2601.10387}{Link} \\

\citeauthor{rimsky-etal-2024-steering} & \ObjRes & \LocCausal & \StrVec & ACL & 2024 & \href{https://aclanthology.org/2024.acl-long.828/}{Link} \\

\citeauthor{poterti-etal-2025-role} & \ObjRes & \LocCausal & \StrVec & EMNLP & 2025 & \href{https://aclanthology.org/2025.findings-emnlp.963/}{Link} \\

\citeauthor{chen2025persona} & \ObjRes & \LocCausal & \StrVec & ArXiv & 2025 & \href{https://arxiv.org/abs/2507.21509}{Link} \\

\citeauthor{handa2025personality} & \ObjRes & \LocCausal & \StrVec & NeurIPS & 2025 & \href{https://openreview.net/forum?id=TWbcIU0DBr}{Link} \\

\citeauthor{tak-etal-2025-mechanistic} & \ObjRes & \LocProb & \StrVec & ACL & 2025 & \href{https://aclanthology.org/2025.findings-acl.679/}{Link} \\

\citeauthor{yuan2025monolingual} & \ObjRes & \LocProb & \None & ArXiv & 2025 & \href{https://arxiv.org/abs/2508.02502}{Link} \\

\citeauthor{ju2025probing} & \ObjRes & \LocProb & \StrOpt & COLM & 2025 & \href{https://openreview.net/pdf?id=z9SbcYYP0M}{Link} \\

\citeauthor{karny2025neural} & \ObjRes & \LocCausal & \None & ArXiv & 2025 & \href{https://arxiv.org/abs/2511.00230}{Link} \\

\citeauthor{banayeeanzade2025psychological} & \ObjRes & \LocCausal & \StrVec & ArXiv & 2025 & \href{https://arxiv.org/abs/2510.04484}{Link} \\

\citeauthor{bas2025steering} & \ObjRes & \LocCausal& \StrVec & ArXiv & 2025 & \href{https://arxiv.org/abs/2511.18284}{Link}\\ 

\citeauthor{sun-etal-2025-personality} & \ObjRes & \LocCausal& \StrVec & EMNLP & 2025 & \href{https://aclanthology.org/2025.emnlp-main.1253/}{Link} \\ 

\citeauthor{pai2025billy} & \ObjRes & \LocCausal& \StrVec & ArXiv & 2025 & \href{https://arxiv.org/abs/2510.10157}{Link}  \\

\citeauthor{joshi2024personas} & \ObjRes & \LocProb & \None & EMNLP & 2024 & \href{https://aclanthology.org/2024.emnlp-main.364.pdf}{Link}  \\ 

\citeauthor{ghandeharioun2024whos} & \ObjRes & \LocCausal & \StrVec & NeurIPS & 2024 & \href{https://proceedings.neurips.cc/paper_files/paper/2024/file/e40d5118ee8f837729fa877add71c38f-Paper-Conference.pdf}{Link}  
\\

\rowcolor{gray!15} 
\multicolumn{7}{c}{\textbf{\textit{Multilingualism (Improve Capability)}}} \\
\midrule

\citeauthor{xie-etal-2021-importance} & \ObjNeuron & \LocMag & \StrAmp & ACL & 2021 & \href{https://aclanthology.org/2021.acl-long.445/}{Link} \\

\citeauthor{kojima-etal-2024-multilingual} & \ObjNeuron & \LocMag & \StrAmp & NAACL & 2024 & \href{https://aclanthology.org/2024.naacl-long.384/}{Link} \\
\citeauthor{tang-etal-2024-language} & \ObjNeuron & \LocMag & \StrAmp & ACL & 2024 & \href{https://aclanthology.org/2024.acl-long.309/}{Link} \\

\citeauthor{zhao-etal-2024-multilingual} & \ObjNeuron & \LocMag & \StrAmp & NeurIPS & 2024 & \href{https://proceedings.neurips.cc/paper_files/paper/2024/file/1bd359b32ab8b2a6bbafa1ed2856cf40-Paper-Conference.pdf}{Link} \\

\citeauthor{gurgurov2025languagearithmeticssystematiclanguage} & \ObjNeuron & \LocMag & \StrAmp & ArXiv & 2025 & \href{https://arxiv.org/abs/2507.22608}{Link} \\

\citeauthor{liu-etal-2025-relation-specific} & \ObjNeuron & \LocMag & \StrAmp & EMNLP & 2025 & \href{https://aclanthology.org/2025.emnlp-main.52/}{Link} \\

\citeauthor{jing-etal-2025-lingualens} & \ObjNeuron & \LocMag & \StrAmp & EMNLP & 2025 & \href{https://aclanthology.org/2025.emnlp-main.1433/}{Link} \\

\citeauthor{andrylie2025sparseautoencoderscapturelanguagespecific} & \ObjSAE & \LocMag & \StrAmp & ArXiv & 2025 & \href{https://arxiv.org/abs/2507.11230}{Link} \\

\citeauthor{brinkmann-etal-2025-large} & \ObjSAE & \LocMag & \StrAmp & NAACL & 2025 & \href{https://aclanthology.org/2025.naacl-long.312/}{Link} \\

\citeauthor{libovicky-etal-2020-language} & \ObjRes & \LocProb & - & EMNLP & 2020 & \href{https://aclanthology.org/2020.findings-emnlp.150/}{Link} \\

\citeauthor{chi-etal-2023-cross} & \ObjRes & \None & \StrVec & ACL & 2023 & \href{https://aclanthology.org/2023.findings-acl.796/}{Link} \\

\citeauthor{philippy2023identifying} & \ObjRes & \LocMag & \StrVec & ACL & 2023 & \href{https://aclanthology.org/2023.sigtyp-1.3/}{Link} \\

\citeauthor{wendler2024llamas} & \ObjRes & \LocVocab & \StrVec & ACL & 2024 & \href{https://aclanthology.org/2024.acl-long.820/}{Link} \\

\citeauthor{mousi2024exploring} & \ObjRes & \LocMag & \StrVec & ACL & 2024 & \href{https://aclanthology.org/2024.acl-long.344/}{Link} \\

\citeauthor{hinck-etal-2024-llava} & \ObjRes & \LocProb & \StrVec & EMNLP & 2024 & \href{https://aclanthology.org/2024.findings-emnlp.783/}{Link} \\

\citeauthor{zhang-etal-2025-shifcon} & \ObjRes & \LocMag  & \StrVec & ACL & 2025 & \href{https://aclanthology.org/2025.acl-long.239/}{Link} \\

\citeauthor{wang-etal-2025-lost-multilinguality} & \ObjRes & \LocVocab & \StrVec & ACL & 2025 & \href{https://aclanthology.org/2025.acl-long.253/}{Link} \\

\citeauthor{wu2025the} & \ObjRes & \LocVocab & - & ICLR & 2025 & \href{https://openreview.net/forum?id=FrFQpAgnGE}{Link} \\

\citeauthor{wang-etal-2025-language-mixing} & \ObjRes & \LocVocab & \StrVec & EMNLP & 2025 & \href{https://aclanthology.org/2025.emnlp-main.132/}{Link} \\

\citeauthor{nie-etal-2025-mechanistic} & \ObjRes & \LocVocab & \StrVec & EMNLP & 2025 & \href{https://aclanthology.org/2025.findings-emnlp.37/}{Link} \\

\citeauthor{liu-etal-2025-tracing} & \ObjRes & \LocVocab & \StrVec & EMNLP & 2025 & \href{https://aclanthology.org/2025.findings-emnlp.113/}{Link}
\\

\rowcolor{gray!15} 
\multicolumn{7}{c}{\textbf{\textit{Knowledge Management (Improve Capability)}}} \\
\midrule

\citeauthor{yang2025fine} & \ObjFFN & \LocGrad & \StrOpt & ICLR & 2026 & \href{https://arxiv.org/abs/2509.22072}{Link} \\
\citeauthor{yang2025ace} & \ObjFFN & \LocCausal & \StrOpt & ICLR & 2026 & \href{https://arxiv.org/abs/2510.07896}{Link} \\
\citeauthor{wang2025lokilowdamageknowledgeimplanting} & \ObjFFN & \LocCausal & \StrOpt & AAAI & 2026 & \href{https://arxiv.org/abs/2505.22120}{Link} \\
\citeauthor{meng2022ccs}               & \ObjFFN   & \LocCausal & \StrOpt & NeurIPS & 2022 & \href{https://openreview.net/forum?id=-h6WAS6eE4}{Link} \\
\citeauthor{meng2023massediting}       & \ObjFFN   & \LocCausal & \StrOpt & ICLR    & 2023 & \href{https://openreview.net/forum?id=MkbcAHIYgyS}{Link} \\

\citeauthor{lai2025jola}               & \ObjAttn  & \LocMag    & \StrOpt & ICML    & 2025 & \href{https://openreview.net/forum?id=Lllg9YjAFX}{Link} \\
\citeauthor{li2025taming}              & \ObjAttn  & \LocMag    & \StrAmp & ICML    & 2025 & \href{https://openreview.net/forum?id=0cEZyhHEks}{Link} \\
\citeauthor{jin2025massivevalues}      & \ObjAttn  & \LocMag    & \StrAmp & ICML    & 2025 & \href{https://openreview.net/forum?id=1SMcxxQiSL}{Link} \\
\citeauthor{jin2024ph3}                & \ObjAttn  & \LocCausal   & \StrAmp & ACL     & 2024 & \href{https://doi.org/10.18653/v1/2024.findings-acl.70}{Link} \\
\citeauthor{lvcausaliarXiv2024}        & \ObjAttn  & \LocCausal & \StrAmp & ArXiv & 2024 & \href{https://arxiv.org/abs/2403.19521}{Link} \\
\citeauthor{li2026attributing} & \ObjAttn & \LocMag & \StrAmp & ICLR & 2026 & \href{https://arxiv.org/abs/2505.16415}{Link} \\
\citeauthor{niu-etal-2025-llama}           & \ObjAttn  & \LocCausal & \StrAmp & ACL     & 2025 & \href{https://aclanthology.org/2025.acl-long.791/}{Link} \\
\citeauthor{emnlp2025_headprobe}       & \ObjAttn  & \LocProb   & \StrOpt & EMNLP    & 2025 & \href{https://aclanthology.org/2025.emnlp-main.1450/}{Link} \\

\citeauthor{yadav2023tiesmerging}      & \ObjFFN~\&~\ObjAttn  & \LocMag   & \StrVec & NeurIPS & 2023 & \href{https://proceedings.neurips.cc/paper_files/paper/2023/hash/1644c9af28ab7916874f6fd6228a9bcf-Abstract-Conference.html}{Link} \\
\citeauthor{yumaagnitude_emnlp2024}    & \ObjFFN~\&~\ObjAttn  & \LocMag   & \StrAmp & EMNLP & 2024 & \href{https://aclanthology.org/2024.emnlp-main.191.pdf}{Link} \\
\citeauthor{zhang2024cofitune}         & \ObjFFN~\&~\ObjAttn  & \LocMag   & \StrOpt & ACL     & 2024 & \href{https://doi.org/10.18653/v1/2024.findings-acl.445}{Link} \\
\citeauthor{chen2024querylocalization} & \ObjFFN~\&~\ObjAttn  & \LocMag   & \StrAmp & ICLR    & 2025 & \href{https://openreview.net/forum?id=tfyHbvFZ0K}{Link} \\
\citeauthor{gmt2025}                   & \ObjFFN~\&~\ObjAttn  & \LocMag   & \StrOpt & AAAI & 2025 & \href{https://doi.org/10.1609/aaai.v39i23.34621}{Link} \\
\citeauthor{muhamed2025geometry}       & \ObjFFN~\&~\ObjAttn  & \LocMag   & \None   & ICML    & 2025 & \href{https://openreview.net/forum?id=qKh7Aip3JC}{Link} \\
\citeauthor{yao2024circuits}           & \ObjFFN~\&~\ObjAttn   & \LocCirc   & \StrAmp & NeurIPS & 2024 & \href{https://proceedings.neurips.cc/paper_files/paper/2024/file/d6df31b1be98e04be48af8bedb95b499-Paper-Conference.pdf}{Link} \\
\citeauthor{du2024tst}                 & \ObjFFN~\&~\ObjAttn  & \LocProb  & \StrOpt & ArXiv   & 2024 & \href{https://doi.org/10.48550/arXiv.2410.04524}{Link} \\
\citeauthor{zhang2024linguistic}       & \ObjFFN~\&~\ObjAttn  & \LocGrad  & \StrOpt & ACL     & 2024 & \href{https://aclanthology.org/2024.acl-long.338/}{Link} \\
\citeauthor{liu2025sensmerging}        & \ObjFFN~\&~\ObjAttn  & \LocGrad  & \StrVec & ACL     & 2025 & \href{https://aclanthology.org/2025.findings-acl.984/}{Link} \\
\citeauthor{yao2025activation}        & \ObjFFN~\&~\ObjAttn   & \LocMag   & \StrVec & NeurIPS & 2025 & \href{https://openreview.net/pdf?id=ayzWTxb9ZD}{Link} \\
\citeauthor{geva-etal-2023-dissecting}         & \ObjFFN~\&~\ObjAttn   & \LocCausal   & \None & EMNLP & 2023 & \href{https://aclanthology.org/2023.emnlp-main.751.pdf}{Link} \\

\citeauthor{zhang2024lulafns}          & \ObjNeuron & \LocMag   & \StrOpt & COLING  & 2025 & \href{https://aclanthology.org/2025.coling-main.385/}{Link} \\
\citeauthor{chengrad_aaai2024}         & \ObjNeuron & \LocGrad  & \StrAmp & AAAI & 2024 & \href{https://ojs.aaai.org/index.php/AAAI/article/view/29735}{Link} \\
\citeauthor{ircan_neurips2024}         & \ObjNeuron & \LocGrad  & \StrAmp & NeurIPS  & 2024 & \href{http://papers.nips.cc/paper\_files/paper/2024/hash/08a9e28c96d016dd63903ab51cd085b0-Abstract-Conference.html}{Link} \\
\citeauthor{chen2024qrnca}             & \ObjNeuron & \LocGrad  & \StrAmp & AAAI    & 2025 & \href{https://doi.org/10.1609/aaai.v39i22.34529}{Link} \\
\citeauthor{kassem2025mneme}           & \ObjNeuron & \None     & \StrAmp & EMNLP   & 2025 & \href{https://aclanthology.org/2025.emnlp-main.1641/}{Link} \\

\citeauthor{muhamed2025dsg}            & \ObjSAE   & \LocMag   & \StrAmp & ICML    & 2025 & \href{https://openreview.net/pdf?id=8gFO7ebDLT}{Link} \\
\citeauthor{zhao2025steering} & \ObjSAE & \LocMag & \StrAmp & NAACL & 2025 & \href{https://aclanthology.org/2025.naacl-long.264.pdf}{Link} \\
\citeauthor{goyalBreakingBadTokens2025}& \ObjSAE   & \LocMag   & \StrAmp & EMNLP   & 2025 & \href{https://aclanthology.org/2025.emnlp-main.641/}{Link} \\
\citeauthor{Markscircuts_iclr2025}     & \ObjSAE   & \LocCirc   & \StrAmp & ICLR & 2025 & \href{https://arxiv.org/pdf/2403.19647}{Link} \\

\citeauthor{kangprobing_emnlp2023}         & \ObjRes   & \LocProb   & \None & EMNLP & 2023 & \href{https://aclanthology.org/2023.findings-emnlp.518.pdf}{Link} \\

\citeauthor{katz2024backwardlens}      & \ObjRes & \LocVocab & \StrOpt & EMNLP   & 2024 & \href{https://doi.org/10.18653/v1/2024.emnlp-main.142}{Link} \\

\citeauthor{wu2024reft}                & \ObjRes   & \LocCausal & \StrOpt & NeurIPS & 2024 & \href{https://proceedings.neurips.cc/paper_files/paper/2024/hash/75008a0fba53bf13b0bb3b7bff986e0e-Abstract-Conference.html}{Link} \\

\citeauthor{arxiv2410_knowledgeconflict}& \ObjRes   & \LocProb   & \None & ArXiv & 2024 & \href{https://arxiv.org/abs/2410.16090}{Link} \\

\citeauthor{juprobing_coling2024}         & \ObjRes   & \LocProb   & \None & COLING & 2024 & \href{https://aclanthology.org/2024.lrec-main.722.pdf}{Link} \\

\citeauthor{jinprobing_coling2025}         & \ObjRes   & \LocProb   & \None & COLING & 2025 & \href{https://aclanthology.org/2025.coling-main.37.pdf}{Link} \\

\citeauthor{chen2025stitching}    & \ObjRes   & \LocProb   & \StrVec & NeurIPS & 2025 & \href{https://openreview.net/forum?id=Qvvy0X63Fv}{Link}
\\

\rowcolor{gray!15} 
\multicolumn{7}{c}{\textbf{\textit{Logic and Reasoning (Improve Capability)}}} \\
\midrule

\citeauthor{wu2023analyzing} & \ObjToken & \LocGrad & \None & ICML & 2023 & \href{https://arxiv.org/abs/2307.13339}{Link} \\
\citeauthor{you-etal-2025-probabilistic_emnlp2025} & \ObjToken & \LocMag & \None & EMNLP & 2025 & \href{https://aclanthology.org/2025.emnlp-main.382/}{Link} \\
\citeauthor{cywinski2025interpretlatentreasoning} & \ObjToken & \LocCausal & \StrAmp & Blog & 2025 & \href{https://www.alignmentforum.org/posts/YGAimivLxycZcqRFR/can-we-interpret-latent-reasoning-using-current-mechanistic}{Link} \\
\citeauthor{cywinski2025interpretlatentreasoning} & \ObjToken & \LocCausal & \StrAmp & Blog & 2025 & \href{https://www.alignmentforum.org/posts/YGAimivLxycZcqRFR/can-we-interpret-latent-reasoning-using-current-mechanistic}{Link} \\
\citeauthor{wang2025two} & \ObjFFN & \LocMag & \StrAmp & ArXiv & 2025 & \href{https://doi.org/10.48550/arXiv.2505.14681}{Link} \\
\citeauthor{zhang2025understanding} & \ObjAttn & \LocProb & \StrAmp & ArXiv & 2025 & \href{https://arxiv.org/abs/2512.24574}{Link} \\
\citeauthor{yucausal_emnlp2024} & \ObjAttn & \LocMag & \StrAmp & EMNLP & 2024 & \href{https://doi.org/10.18653/v1/2024.emnlp-main.192}{Link} \\
\citeauthor{zhang2024interpreting} & \ObjAttn & \LocCausal & \StrOpt & ICML & 2024 & \href{https://openreview.net/forum?id=CfOtiepP8s}{Link} \\
\citeauthor{yu-ananiadou-2024-interpreting} & \ObjAttn & \LocCausal & \StrAmp & EMNLP & 2024 & \href{https://aclanthology.org/2024.emnlp-main.193/}{Link} \\
\citeauthor{yu-etal-2025-back} & \ObjAttn & \LocCausal & \None & EMNLP & 2025 & \href{https://aclanthology.org/2025.emnlp-main.567/}{Link} \\

\citeauthor{stolfo-etal-2023-mechanistic} & \ObjFFN~\&~\ObjAttn & \LocCausal & \None & EMNLP & 2023 & \href{https://doi.org/10.18653/v1/2023.emnlp-main.435}{Link} \\
\citeauthor{Aktercausal_compsac2024} & \ObjFFN~\&~\ObjAttn & \LocCausal & \None & COMPSAC & 2024 & \href{https://doi.org/10.1109/COMPSAC61105.2024.00143}{Link} \\
\citeauthor{yang2024chainofthoughtlargelanguagemodels} & \ObjFFN~\&~\ObjAttn  & \LocMag & \None & ArXiv & 2024 & \href{https://arxiv.org/abs/2412.03944}{Link} \\
\citeauthor{quirke2024understanding} & \ObjFFN~\&~\ObjAttn & \LocCausal & \StrAmp & ICLR & 2024 & \href{https://openreview.net/forum?id=rIx1YXVWZb}{Link} \\
\citeauthor{chen-etal-2025-inner} & \ObjFFN~\&~\ObjAttn & \LocGrad & \StrOpt & ACL & 2025 & \href{https://aclanthology.org/2025.acl-long.1369/}{Link} \\
\citeauthor{Hannacicuits_nips2023} & \ObjFFN~\&~\ObjAttn & \LocCirc & \None & NeurIPS & 2023 & \href{http://papers.nips.cc/paper\_files/paper/2023/hash/efbba7719cc5172d175240f24be11280-Abstract-Conference.html}{Link} \\
\citeauthor{Nikankincircuits_iclr2025} & \ObjFFN~\&~\ObjAttn & \LocCirc & \None & ICLR & 2025 & \href{https://openreview.net/forum?id=O9YTt26r2P}{Link} \\

\citeauthor{zhang2025fantastic} & \ObjSAE & \LocMag & \StrAmp & ArXiv & 2025 & \href{https://arxiv.org/abs/2512.23988}{Link} \\

\citeauthor{kim2026reasoning} & \ObjSAE & \LocMag & \StrAmp & arXiv & 2026 & \href{https://arxiv.org/abs/2601.10825}{Link} \\

\citeauthor{galichin2025have} & \ObjSAE & \LocMag & \StrAmp & ArXiv & 2025 & \href{https://arxiv.org/abs/2503.18878}{Link} \\ 
\citeauthor{pach2025sparse} & \ObjSAE & \LocMag & \StrAmp & ArXiv & 2025 & \href{https://doi.org/10.48550/arXiv.2504.02821}{Link} \\

\citeauthor{troitskii-etal-2025-internal_emnlp2025} & \ObjSAE & \LocMag & \StrAmp & EMNLP & 2025 & \href{https://aclanthology.org/2025.findings-emnlp.1012/}{Link} \\

\citeauthor{venhoff2025understandingreasoningthinkinglanguage} & \ObjRes & \LocCausal & \StrVec & ICLR & 2025 & \href{https://arxiv.org/pdf/2506.18167}{Link} \\

\citeauthor{hjer2025improvingreasoningperformancelarge} & \ObjRes & \LocCausal & \StrVec & ICLR & 2025 & \href{https://arxiv.org/abs/2504.19483}{Link} \\

\citeauthor{tang-etal-2025-unlocking} & \ObjRes & \LocCausal & \StrVec & ACL & 2025 & \href{https://aclanthology.org/2025.acl-long.339/}{Link} \\

\citeauthor{hong-etal-2025-reasoning} & \ObjRes & \LocCausal & \StrVec & ACL & 2025 & \href{https://aclanthology.org/2025.findings-acl.1111/}{Link} \\

\citeauthor{zhang2025uncoveringlatentchainthought} & \ObjRes & \LocCausal & \StrVec & ICLR & 2025 & \href{https://arxiv.org/abs/2409.14026}{Link} \\

\citeauthor{liu2025fractionalreasoninglatentsteering} & \ObjRes & \LocCausal & \StrVec & ArXiv & 2025 & \href{https://arxiv.org/abs/2506.15882}{Link} \\

\citeauthor{sinii2025steeringllmreasoningbiasonly} & \ObjRes & \LocCausal & \StrVec & EMNLP & 2025 & \href{https://arxiv.org/abs/2505.18706}{Link} \\

\citeauthor{li-etal-2025-feature} & \ObjRes & \LocCausal & \StrVec & EMNLP & 2025 & \href{https://aclanthology.org/2025.emnlp-main.552/}{Link} \\

\citeauthor{ward2025reasoningfinetuning_arxiv2025} & \ObjRes & \LocCausal & \StrVec & ICML & 2025 & \href{https://doi.org/10.48550/arXiv.2507.12638}{Link} \\
\citeauthor{Biranprobing_emnlp2024} & \ObjRes & \LocProb & \None & EMNLP & 2024 & \href{https://doi.org/10.18653/v1/2024.emnlp-main.781}{Link} \\
\citeauthor{yeprobing_iclr2025} & \ObjRes & \LocProb & \None & ICLR & 2025 & \href{https://openreview.net/forum?id=Tn5B6Udq3E}{Link} \\
\citeauthor{sun-etal-2025-probing_emnlp2025} & \ObjRes & \LocProb & \None & EMNLP & 2025 & \href{https://aclanthology.org/2025.emnlp-main.411/}{Link} \\
\citeauthor{wangelicitingcot_aaai2026} & \ObjRes & \LocProb & \StrVec & AAAI & 2026 & \href{https://arxiv.org/abs/2511.19131}{Link} \\
\citeauthor{nguyen2026atlas} & \ObjRes & \LocProb & \StrVec & ArXiv & 2026 & \href{https://arxiv.org/abs/2601.03093}{Link} \\
\citeauthor{tanvocab_2025arxiv} & \ObjRes & \LocVocab & \StrOpt & ArXiv & 2025 & \href{https://arxiv.org/abs/2512.19673v1}{Link}
\\

\rowcolor{gray!15} 
\multicolumn{7}{c}{\textbf{\textit{Efficient Training (Improve Efficiency)}}} \\
\midrule

\citeauthor{panigrahi2023taskspecificskilllocalizationfinetuned} & \ObjNeuron & \LocMag & \StrOpt & ICML & 2023 & \href{https://arxiv.org/abs/2302.06600}{Link} \\

\citeauthor{zhu-etal-2024-landermt} & \ObjNeuron & \LocGrad & \StrOpt & ACL & 2024 & \href{https://aclanthology.org/2024.acl-long.656/}{Link} \\

\citeauthor{song-etal-2024-sift} & \ObjNeuron & \LocGrad & \StrOpt & ICML & 2024 & \href{https://dl.acm.org/doi/10.5555/3692070.3693945}{Link} \\

\citeauthor{zhang-etal-2023-fine} & \ObjNeuron & \LocMag & \StrOpt & ACL & 2023 & \href{https://aclanthology.org/2023.acl-long.95/}{Link} \\

\citeauthor{xu-etal-2025-lets} & \ObjNeuron & \LocMag & \StrOpt & COLING & 2025 & \href{https://aclanthology.org/2025.coling-main.630/}{Link} \\

\citeauthor{mondal-etal-2025-language} & \ObjNeuron & \LocMag & \StrOpt & ACL & 2025 & \href{https://aclanthology.org/2025.insights-1.6/}{Link} \\

\citeauthor{gurgurov2025sparsesubnetworkenhancementunderrepresented} & \ObjNeuron & \LocMag & \StrOpt & AACL & 2025 & \href{https://arxiv.org/abs/2510.13580}{Link} \\

\citeauthor{zhao-etal-2024-multilingual} & \ObjNeuron & \LocCausal & \StrOpt & NeurIPS & 2024 & \href{https://proceedings.neurips.cc/paper_files/paper/2024/file/1bd359b32ab8b2a6bbafa1ed2856cf40-Paper-Conference.pdf}{Link} \\
\citeauthor{chenneuron} & \ObjNeuron & \LocMag & - & ICLR & 2026 & \href{https://openreview.net/forum?id=uq6UWRgzMr}{Link} \\
\citeauthor{li2025find} & \ObjNeuron & \LocMag & \None & ArXiv & 2025 & \href{https://openreview.net/forum?id=cG1EbmWiSs}{Link}
\\

\citeauthor{sergeev2025optimizingmultimodallanguagemodels} & \ObjAttn & \LocMag & \StrOpt & ICAI & 2025 & \href{https://arxiv.org/abs/2511.23375}{Link} \\

\citeauthor{olsson2022incontextlearninginductionheads} & \ObjAttn & \LocMag & \None & ArXiv & 2022 & \href{https://arxiv.org/abs/2209.11895}{Link} \\

\citeauthor{wang2024transformers} & \ObjAttn & \LocMag & \None & ArXiv & 2024 & \href{https://openreview.net/forum?id=1lFZusYFHq}{Link} \\

\citeauthor{singh2024needs} & \ObjAttn & \LocMag & \None & ICML & 2024 & \href{https://dl.acm.org/doi/abs/10.5555/3692070.3693925}{Link} \\

\citeauthor{hoogland2402developmental} & \ObjAttn & \LocMag & \None & TLMR & 2025 & \href{https://arxiv.org/abs/2402.02364}{Link} \\

\citeauthor{minegishi2025context} & \ObjAttn & \LocMag & \None & ICLR & 2025 & \href{https://openreview.net/forum?id=LNMfzv8TNb}{Link} \\

\citeauthor{lai2025jola} & \ObjAttn & \LocMag & \StrVec & ICML & 2025 & \href{https://openreview.net/pdf?id=Lllg9YjAFX}{Link}\\

\citeauthor{thilak2022slingshot} & \ObjFFN~\&~\ObjAttn& \LocMag & \None & NeurIPS & 2022 & \href{https://openreview.net/pdf?id=lY1e0PNkSJ}{Link} \\

\citeauthor{varma2023explaining} & \ObjFFN~\&~\ObjAttn& \LocMag & \None & ArXiv & 2023 & \href{https://arxiv.org/pdf/2309.02390}{Link} \\

\citeauthor{furutatowards} & \ObjFFN~\&~\ObjAttn& \LocMag & \None & TMLR & 2024 & \href{https://openreview.net/forum?id=MzSf70uXJO}{Link} \\

\citeauthor{nandaprogress} & \ObjFFN~\&~\ObjAttn& \LocMag & \None & ICLR & 2023 & \href{https://openreview.net/forum?id=9XFSbDPmdW}{Link} \\

\citeauthor{notsawo2023predicting} & \ObjFFN~\&~\ObjAttn& \LocMag & \None & ArXiv& 2023 & \href{https://arxiv.org/abs/2306.13253}{Link} \\

\citeauthor{qiye2024exploring} & \ObjFFN~\&~\ObjAttn& \LocMag & \None & ArXiv & 2024 & \href{https://arxiv.org/pdf/2412.10898}{Link} \\

\citeauthor{liu2023omnigrok} & \ObjFFN~\&~\ObjAttn& \LocMag & \None & ICLR & 2023 & \href{https://openreview.net/forum?id=zDiHoIWa0q1}{Link} \\

\citeauthor{wang2024grokking} & \ObjFFN~\&~\ObjAttn& \LocMag & \None & NeurIPS & 2024 & \href{https://proceedings.neurips.cc/paper_files/paper/2024/file/ad217e0c7fecc71bdf48660ad6714b07-Paper-Conference.pdf}{Link} \\

\citeauthor{huang2024unified} & \ObjFFN~\&~\ObjAttn& \LocMag & \None & COLM & 2024 & \href{https://arxiv.org/pdf/2402.15175}{Link} \\

\citeauthor{li2025finetuningsubgraphsearchnew} & \ObjFFN~\&~\ObjAttn& \LocCirc & \StrOpt & ArXiv & 2025 & \href{https://www.arxiv.org/pdf/2502.06106}{Link} \\

\rowcolor{gray!15} 
\multicolumn{7}{c}{\textbf{\textit{Efficient Inference (Improve Efficiency)}}} \\
\midrule

\citeauthor{xia-etal-2025-tokenskip} & \ObjToken & \LocMag & \StrAmp & EMNLP & 2025 & \href{https://aclanthology.org/2025.emnlp-main.165/}{Link} \\

\citeauthor{lei2025generictokencompressionmultimodal} & \ObjToken & \LocGrad & \StrAmp & ArXiv & 2025 & \href{https://arxiv.org/abs/2506.01097}{Link} \\

\citeauthor{guo-etal-2024-attention} & \ObjToken & \LocMag & \StrAmp & EMNLP & 2024 & \href{https://aclanthology.org/2024.emnlp-main.1178/}{Link} \\

\citeauthor{ye2025fit} & \ObjToken & \LocMag & \StrAmp & AAAI & 2025 & \href{https://ojs.aaai.org/index.php/AAAI/article/view/34366}{Link} \\

\citeauthor{he2024zipcache} & \ObjToken & \LocMag & \StrAmp & NeurIPS & 2024 &\href{https://proceedings.neurips.cc/paper_files/paper/2024/hash/7e57131fdeb815764434b65162c88895-Abstract-Conference.html}{Link} \\

\citeauthor{cai2024pyramidkv} & \ObjToken & \LocMag & \StrAmp & COLM & 2025 & \href{https://openreview.net/forum?id=ayi7qezU87}{Link} \\

\citeauthor{tang2024razorattention} & \ObjAttn & \LocCirc & \StrAmp & ICLR & 2025 & \href{https://iclr.cc/virtual/2025/poster/28028}{Link} \\

\citeauthor{xiao2024duoattention} & \ObjAttn & \LocCirc & \StrAmp & ICLR & 2025 & \href{https://openreview.net/forum?id=cFu7ze7xUm}{Link} \\

\citeauthor{bi2025unveiling} & \ObjAttn & \LocMag & \None & CVPR & 2025 & \href{https://openaccess.thecvf.com/content/CVPR2025/papers/Bi_Unveiling_Visual_Perception_in_Language_Models_An_Attention_Head_Analysis_CVPR_2025_paper.pdf}{Link} \\

\citeauthor{luofast} & \ObjAttn & \LocMag & \StrAmp & ICML & 2025 & \href{https://arxiv.org/abs/2505.00598}{Link} \\
\citeauthor{luo2026frost} & \ObjAttn & \LocMag & \StrAmp & ICLR & 2026 & \href{https://arxiv.org/abs/2601.19001}{Link} \\
\citeauthor{su2025rotatekv} & \ObjAttn & \LocMag & \StrAmp & IJCAI & 2025 & \href{https://www.ijcai.org/proceedings/2025/690}{Link} \\

\citeauthor{xiao2023streamingllm} & \ObjAttn & \LocMag & \StrAmp & ICLR & 2024 & \href{https://openreview.net/forum?id=NG7sS51zVF}{Link} \\

\citeauthor{lu2024not} & \ObjFFN & \LocMag & \StrAmp & ACL & 2024 & \href{https://aclanthology.org/2024.acl-long.334/}{Link} \\

\citeauthor{su2025unveiling} & \ObjFFN & \LocMag & \StrAmp & ArXiv & 2025 & \href{https://arxiv.org/abs/2507.23279}{Link} \\

\citeauthor{yu2024super} & \ObjFFN & \LocMag & \StrAmp & Arxiv & 2024 & \href{https://arxiv.org/abs/2411.07191}{Link}
\\

\citeauthor{liu2024unraveling} & \ObjNeuron & \LocMag & \StrAmp & ArXiv & 2024 & \href{https://openreview.net/forum?id=nUtrPN0GHX}{Link} \\

\citeauthor{tan-etal-2024-neuron} & \ObjNeuron & \LocMag & \None & EMNLP & 2024 & \href{https://aclanthology.org/2024.emnlp-main.374/}{Link} \\

\citeauthor{laitenberger2025layerswhenlearningskip} & \ObjRes & \LocMag & \StrAmp & ArXiv & 2025 & \href{https://arxiv.org/abs/2510.13876}{Link} \\

\citeauthor{valade2024acceleratinglargelanguagemodel} & \ObjRes & \LocProb & \StrAmp & ArXiv & 2024 & \href{https://arxiv.org/abs/2407.21082}{Link} \\

\citeauthor{elhoushi-etal-2024-layerskip} & \ObjRes & \LocProb & \StrAmp & ACL & 2024 & \href{https://aclanthology.org/2024.acl-long.681/}{Link} \\

\citeauthor{wang-etal-2023-hadskip} & \ObjRes & \LocMag & \StrAmp & EMNLP & 2023 & \href{https://aclanthology.org/2023.findings-emnlp.283/}{Link} \\

\citeauthor{lawson2025learningskipmiddlelayers} & \ObjRes & \LocMag & \StrAmp & ArXiv & 2025 & \href{https://arxiv.org/abs/2506.21103}{Link} \\

\citeauthor{men-etal-2025-shortgpt} & \ObjRes & \LocMag & \StrAmp & ACL & 2025 & \href{https://aclanthology.org/2025.findings-acl.1035/}{Link} \\
\citeauthor{dumitru2024layer} & \ObjRes & \LocMag & \None & ArXiv & 2024 & \href{https://arxiv.org/abs/2406.17415}{Link} \\

\citeauthor{zhang2025towards} & \ObjRes & \LocMag & \None & ArXiv & 2025 & \href{https://arxiv.org/abs/2503.06518}{Link} \\

\citeauthor{xiao2025exploring} & \ObjRes & \LocMag & \None & ArXiv & 2025 & \href{https://arxiv.org/abs/2508.03332}{Link} \\

\citeauthor{ranjan2025mix} & \ObjRes & \LocGrad & \None & ArXiv & 2025 & \href{https://arxiv.org/abs/2501.06357}{Link} \\

\citeauthor{zeng2024lsaq} & \ObjRes & \LocVocab & \None & ArXiv & 2024 & \href{https://arxiv.org/abs/2412.18135}{Link} \\

\citeauthor{shelke2024towards} & \ObjRes & \LocMag & \StrAmp & ACL & 2024 & \href{https://aclanthology.org/2024.paclic-1.68/}{Link} \\

\citeauthor{zhang2026beyond} & \ObjFFN \ \& \ObjAttn & \LocMag & - & ArXiv & 2026 & \href{https://arxiv.org/abs/2603.17354}{Link} \\

\citeauthor{lin2024awq} & \ObjFFN~\&~\ObjAttn & \LocMag & \StrAmp & MLSyS & 2024 & \href{https://proceedings.mlsys.org/paper_files/paper/2024/file/42a452cbafa9dd64e9ba4aa95cc1ef21-Paper-Conference.pdf}{Link} \\

\citeauthor{ashkboos2024quarot} & \ObjFFN~\&~\ObjAttn & \LocMag & \StrAmp & NeurIPS & 2025 & \href{https://dl.acm.org/doi/10.5555/3737916.3741096}{Link} \\

\citeauthor{su2025kvsink} & \ObjFFN~\&~\ObjAttn & \LocCirc & \None & COLM & 2025 & \href{https://openreview.net/forum?id=gIqb6zWZoO}{Link} \\

\citeauthor{xiao2023smoothquant} & \ObjFFN~\&~\ObjAttn & \LocMag & \StrAmp & NeurIPS & 2022 & \href{https://neurips.cc/virtual/2022/61716}{Link} \\

\citeauthor{sun2024massive} & \ObjFFN~\&~\ObjAttn & \LocMag & \None & NeurIPS & 2024 & \href{https://openreview.net/forum?id=F7aAhfitX6}{Link} \\

\citeauthor{an2025systematic} & \ObjFFN~\&~\ObjAttn & \LocCirc & \None & ICLR & 2025 & \href{https://openreview.net/forum?id=rLX7Vyyzus}{Link} \\

\citeauthor{NEURIPS2023_edbcb758} & \ObjFFN~\&~\ObjAttn & \LocCirc & \None & NeurIPS & 2023 & \href{https://proceedings.neurips.cc/paper_files/paper/2023/hash/edbcb7583fd8921dad78adecfe06a99b-Abstract-Conference.html}{Link} \\

\end{longtable}
\end{small}

\clearpage
\phantomsection
\bibliographystyle{refstyle}
\bibliography{all_bib}

\end{document}

%% file: Tables/summary_object_notation.tex
\begin{table*}[!t]
    \centering
    \caption{Core interpretable objects of LLMs and their mathematical notations in this paper. Here, dimensions are denoted as follows: $d_{\text{model}}$ is the model dimension, $T$ is the sequence length, $|\mathcal{V}|$ is the vocabulary size, $H$ is the number of attention heads, $d_\text{head}$ is the head dimension ($d_\text{model}/H$), $d_\text{ffn}$ is the FFN hidden dimension, and $d_\text{SAE}$ is the dictionary size of the Sparse Autoencoder.}
    \label{tab:interpretable_objects}
    \setlength\tabcolsep{5pt} 
    \fontsize{9}{13}\selectfont 
    \begin{tabular}{l l c c}
    \toprule[1.3pt]
    \textbf{Object} & & \textbf{Notation} & \textbf{Shape} \\
    \midrule
    \textbf{Token Embedding} (\S\ref{sec:token_embedding}) & Embedding Matrix & $\mathbf{W}_E$ & $\mathbb{R}^{|\mathcal{V}| \times d_{\text{model}}}$ \\
    & Token $i$ Embedding (Input) & $\mathbf{x}^0_{i}$ & $\mathbb{R}^{d_{\text{model}}}$ \\
    \addlinespace[2pt] \hdashline[1pt/1pt] \addlinespace[2pt]
    \textbf{Residual Stream} (\S\ref{sect:block_stream}) & Residual Stream State & $\mathbf{x}^l$ & $\mathbb{R}^{T \times d_{\text{model}}}$ \\
    & Intermediate State (Post-Attn) & $\mathbf{x}^{l, \text{mid}}$ & $\mathbb{R}^{T \times d_{\text{model}}}$ \\
    \addlinespace[2pt] \hdashline[1pt/1pt] \addlinespace[2pt]
    \textbf{MHA} (\S\ref{sec:mha}) & Q, K, V, O Weight Matrices & $\mathbf{W}_{Q,K,V,O}^{l,h}$ & $\mathbb{R}^{d_{\text{model}} \times d_\text{head}}$ / $\mathbb{R}^{d_\text{head} \times d_{\text{model}}}$ \\
    & Attention Score Matrix & $\mathbf{A}^{l,h}$ & $\mathbb{R}^{T \times T}$ \\
    & Head Output & $\mathbf{h}_{\text{attn}}^{l,h}$ & $\mathbb{R}^{T \times d_\text{model}}$ \\
    & Block Output & $\mathbf{h}_{\text{attn}}^{l}$ & $\mathbb{R}^{T \times d_{\text{model}}}$ \\
    \addlinespace[2pt] \hdashline[1pt/1pt] \addlinespace[2pt]
    \textbf{FFN} (\S\ref{sec:ffn}) & In Projection (Key) Matrix & $\mathbf{W}_{\text{in}}^l$ & $\mathbb{R}^{d_{\text{model}} \times d_\text{ffn}}$\\
    & Out Projection (Value) Matrix & $\mathbf{W}_{\text{out}}^l$ & $\mathbb{R}^{d_\text{ffn} \times d_{\text{model}}}$ \\
    & Block Output & $\mathbf{h}_\text{ffn}^{l}$ & $\mathbb{R}^{T \times d_{\text{model}}}$ \\
    \addlinespace[2pt] \hdashline[1pt/1pt] \addlinespace[2pt]
    \textbf{Neuron} (\S\ref{sec:ffn})
    & Neuron Activation State & $\mathbf{s}^l$ & $\mathbb{R}^{d_\text{ffn}}$ \\
    & $j$-th Neuron Activation & ${s}^l_j$ & $\mathbb{R}$ (Scalar) \\
    & $j$-th Neuron Key Weight & $\mathbf{k}_{j}^l$ & $\mathbb{R}^{d_{\text{model}}}$ \\
    & $j$-th Neuron Value Weight & $\mathbf{v}_{j}^l$ & $\mathbb{R}^{d_{\text{model}}}$ \\
    \addlinespace[2pt] \hdashline[1pt/1pt] \addlinespace[2pt]
    \textbf{SAE Feature} (\S\ref{sec:sae_feature}) & Feature Activation State& $\mathbf{a}$ & $\mathbb{R}^{d_{\text{SAE}}}$ \\
    & $j$-th Feature Activation & $a_j$ & $\mathbb{R}$ (Scalar) \\
    & $j$-th Feature & $\mathbf{f}_j$ & $\mathbb{R}^{d_{\text{model}}}$ \\
    \bottomrule[1.3pt]
    \end{tabular}
    \end{table*}

%% file: Tables/comparative_localizing_methods.tex
\begin{table*}[!t]
    \centering
    \caption{Comparative analysis of localizing methods. These methods differ mainly in causal strength, computational cost, required access, and typical limitations.}
    \label{tab:localizing_methods_comparison}
    \small
    \setlength{\tabcolsep}{3pt}
    \renewcommand{\arraystretch}{1.1}
    \begin{tabular}{p{0.28\textwidth} p{0.15\textwidth} p{0.09\textwidth} p{0.20\textwidth} p{0.23\textwidth}}
    \toprule[1.3pt]
    \textbf{Method} & \textbf{Causal Strength} & \textbf{Cost} & \textbf{Required Access} & \textbf{Key Limitation} \\
    \midrule
    \textbf{Magnitude Analysis} (\S\ref{sec:magtitude_analysis}) & Correlational & Low & Activations / Weights & Salience may not reflect causal necessity \\
    \textbf{Gradient Detection} (\S\ref{sec:gradient_detection}) & Moderate & Medium & Gradients / Weights & Local proxy; May fail sanity checks \\
    \textbf{Causal Attribution} (\S\ref{sec:causal_Attribution}) & Strong & High & Activations / Hidden States / Weights & Expensive; Intervention may introduce artifacts \\
    \textbf{Probing} (\S\ref{sec:probing}) & Weak & Medium & Hidden States / Labeled Data & Decodability does not imply causality \\
    \textbf{Vocabulary Projection} (\S\ref{sec:vocab_project}) & Correlational & Low & Hidden States / Unembedding Matrix & Depends on representation alignment \\
    \textbf{Circuit Discovery} (\S\ref{sec:circuit_discovery}) & Strong & High & Activations / Gradients / Hidden States / Weights & Sensitive to setup and contrastive inputs \\
    \bottomrule[1.3pt]
    \end{tabular}
\end{table*}

%% file: Tables/comparative_steering_methods.tex
\begin{table*}[!t]
    \centering
    \caption{Comparative analysis of steering methods. These interventions differ mainly in steering strength, side effects, required access, computational cost, and typical limitations.}
    \label{tab:steering_methods_comparison}
    \small
    \setlength{\tabcolsep}{3pt}
    \renewcommand{\arraystretch}{1.1}
    \begin{tabular}{L{0.27\textwidth} L{0.10\textwidth} L{0.11\textwidth} L{0.18\textwidth} L{0.08\textwidth} L{0.18\textwidth}}
    \toprule[1.3pt]
    \textbf{Method} & \textbf{Steering Strength} & \textbf{Side Effects} & \textbf{Required Access} & \textbf{Cost} & \textbf{Key Limitation} \\
    \midrule
    \textbf{Amplitude Manipulation} (\S\ref{sec:amplitude_manipulation}) & Moderate & Moderate & Activations / Hidden States / Weights & Low & Polysemantic features may cause collateral changes \\
    \textbf{Targeted Optimization} (\S\ref{sec:target_optimization}) & High & Low & Weights / Gradients & High & Depends on accurate localization \\
    \textbf{Vector Arithmetic} (\S\ref{sec:vector_arithmetic}) & Moderate & Moderate & Hidden States / Weights & Low & Assumes linear and separable directions \\
    \bottomrule[1.3pt]
    \end{tabular}
    \end{table*}

%% file: Tables/evaluation_protocal.tex
\begin{table*}[ht!]
    \centering
    \caption{A minimal evaluation benchmark framework for actionable MI. This framework combines primary success metrics with essential side effect checks to ensure a holistic assessment of any intervention.}
    \label{tab:evaluation_framework}
    \small
    \renewcommand{\arraystretch}{1.3} 
    \begin{tabularx}{\textwidth}{
      >{\raggedright\arraybackslash}p{0.16\textwidth} 
      >{\raggedright\arraybackslash}X                 
      >{\raggedright\arraybackslash}X 
      >{\raggedright\arraybackslash}X 
      >{\raggedright\arraybackslash}X 
    }
    \toprule
    \textbf{Capability} & \textbf{Feature to Localize/ Steer} & \textbf{Primary Metric} & \textbf{Side Effect Check} & \textbf{Example Datasets} \\
    \midrule
    \textbf{Math Reasoning} & Arithmetic and Reflection Features & Accuracy, Reflection Token Ratio & General Capabilities & GSM8K, AIME25 \\
    \addlinespace 
    \textbf{Safety} & Refusal Features & Refusal Rate, Attack Success Rate & Over-Refusal & SORRY-Bench, OR-Bench \\
    \addlinespace
    \textbf{Knowledge} & Factual Associations & Edit Success Rate & Locality, Fluency & CounterFact, ZsRE \\
    \bottomrule
    \end{tabularx}
    \end{table*}